\documentclass[sigconf,noacm]{acmart}
\setcopyright{none}
\renewcommand\footnotetextcopyrightpermission[1]{}
\usepackage{multirow}
\usepackage{longtable}
\usepackage{xurl}  
\usepackage{enumitem}
\usepackage{stfloats}
\usepackage{caption}
\usepackage{float}
\usepackage{placeins}
\usepackage{float}
\author{Xin Luo}
\authornote{These authors contributed equally to this research.}
\affiliation{%
  \institution{University of Michigan}
  \city{Ann Arbor}
  \state{Michigan}
  \country{USA}
}
\email{luosanj@umich.edu}

\author{Yicheng Tao}
\authornotemark[1] 
\affiliation{%
  \institution{University of Michigan}
  \city{Ann Arbor}
  \state{Michigan}
  \country{USA}
}
\email{yctao@umich.edu}

\author{Haoxuan Zeng}
\authornotemark[1] 
\affiliation{%
  \institution{University of Michigan}
  \city{Ann Arbor}
  \state{Michigan}
  \country{USA}
}
\email{zchx@umich.edu}

\author{Suyuan Wang}
\affiliation{%
  \institution{University of Michigan}
  \city{Ann Arbor}
  \state{Michigan}
  \country{USA}
}
\email{suyuanw@umich.edu}

\author{Chenzi Ouyang}
\affiliation{%
  \institution{University of Michigan}
  \city{Ann Arbor}
  \state{Michigan}
  \country{USA}
}
\email{oooycz@umich.edu}

\author{Meiqi Zhu}
\affiliation{%
  \institution{University of Michigan}
  \city{Ann Arbor}
  \state{Michigan}
  \country{USA}
}
\email{meiqizhu@umich.edu}

\author{Kai Liu}
\affiliation{%
  \institution{University of Michigan}
  \city{Ann Arbor}
  \state{Michigan}
  \country{USA}
}
\email{kailiua@umich.edu}

\author{Shuibing Chen}
\authornote{Corresponding author.}
\affiliation{%
  \institution{Weill Cornell Medicine}
  \city{New York}
  \state{New York}
  \country{USA}
}
\email{shc2034@med.cornell.edu}

\author{Jie Liu}
\authornotemark[2]
\affiliation{%
  \institution{University of Michigan}
  \city{Ann Arbor}
  \state{Michigan}
  \country{USA}
}
\email{drjieliu@umich.edu}

\title[VOICE: A Vision-Omics Foundation Model]{
VOICE: A Vision-Omics Foundation Model Integrating Direct and Retrieval-Based Prediction of In-situ Single-Cell Gene Expression
}

\acmConference[]{}{}{}

\begin{document}
\begin{abstract}
Spatial transcriptomics can resolve gene expression at single-cell resolution, but it is costly, limited to targeted panels of a few hundred to thousand genes, and applicable to only a small number of samples. H\&E imaging, by contrast, is cheap and collected routinely at scale. This makes predicting single-cell expression directly from morphology a practical way to bring molecular analysis to large tissue archives. We therefore present VOICE, a multimodal foundation model that predicts single-cell gene expression from H\&E images using paired Xenium data. VOICE first aligns cell-centered H\&E morphology from a pathology foundation model with single-cell expression embeddings from a transcriptome foundation model, trained using contrastive learning over 23 million cells. Next it predicts expression through two branches. One branch directly regresses expression from morphology. The other branch retrieves measured expression from similar reference cells, complementing signals poorly captured by direct prediction. Because genes vary in morphological predictability, VOICE fuses the two branches with a per-gene weight. After training, VOICE generalizes across held-out Xenium slides and partially overlapping gene panels, with evaluation on ten slides across seven tissues, including two tissues absent from training. Downstream analyses further show recovery of spatial domains, pathway activity, proliferation, and expression differences between cell subtypes.
\end{abstract}
\maketitle

\section{Introduction}
Single-cell spatial transcriptomics technologies, including 10x Genomics Xenium \cite{janesick2023high} and CosMx Spatial Molecular Imaging ~\cite{he2022high} data, enable transcriptomic profiling of individual cells while preserving their native tissue context. These assays provide an increasingly detailed view of how cellular states, tissue morphology, and local microenvironments interact in situ, creating new opportunities to study cell-type organization and spatially cellular interactions. However, single-cell spatial transcriptomics data remains costly and technically demanding. A single Xenium or CosMx slide typically costs several thousand US dollars, limited to its measured gene panels of a few hundred to thousand genes. In contrast, hematoxylin and eosin (H\&E) imaging is much cheaper and easy to acquire, which costs only a few tens of dollars per slide, two to three orders of magnitude less than Xenium and CosMx. H\&E slides are also routinely collected at scale in both research studies and clinical applications. Learning transferable relationships between histological morphologies and molecular states among cells could therefore extend spatial molecular analysis to large retrospective tissue collections.

Recent methods have demonstrated that cellular gene expression can be inferred from H\&E images by training with spatial transcriptomics data. For example, GHIST directly integrates paired histology and subcellular spatial transcriptomics to jointly model nuclear morphology, cell identity, local neighborhood composition, and single-cell gene expression ~\cite{ghist2025}. sCellST uses weakly supervised multiple-instance learning to decompose Visium spot-level gene expression into cell-level predictions \cite{scellst2026}. These methods establish that tissue morphology contains informative molecular signals, but their learned predictors are generally limited to a particular cohort, tissue type, spatial platform, or predefined gene panel, which cannot learn morphology–molecular relationships that transfer across datasets. To resolve this limitation, recently there are a few foundation models trained to predict spatial transcriptomics from H\&E images like OmiCLIP ~\cite{omiclip2025}, STORM~\cite{storm2026}, and DeepSpot-M~\cite{nonchev2026deepspotm}, inspired by the recent advances in general-purpose multimodal foundation models, including vision-language models and world models ~\cite{radford2021learning,rombach2022high,dong2026dynamic,zhu2026ants,zhang2024dual,ke2026deformba,ke2025mambev,chow2026masked,kong2026driving,xiao2026reversible,liu2026affordance,zhou2026comem}. However, these foundation models are primarily trained on spot-resolution data and treat all genes uniformly. In practice, some genes are strongly associated with cell morphology and can be predicted directly from the image, while others are better estimated by referring to cells with similar transcriptomic profiles. A single prediction strategy is therefore unlikely to be optimal for all genes.

In this work, we propose VOICE, a multimodal spatial foundation model that predicts single-cell gene expression from H\&E using paired Xenium data. Our model first aligns cell-centered H\&E morphology with single-cell transcriptomic foundation model in a shared embedding space through contrastive learning. Then it predicts gene expressions of every query cell by two complementary branches from this space. One branch directly regresses gene expression from cell morphology, and the other branch retrieves expression from reference cells whose aligned embeddings are closest to the query's, complementing signals poorly captured by direct prediction. Because genes vary in morphological predictability, VOICE fuses the two branches' predictions per gene to produce the final prediction. After training, VOICE generalizes to held-out patients and slides. It can predict genes measured in the target tissue's test set but never seen in any training slides from the same tissue. Therefore, it can transfer knowledge of these genes from other tissues.

We summarize our contributions as follows:
\begin{itemize}[topsep=2pt, itemsep=2pt, partopsep=0pt, parsep=0pt]
\item \textbf{Panel-independent contrastive alignment.} We align cell-level morphology from UNI2-h~\cite{chen2024towards} with frozen scFoundation~\cite{hao2024large} expression embeddings using LoRA adaptation and a symmetric InfoNCE loss over 23 million cells from 15 tissues and 75 slides. We represent cell morphology representation with a cell-mask pooling strategy, reducing noise from surrounding cells. This establishes a panel-independent vision--omics representation that links cellular morphology with gene-expression state.
\item \textbf{Gene-supervised parametric continuation.} Starting from the aligned checkpoint of UNI2-h, we attach an SE(2)-equivariant spatial-attention decoder and a count head over a shared 6{,}029-gene vocabulary pooled across all slides, and continue LoRA finetuning end-to-end, trained by a log-space MSE plus a negative-binomial likelihood. This count head is shared across slides and masked to each slide's own panel, so heterogeneous panels train one predictor together, giving an accurate per-cell estimate.
\item \textbf{Per-gene gated fusion of two branch prediction results.} The cell morphology representation is shared by two prediction branches and is used to construct a \textbf{reference-cell bank} that stores image-derived reference features together with their measured expression profiles. One branch retrieves expression from similar cells, the other predicts morphology-related genes, so the two make partly independent errors. We fuse them with a per-gene convex gate whose weight is fit per gene to favor the more reliable branch to yield the final prediction.
\end{itemize}

\section{Related Work}
\label{sec:related}

\noindent\textit{Histology-based spatial transcriptomics prediction.}
Early methods predicted spot-level gene expression from H\&E image patches ~\cite{zhang2024istar}.
Foundation models such as OmiCLIP~\cite{omiclip2025} improved spot-level
accuracy and cross-cohort transfer through large-scale contrastive alignment
of histology and spatial transcriptomic profiles, with its Loki platform
estimating expression for a query patch by retrieving and aggregating
similar reference spots. PathOmCLIP~\cite{lee2024pathomclip} likewise aligns histology with transcriptomic foundation-model embeddings and aggregates a fixed set of neighboring patches at Visium spot resolution.
These approaches, however, operate at the spot level and cannot resolve the
cellular heterogeneity within each capture region.

To reach single-cell resolution, one line of work trains on paired histology
and subcellular spatial transcriptomics data.
GHIST~\cite{ghist2025} jointly predicts cell type, morphology, neighborhood
composition, and single-cell expression by combining cellular and tissue
context.
STORM~\cite{storm2026} and DeepSpot-M~\cite{nonchev2026deepspotm} are multimodal foundation models pretrained on spot-level data and adapted to
high-resolution assays such as Xenium and CosMx.
STORM couples histology, expression, coordinates, and neighborhood in a hierarchical architecture, while DeepSpot-M uses heterogeneous molecular
embeddings and gene queries to attend to image features, enabling
adaptation to new cohorts from limited data. More recently, Phoenix~\cite{phoenix2026} scales direct single-cell
supervision across tissues and gene panels using millions of paired histology--expression observations from Xenium. It employs a latent flow-matching generative model conditioned on pathology image features to predict single-cell gene expression, and demonstrates zero-shot transfer across independent cohorts, including previously unseen tissues and organs. Despite these advances, directly supervised approaches remain constrained by the gene panels available in high-resolution spatial assays, making generalization across datasets with heterogeneous gene sets challenging.

A separate line of work learns cell-level expression from spot-level
supervision alone. sCellST~\cite{scellst2026} treats cells within each
Visium spot as a bag and mean-pools per-cell predictions to match the
observed spot expression. DeepSpot2Cell~\cite{deepspot2cell2025} follows
a similar strategy but adds multiscale context from the cell and its
neighboring spots. These methods avoid subcellular spatial data but face
an underdetermined decomposition, as multiple cellular configurations can
explain the same spot-level observation.\\

\noindent\textit{Pathology foundation models.}
Self-supervised pretraining on large collections of whole-slide images has become a standard approach for learning transferable patch-level representations in computational pathology ~\cite{deng2009imagenet}. Representative models include CTransPath, which combines contrastive learning with a hierarchical vision transformer, and UNI, which pretrains a ViT-L encoder on approximately 100,000 diagnostic slides ~\cite{wang2022transformer,chen2024towards}. Although these models differ in architecture and pretraining strategy, both are designed primarily to capture histological morphology. Consequently, their embeddings capture rich visual patterns but do not explicitly represent gene-expression programs.\\

\noindent\textit{Single-cell transcriptomic foundation models.}
Large-scale pretraining on single-cell RNA-seq atlases has produced
general-purpose cell embeddings that capture transcriptomic programs across
tissues and conditions, notably
scGPT~\cite{cui2024scgpt},
Geneformer~\cite{theodoris2023transfer}, and
scFoundation~\cite{hao2024large}, which pretrains on over 50 million cells
to produce unified cell and gene embeddings.
These models define a semantically meaningful transcriptomic latent space
but operate entirely on molecular measurements with no access to tissue
morphology.\\

\noindent\textit{Cross-modal transfer and representation learning.}
Cross-modal learning exploits complementary measurements through knowledge
transfer or joint representation learning. OC-Distill transfers information
from clinical notes to a vital-sign model during training while requiring only
vital signs at inference~\cite{liang2026oc}. RiVaT-Fuse estimates a fused
image--metadata representation using learned reliability
operators~\cite{xu2026rivatfuse}. Complementing representation-level fusion, validation-stage Combinatorial Fusion Analysis (CFA) selects a small subset of complementary classifiers and diversity-aware fusion weights for credit-card fraud detection~\cite{han2026validationstagecombinatorialfusionanalysis}. These approaches focus on cross-modal
representations, leaving the integration of direct and reference-based predictions as a separate problem.\\

\noindent\textit{Retrieval-augmented prediction.}
Retrieval-augmented prediction supplements learned models with evidence from
relevant reference examples. In cross-video procedure planning, OSEF avoids
committing to a single candidate window by passing soft evidence to a downstream
planner~\cite{ye2026osef}. These studies address representation learning or classifier combination, while gene-specific integration of direct and reference-based expression predictions remains a distinct problem.\\

\section{Methodology}
\label{sec:method}

\begin{figure*}[t]
  \centering
  \includegraphics[width=\textwidth]{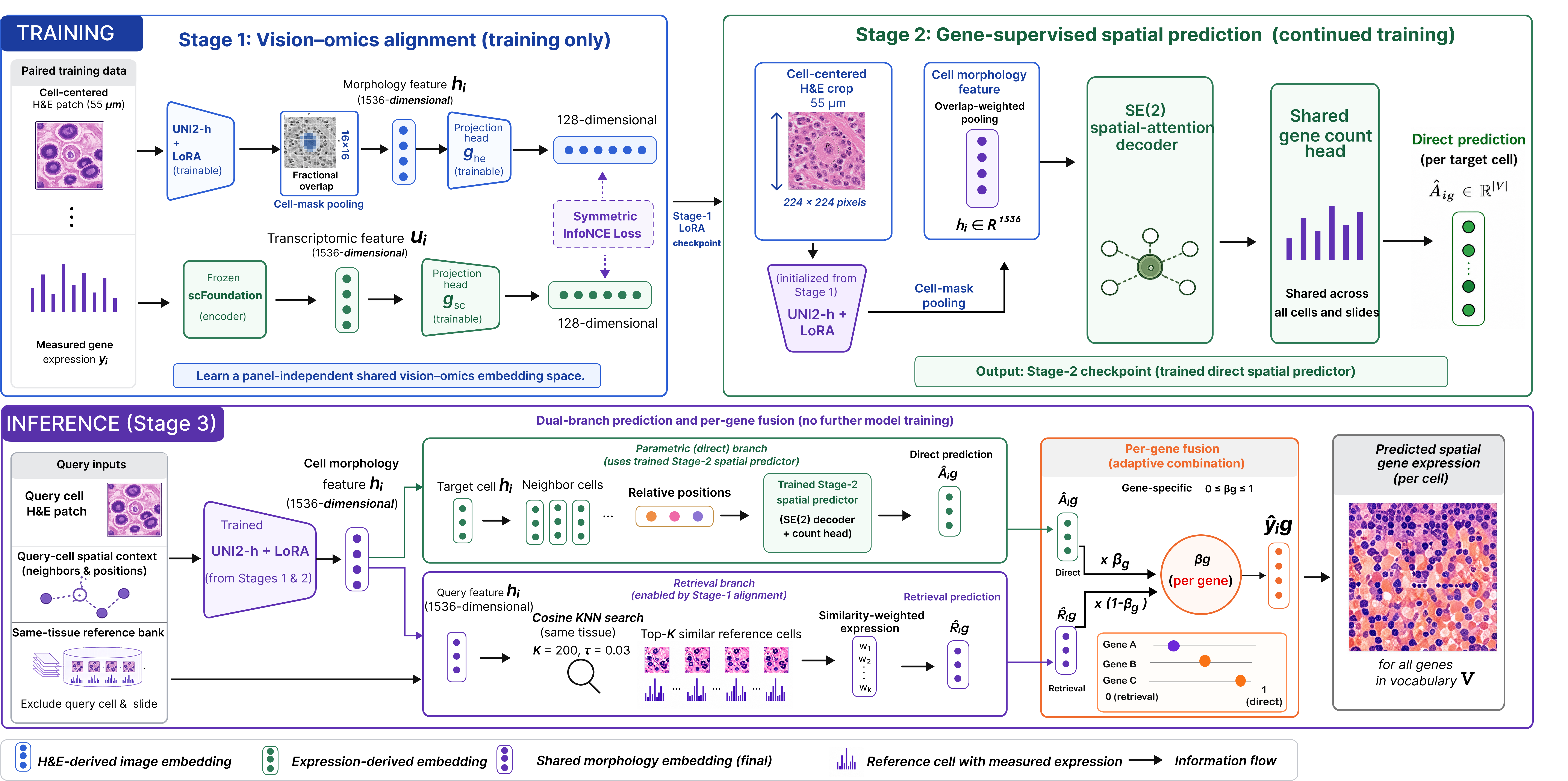}
  \caption{\textbf{Overview of VOICE.} VOICE predicts a single cell's gene expression from its H\&E appearance in three stages.}
  \label{fig:arch}
\end{figure*}

\subsection{Problem setting and notation}
Let $\mathcal{S}$ be a set of tissue slides, each slide $s\in\mathcal{S}$ provides a registered H\&E whole-slide image together with a Xenium assay.  For each of its $N_s$ segmented cells, we observe its image centroid coordinate $p_i\in\mathbb{R}^2$, a cell-boundary polygon $\Omega_i$, and raw counts $y_i\in\mathbb{Z}_{\geq0}^{G_s}$ for the $G_s$ genes measured by that slide's panel. Writing $z_i=\log(1+y_i)$, our goal is to predict $z_i$ from H\&E cell morphology and similar reference cells.

VOICE learns the prediction of genes in two training stages and fuses the prediction results at per gene level from two branches (Fig.~\ref{fig:arch}). Stage~1 aligns cell morphology with its gene-expression profile through contrastive learning. Specifically, LoRA is used to fine-tune the UNI2-h image encoder, producing an image adapter that maps morphological embeddings into the transcriptomic latent space defined by scFoundation~\cite{chen2024towards,hao2024large}. Stage~2 continues finetuning the same image adapter against measured gene counts jointly with an SE(2) spatial decoder learning the relationship between each cell and its neighboring cells. The two prediction branches then use the same final morphology representation from the model, with one regressing gene expression from the cell morphology embedding, and the other retrieving similar reference cells from the cell embedding space. Finally, a per-gene gate fuses the two predictions, relying on the direct branch where morphology predicts a gene well and on retrieval otherwise.

\subsection{Cell-resolved image features}
\begin{figure}[t]
    \centering
    \includegraphics[
        width=\columnwidth
    ]{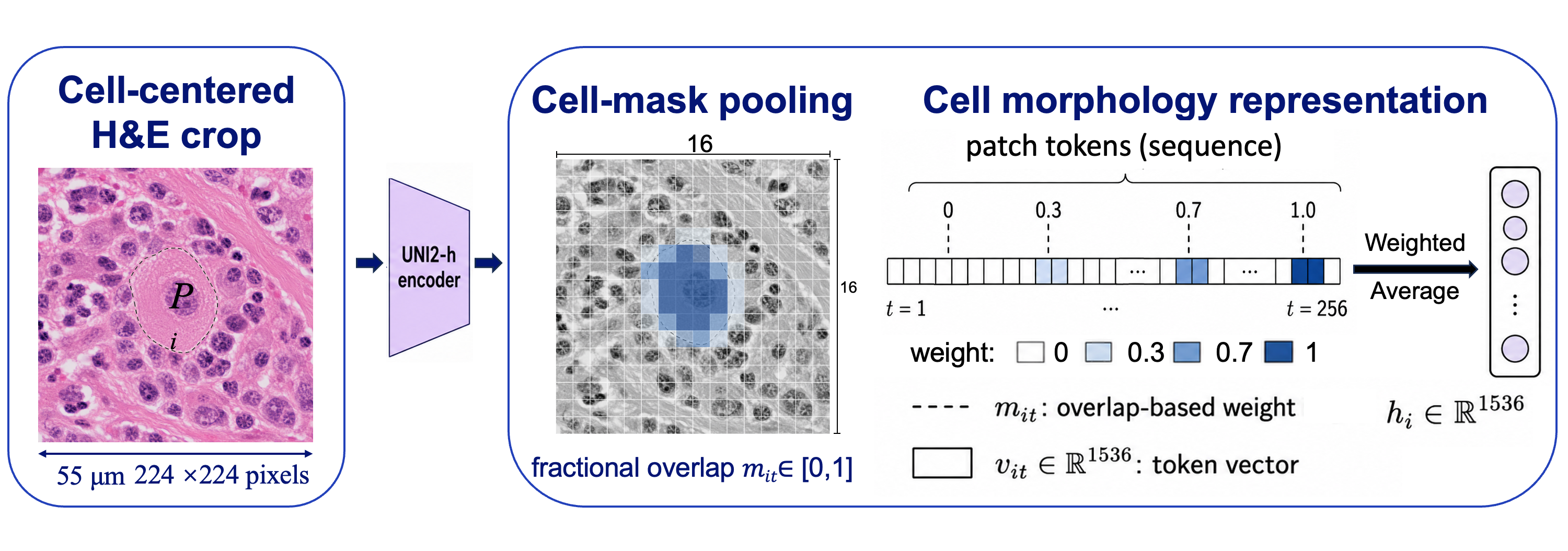}
    \caption{Cell–patch overlap-weighted token pooling for cell morphology representation.
    }
    \label{fig:cell-resolved-features}
\end{figure}

To encode cell morphology from H\&E images, for each cell $i$, we extract a $55\,\mu$m H\&E field of view centered at $p_i$ and resample it to $224\times224$ pixels (Appendix~\ref{app:preproc}). This crop includes the surrounding region of the cell so that passing it through UNI2-h attends it to its neighborhood. UNI2-h converts the crop into a $16\times16$ grid of 1536-dimensional patch tokens which interacts with other tokens to learn spatial relationship. However, if we pool over the whole crop, we would mix the cell with its neighborhood and wash out its own morphology signal. Therefore, we rasterize the cell boundary $\Omega_i$ within the image crop and compute the fractional overlap $m_{it}\in[0,1]$ between cell $i$ and patch token $\mathbf{v}_{it}$. The cell morphology representation is calculated as an overlap-weighted average of the patch token embeddings (Fig.~\ref{fig:cell-resolved-features}). 
\begin{equation}
  h_i
  = \frac{\sum_{t=1}^{256} m_{it}v_{it}}
         {\sum_{t=1}^{256} m_{it}},
  \qquad v_{it}\in\mathbb{R}^{1536}.
  \label{eq:cellmask}
\end{equation}
Replacing the cell-mask pooling with a uniform average over the crop lowers All-gene PCC by $0.086$ at a matched crop size (Appendix~\ref{sec:ablation}).

\subsection{Stage 1: contrastive LoRA alignment to transcriptomic embedding space}
\label{sec:stage1}

Our aim in this stage is to align cell morphology representations to the transcriptomic embedding space, such that cells with similar transcriptomic profiles are also close in the morphology embedding space. 
We achieve this by contrastively aligning morphology embeddings from UNI2-h image encoder with transcriptomic embeddings produced by scFoundation. The morphology representation is the embedding obtained by the cell-mask pooling Eq.~\eqref{eq:cellmask} to the target cell.

UNI2-h has 24 attention layers. We insert LoRA adapters (rank $16$, $\alpha = 32$, dropout $0.05$) into the query, key, value, and output projections of the final $12$ attention layers, and leave all original weights frozen. Writing $f_\theta$ for this adapted encoder followed by the pooling of Eq.~\eqref{eq:cellmask}, we denote the adapted feature of cell $i$ by $h_i^\theta=f_\theta(x_i,\Omega_i)$, where $x_i$ is the cell's crop and $\Omega_i$ is its boundary. 

To construct the transcriptomic alignment target, we pass each training cell's measured expression profile through scFoundation~\cite{hao2024large}, and the resulting embedding $u_i \in \mathbb{R}^{3072}$ is fixed. Because scFoundation represents each cell using its own unified gene vocabulary rather than the gene panel of a particular Xenium assay, cells measured with different panels are embedded in a common transcriptomic space. This allows the model to learn jointly from datasets with heterogeneous gene panels. We use two lightweight projection heads, $g_{\text{he}} : \mathbb{R}^{1536}\!\to\!\mathbb{R}^{128}$ and $g_{\text{sc}} : \mathbb{R}^{3072}\!\to\!\mathbb{R}^{128}$ to map the morphology feature $h_i^\theta$ and the transcriptomic embedding $u_i$ into one shared $128$-dimensional space. We then optimize the symmetric CLIP objective~\cite{radford2021clip}, an InfoNCE loss, over a batch $\mathcal{B}$ that pulls each cell's image and its own expression together while pushing apart mismatched pairs:
\begin{equation}
\begin{split}
  \mathcal{L}_{\text{Stage-1}}
  = -\frac{1}{2|\mathcal{B}|}\sum_{i\in\mathcal{B}}
  \Bigg[ \;
    &\log \frac{e^{\,\langle a_i, b_i\rangle/\tau_c}}{\sum_{j\in\mathcal{B}} e^{\,\langle a_i, b_j\rangle/\tau_c}} \\[-2pt]
    +\; &\log \frac{e^{\,\langle a_i, b_i\rangle/\tau_c}}{\sum_{j\in\mathcal{B}} e^{\,\langle a_j, b_i\rangle/\tau_c}}
  \Bigg],
\end{split}
\label{eq:infonce}
\end{equation}
with $a_i = \overline{g_{\text{he}}(h_i^\theta)}$, $b_i = \overline{g_{\text{sc}}(u_i)}$ ($\overline{\,\cdot\,}$ denotes $\ell_2$ normalization) and a \emph{fixed} temperature $\tau_c = 0.07$ to sharpen the softmax distribution into true pairs. 

We train the adapter and both projection heads for three epochs with AdamW optimizer. The adapter learning rate is $1.5\times10^{-4}$, the projection-head learning rate is $10^{-3}$, weight decay is $0.01$, and the first $150$ updates linearly warm up the learning rates. In addition, we use GradCache to achieve a large effective contrastive batch of $2048$ cells while forwarding $128$ cells at a time. 

\textit{Retrieval Prediction}
Given a bank of cells with known expression, we predict a query cell's profile as a similarity-weighted average of its neighbours' profiles,
\begin{equation}
\begin{split}
  \hat{y}_i &\;=\; \sum_{k=1}^{K} w_{ik}\, \log(1 + y_{\pi_k(i)}), \\
  w_{ik} &\;=\; \frac{\exp\!\big(\cos(h_i^\theta, h_{\pi_k(i)}^\theta)/\tau\big)}{\sum_{k'} \exp\!\big(\cos(h_i^\theta, h_{\pi_{k'}(i)}^\theta)/\tau\big)},
\end{split}
\label{eq:knn}
\end{equation}
where $\pi_k(i)$ indexes the $k$-th nearest bank cell under cosine similarity. We compare cells in the raw $1536$-dimensional feature $h_i^\theta$ rather than in its $128$-dimensional projection $g_{\text{he}}(h_i^\theta)$, because the projection is trained only to align the two modalities and disregards morphology information that remains useful to identify similar cells. We use $K = 200$ neighbors and $\tau = 0.03$ in all experiments.


For cross-slide prediction, the reference bank consists of training cells from the same tissue and excludes the query slide. In-slide prediction instead uses reference cells from the training bands of the same slide. Because different slides can measure different gene panels, retrieval is performed separately for each gene. A gene is predicted using only reference cells whose panels contain that gene. When the bank includes multiple slides, each gene is standardized separately within each slide before the cells are pooled. This reduces slide-specific differences in expression scale and sequencing depth. If none of the bank cell measures a requested gene, retrieval cannot predict it and returns zero.

\subsection{Stage 2: Spatial Gene-Expression Prediction by Continued LoRA Fine-Tuning}
\label{sec:decoder}

In Stage 2, we retain the LoRA-adapted encoder from Stage 1 and continue finetuning it jointly with a spatial decoder $D_\phi$ to predict gene expressions. Because cellular expression is related to both a cell's morphology and its local environment, the decoder predicts cell $i$ from its feature $h_i^\theta$ together with the features and coordinates of cells in its spatial neighborhood $\mathcal{N}(i)$. It uses an SE(2)-equivariant transformer \cite{zhou2023uni} to aggregate this local context. For each target cell $i$, the attention weight assigned to a neighboring cell $j$ is computed as 
\begin{equation} \alpha_{ij} = \operatorname{softmax}_{j\in\mathcal{N}(i)} \left( \frac{\mathbf{q}_i^{\top}\mathbf{k}_j}{\sqrt{d}} + b(r_{ij}) \right), 
\end{equation} 
where $\mathbf{q}_i$ is the query embedding of target cell $i$, $\mathbf{k}_j$ is the key embedding of neighboring cell $j$, and $d$ is their dimensionality. The relative distance between the two cells is $r_{ij}=\lVert\mathbf{p}_j-\mathbf{p}_i\rVert_2$, where $\mathbf{p}_i$ and $\mathbf{p}_j$ denote their spatial coordinates. The scalar distance $r_{ij}$ is expanded using $K$ Gaussian radial basis functions, \begin{equation} 
\phi_k(r_{ij}) = \exp\left( -\frac{(r_{ij}-\mu_k)^2}{2\sigma^2} \right), \qquad k=1,\ldots,K, 
\end{equation} 
where $\mu_k$ is the center of the $k$-th basis function and $\sigma$ controls its width. The resulting radial basis embedding is mapped to the geometric bias $b(r_{ij})$. The transformer then aggregates neighboring cell embeddings according to both morphological similarity and spatial separation. Because the attention depends only on relative distances, the decoder is invariant to global translations and rotations of the slide. The resulting neighborhood-aware representation is passed to the count prediction head.

The count head is a two-layer MLP with a softplus output, so it yields one non-negative value $\mu_{ig}$ for every gene in cell $i$, over a shared gene vocabulary $\mathcal{V}$ pooled across all training slides. We train the decoder with two terms, a squared error in log space plus a negative-binomial (NB) likelihood in count space:
\begin{equation}
\begin{split}
  \mathcal{L}_{\text{Stage-2}}
  = \sum_{i}\sum_{g \in \text{panel}(s_i)} \Big[\, &\big(\log(1{+}\mu_{ig}) - \log(1{+}y_{ig})\big)^2 \\[-1pt]
  &-\; \tfrac{1}{2}\,\log \mathrm{NB}\!\left(y_{ig} \,\middle|\, \mu_{ig}, \theta_g\right) \Big],
\end{split}
\label{eq:nb}
\end{equation}
where the cell's predicted mean $\mu_{i} = D_\phi(\{h_j^\theta, p_j\}_{j \in \mathcal{N}(i)})$ is computed from the features and positions of the cells in its neighborhood $\mathcal{N}(i)$, and $\theta_g$ is a learnable per-gene dispersion used only by the NB term. The count head has one output for every gene in the pooled vocabulary $\mathcal{V}$, whereas each slide measures only a subset of these genes. Slides with heterogeneous panels therefore train different subsets of the same shared head and collectively provide supervision across the full vocabulary. The two loss terms are complementary, with log-space MSE reducing the dominance of highly expressed genes, while the negative-binomial likelihood models gene-specific variability and overdispersion for sparsely expressed genes.

For Stage 2 optimization, the first three LoRA-adapted attention blocks remain frozen, while the remaining nine are fine-tuned at \(3\times10^{-5}\) and the decoder at \(10^{-4}\), using AdamW with \(10^{-4}\) weight decay and 200 warm-up updates.

\subsection{Stage 3: per-gene fusion of the direct and retrieval predictions}
\label{sec:gate}

Stages 1 and 2 result in gene expression predictions from two branches for the target cell. The final LoRA-adapted image encoder is shared by both prediction branches. For each target cell, the direct branch passes its Stage-2 encoder feature $h_i^\theta$ to the spatial decoder of Eq.~\eqref{eq:nb}, producing $\hat{A}_{ig}=\log(1+\mu_{ig})$. The retrieval branch uses the same feature $h_i^\theta$ to identify similar reference cells and obtains $\hat{R}_{ig}$ through the neighbor average of Eq.~\eqref{eq:knn}. Because the relative usefulness of morphology-based inference and reference-cell retrieval varies across genes, Stage 3 combines the two predictions using a gene-specific weighted average:
\begin{equation}
  \hat{y}_{ig} \;=\; \beta_g\,\hat{A}_{ig} \;+\; (1-\beta_g)\,\hat{R}_{ig},
  \qquad \beta_g \in [0,1],
  \label{eq:gate}
\end{equation}
The fusion weights are estimated separately for each tissue. For each gene $g$, we search the grid $\{0,0.025,\ldots,1\}$ on each training reference slide and select the weight that maximizes within-slide PCC. Reference-slide retrieval excludes that slide from its bank. We then average the slide-specific optima over same-tissue reference slides measuring $g$ and transfer the resulting $\beta_g$ to the test slide (Fig.~\ref{fig:gate}). Genes without a fitted weight or retrieval coverage, and tissues absent from training, use $\beta_g=1$.

The direct branch supplies $\log(1+\mu)$, while retrieval averages reference $\log(1+y)$ values, standardized per gene within each reference slide for cross-slide prediction. The gate is fitted on these branch outputs. The fused values are expression scores for comparing cells within each gene; the count mean $\mu$ is supplied by the direct head. Supporting-slide counts, weight stability, and held-out calibration are reported in Appendix~\ref{app:gate-calibration}.

\section{Results}
\label{sec:results}

\subsection{Experimental setup}
\label{sec:benchmark}

\textit{Data and model.}
We train VOICE at two data scales. VOICE-23M is trained on the full-scale training set of
$75$ human Xenium slides and $23{,}232{,}915$ cells from $15$ tissues, $51$ patient groups, and $24$ cohorts
(Table~\ref{tab:data-role}). VOICE-7M is trained on a 24-slide subset to perform ablation studies to determine the best model architecture and training strategies and to assess the impact of data scaling compared to the 23-million-cell data set.

We evaluate on three in-slide benchmarks (breast, lung, and skin) and ten cross-slide targets across seven tissues: two slides each from breast, lung, and ovary, and one each from kidney, pancreas, head and neck, and stomach. Head and neck and stomach are tissues absent from the training dataset. The in-slide cohort supports comparisons under the protocol used by prior methods, while the cross-slide cohort tests transfer across slides, patients, and tissues. All ten cross-slide targets are held out at the slide level (Appendix~\ref{app:provenance}). Slide identifiers and public sources are listed in Table~\ref{tab:test-provenance}.\\


\textit{Evaluation metrics.}
For every benchmark method, we compute Pearson Correlation between predicted and measured log-transformed expression separately for each gene across all cells of a slide. We report seven metrics, each computed as an unweighted average across genes in the corresponding evaluation set: all genes in the evaluated panel (\textsc{All}); the top 20, 50, and 100 highly variable genes (\textsc{H20}, \textsc{H50}, and \textsc{H100}), ranked by the variance of measured log-transformed expression; and the top 20, 50, and 100 spatially variable genes (\textsc{S20}, \textsc{S50}, and \textsc{S100}), ranked by Moran's (I) computed from measured log-transformed expression and cell coordinates. A prediction is considered unavailable when a gene is included in the evaluation panel but falls outside a method's output gene set or not measured in any reference slide for retrieval-based methods. Such genes are assigned a correlation of $0$ (Appendix~\ref{app:metrics}).\\

\textit{Baselines.}
Of the methods surveyed in \S\ref{sec:related}, we benchmark against those that (i) predict expression at single-cell resolution, and (ii) provide public code or pretrained weights, so that they can be run under a common protocol. Four methods meet both criteria: GHIST~\cite{ghist2025}, sCellST~\cite{scellst2026}, DeepSpot2Cell~\cite{deepspot2cell2025} and DeepSpot-M~\cite{nonchev2026deepspotm}. Reproduction details are given in Appendix~\ref{app:baselines}.\\

\textit{In-slide protocol.}
For every benchmark method, we divide each of the three in-slide evaluation slides into the same five contiguous vertical bands. In each fold, one band is held out for testing.  We reserve an inner-validation strip from the other four bands and use the remaining cells for training. All methods use the same training, inner-validation, and test datasets while retain their own architectures, objective functions, and preprocessing pipelines. Each band serves as the test set once, so every cell on the slide is evaluated exactly once.

For the VOICE direct branch, the pretrained image encoder is frozen, while the spatial decoder and count-prediction head are further finetuned. For the VOICE retrieval branch, reference cells are drawn from the same slide, excluding all cells in the held-out test band. For the VOICE fusion module, we find the per-gene weight on the training bands and transfer them to the test band.\\
\begin{table*}[!t]
\centering
\small
\setlength{\tabcolsep}{4pt}

\caption{
  \textbf{In-slide comparison under five-fold cross-validation.}
  The best reported result in each column is highlighted in bold.
}
\label{tab:single-slide-expanded}

\resizebox{\textwidth}{!}{%
\begin{tabular}{l ccccccc ccccccc ccccccc}
\toprule

& \multicolumn{7}{c}{Breast}
& \multicolumn{7}{c}{Lung}
& \multicolumn{7}{c}{Skin} \\

\cmidrule(lr){2-8}
\cmidrule(lr){9-15}
\cmidrule(lr){16-22}

Model
& All & S20 & S50 & S100 & H20 & H50 & H100
& All & S20 & S50 & S100 & H20 & H50 & H100
& All & S20 & S50 & S100 & H20 & H50 & H100 \\

\midrule

GHIST (reduced)
& 0.306 & 0.741 & 0.602 & 0.524 & 0.634 & 0.583 & 0.509
& 0.163 & 0.442 & 0.394 & 0.358 & 0.524 & 0.447 & 0.385
& 0.267 & 0.672 & 0.659 & 0.528 & 0.793 & 0.633 & 0.539 \\

sCellST
& 0.094 & 0.291 & 0.205 & 0.164 & 0.204 & 0.198 & 0.153
& 0.091 & 0.263 & 0.205 & 0.171 & 0.319 & 0.246 & 0.186
& 0.105 & 0.403 & 0.327 & 0.226 & 0.291 & 0.224 & 0.186 \\

DeepSpot2Cell
& 0.269 & 0.671 & 0.547 & 0.464 & 0.563 & 0.496 & 0.428
& 0.109 & 0.306 & 0.273 & 0.237 & 0.363 & 0.304 & 0.247
& 0.276 & 0.711 & 0.632 & 0.540 & 0.588 & 0.513 & 0.456 \\

DeepSpot-M (zero-shot)
& 0.069 & 0.183 & 0.136 & 0.122 & 0.148 & 0.129 & 0.108
& 0.044 & 0.118 & 0.107 & 0.094 & 0.103 & 0.094 & 0.079
& 0.135 & 0.392 & 0.338 & 0.248 & 0.163 & 0.221 & 0.207 \\

DeepSpot-M (fine-tuned)
& 0.421 & 0.842 & 0.729 & 0.630 & 0.761 & 0.687 & 0.600
& 0.235 & 0.519 & 0.470 & 0.440 & 0.599 & 0.506 & 0.447
& 0.432 & 0.906 & 0.842 & 0.738 & 0.856 & 0.743 & 0.665 \\

VOICE-7M
& 0.422 & 0.829 & 0.712 & 0.631 & 0.747 & 0.682 & 0.605
& 0.300 & 0.671 & 0.591 & 0.534 & 0.679 & 0.590 & 0.517
& 0.450 & 0.908 & 0.849 & 0.763 & 0.859 & 0.750 & 0.677 \\

\textbf{VOICE-23M}
& \textbf{0.464} & \textbf{0.857} & \textbf{0.758} & \textbf{0.678}
& \textbf{0.788} & \textbf{0.725} & \textbf{0.648}
& \textbf{0.323} & \textbf{0.716} & \textbf{0.634} & \textbf{0.567}
& \textbf{0.695} & \textbf{0.606} & \textbf{0.535}
& \textbf{0.478} & \textbf{0.925} & \textbf{0.868} & \textbf{0.788}
& \textbf{0.872} & \textbf{0.773} & \textbf{0.699} \\

\bottomrule
\end{tabular}%
}

\end{table*}

\textit{Cross-slide evaluation protocol.}
We evaluate the cross-slide targets without using them to train the model. Each method retains its own architecture, objective functions, and preprocessing pipeline. We evaluate all methods on each target slide's gene set.

For VOICE, we apply the pretrained model to each test slide without updating any model parameters, so the direct branch operates in a zero-shot manner. The retrieval branch uses only training slides from the same tissue to construct its reference bank.  A gene is retrieved only from reference slides on which that gene was measured.  Before pooling cells from different reference slides, we standardize each gene within each slide to remove slide-specific shifts and scales. We fit the per-gene fusion weights on the same tissue's training slides and transfer them to the test slide. For the unseen head-and-neck and stomach tissues, we use direct prediction alone ($\beta=1$). We also evaluate DeepSpot-M (pretrained), which is pretrained on the same training data as VOICE, and a matched VOICE variant that omits Stage~1, with both prediction branches and fusion weights recomputed using the ablated model.

Because different baselines predict different gene sets, we use three evaluation
panels. The \emph{full target panel} contains all transcript genes measured on the held-out Xenium test slide and compares DeepSpot-M with VOICE. The \emph{Xenium-shared panel} contains genes present in both the Xenium test slide and the Xenium training slides used by baselines, and compares GHIST, DeepSpot2Cell, DeepSpot-M, and VOICE. The \emph{Visium-shared panel} intersects the genes measured on the test slide with the \(1{,}000\) genes that have the highest log-transformed expression variance in the Visium training slides. These genes define both the training targets and the output gene set of the corresponding sCellST model. We use this panel to compare sCellST and GHIST with VOICE; DeepSpot-M is also evaluated both zero-shot and after fine-tuning on the same-tissue Xenium training slides. Head and neck and stomach, two tissues absent from the
training dataset are not compared in the xenium-shared and visium-shared pannel since GHIST and sCellST can't predict unseen tissue's expression.\\

\subsection{In-slide benchmark under five-fold cross-validation}
\label{sec:inslide}

\begin{figure*}[!t]
\centering
\includegraphics[
    width=0.70\textwidth,
    height=0.52\textheight,
    keepaspectratio
]{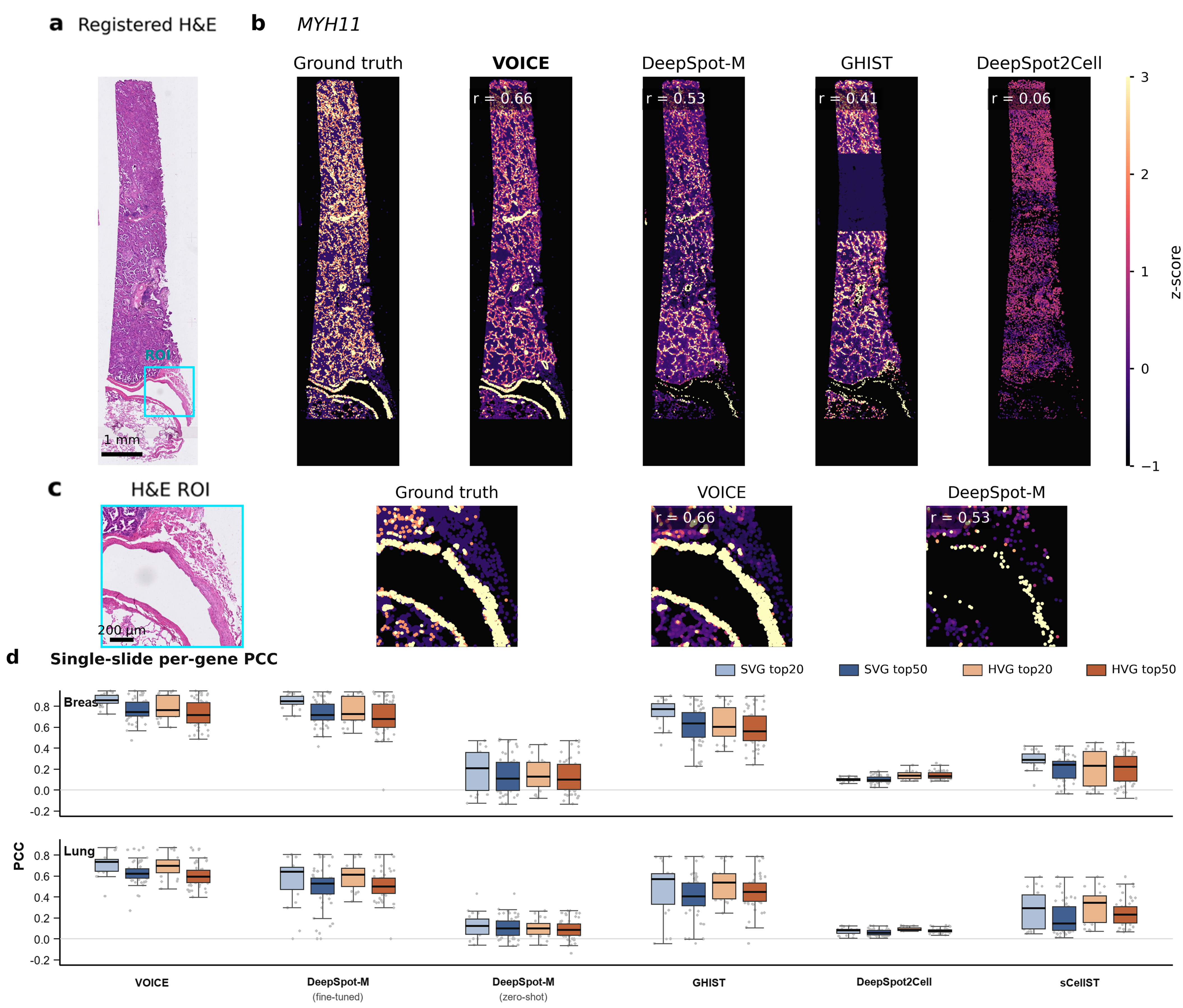}
\caption{\textbf{In-slide comparison.} \textbf{(a--c)} Spatial heatmap of specific gene expressions between ground truth and predicted values from VOICE-23M and baselines;
\textbf{(d)} per-gene PCC. Expression is shown as within-panel $z$-scores. The displayed GHIST predictions use the reduced variant.}
\Description{In-slide comparison on lung, with an H\&E image, measured and predicted MYH11 maps, an enlarged region, and per-gene PCC distributions on breast and lung.}
\label{fig:inslide-spatial}
\end{figure*}

\begin{table*}[!t]
\centering
\scriptsize
\setlength{\tabcolsep}{2.2pt}
\renewcommand{\arraystretch}{1.05}

\caption{
\textbf{Cross-slide generalization under baseline-compatible gene panels.}
Models are compared on the same genes within each shared panel.
Results are reported for spatially variable gene (S20 and S50) and highly
variable gene (H20 and H50) subsets. Complete results, including All, S100, and H100, are provided
in the appendix. GHIST (full) retains its auxiliary tasks. DeepSpot-M (pretrained) is pretrained on the same training data as VOICE. Slide identifiers are defined in Table~\ref{tab:test-provenance}. Shared-panel rows use identical panel-specific gene lists and rankings across models (Appendix~\ref{app:metrics}). Bold marks the best result within each panel.
}
\label{tab:cross-slide-summary}

\resizebox{\textwidth}{!}{%
\begin{tabular}{l
cccc
cccc
cccc
cccc
cccc}
\toprule

& \multicolumn{4}{c}{Breast-1}
& \multicolumn{4}{c}{Lung-1}
& \multicolumn{4}{c}{Kidney}
& \multicolumn{4}{c}{Ovary-1}
& \multicolumn{4}{c}{Pancreas} \\

\cmidrule(lr){2-5}
\cmidrule(lr){6-9}
\cmidrule(lr){10-13}
\cmidrule(lr){14-17}
\cmidrule(lr){18-21}

Model
& S20 & S50 & H20 & H50
& S20 & S50 & H20 & H50
& S20 & S50 & H20 & H50
& S20 & S50 & H20 & H50
& S20 & S50 & H20 & H50 \\

\midrule

\multicolumn{21}{l}{\textit{Full target panel}} \\

DeepSpot-M (zero-shot)
& 0.193 & 0.184 & 0.214 & 0.196 & 0.172 & 0.148 & 0.153 & 0.136 & 0.177 & 0.127 & 0.117 & 0.108 & 0.125 & 0.081 & 0.038 & 0.057 & 0.308 & 0.260 & 0.312 & 0.223 \\

DeepSpot-M (fine-tuned)
& 0.348 & 0.339 & 0.355 & 0.332 & 0.522 & 0.447 & 0.500 & 0.437 & 0.563 & 0.479 & \textbf{0.554} & 0.473 & 0.418 & 0.323 & 0.327 & 0.283 & 0.458 & 0.378 & 0.487 & 0.360 \\

DeepSpot-M (pretrained)
& 0.4257 & 0.4136 & 0.4523 & 0.4122 & 0.5882 & 0.5485 & 0.6661 & 0.5624 & 0.5507 & 0.4795 & 0.5483 & 0.4588 & 0.3719 & 0.3134 & 0.3075 & 0.2875 & 0.4451 & 0.4814 & 0.6180 & 0.5273 \\
VOICE-23M w/o Stage~1
& \textbf{0.5336} & \textbf{0.4856} & 0.5430 & 0.4991 & 0.6231 & \textbf{0.6017} & 0.6596 & 0.5850 & 0.5888 & 0.4887 & 0.5334 & \textbf{0.4878} & 0.4167 & 0.3551 & 0.3553 & 0.3186 & 0.5174 & \textbf{0.5264} & \textbf{0.6329} & \textbf{0.5557} \\

VOICE-7M
& 0.424 & 0.382 & 0.469 & 0.441 & 0.601 & 0.530 & 0.653 & 0.547 & 0.489 & 0.448 & 0.456 & 0.432 & \textbf{0.464} & \textbf{0.390} & \textbf{0.439} & \textbf{0.381} & 0.564 & 0.510 & 0.618 & 0.527 \\

VOICE-23M
& 0.515 & 0.480 & \textbf{0.559} & \textbf{0.519} & \textbf{0.673} & 0.599 & \textbf{0.684} & \textbf{0.589} & \textbf{0.602} & \textbf{0.507} & \textbf{0.554} & 0.477 & 0.433 & 0.379 & 0.420 & 0.360 & \textbf{0.566} & 0.517 & 0.592 & 0.519 \\

\midrule

\multicolumn{21}{l}{\textit{Xenium-shared gene panel}} \\

GHIST (reduced)
& 0.367 & 0.223 & 0.327 & 0.220 & 0.280 & 0.229 & 0.281 & 0.259 & 0.333 & 0.323 & 0.405 & 0.340 & 0.230 & 0.170 & 0.297 & 0.236 & 0.245 & 0.148 & 0.279 & 0.155 \\

GHIST (full)
& 0.3747 & 0.2301 & 0.3397 & 0.2266 & 0.2906 & 0.2139 & 0.2836 & 0.2331 & 0.2548 & 0.2549 & 0.2945 & 0.2654 & 0.0868 & 0.0660 & 0.1315 & 0.0859 & 0.2860 & 0.1650 & 0.2840 & 0.1603 \\

DeepSpot2Cell
& 0.231 & 0.149 & 0.215 & 0.144 & 0.312 & 0.221 & 0.280 & 0.220 & 0.165 & 0.145 & 0.187 & 0.149 & 0.267 & 0.170 & 0.247 & 0.196 & 0.181 & 0.114 & 0.162 & 0.104 \\

DeepSpot-M (zero-shot)
& 0.187 & 0.125 & 0.155 & 0.124 & 0.162 & 0.123 & 0.176 & 0.122 & 0.166 & 0.139 & 0.124 & 0.108 & 0.108 & 0.068 & 0.067 & 0.080 & 0.276 & 0.173 & 0.227 & 0.163 \\

DeepSpot-M (fine-tuned)
& 0.423 & 0.304 & 0.392 & 0.294 & 0.448 & 0.387 & 0.424 & 0.421 & 0.464 & 0.457 & 0.527 & 0.458 & 0.413 & 0.299 & 0.322 & 0.270 & 0.424 & 0.269 & 0.414 & 0.273 \\

VOICE-23M w/o Stage~1
& 0.4977 & 0.3864 & 0.4797 & 0.3697 & 0.5898 & 0.4899 & 0.5659 & \textbf{0.5098} & 0.5091 & \textbf{0.4870} & \textbf{0.5375} & \textbf{0.4761} & 0.3986 & 0.3218 & 0.3281 & 0.2854 & 0.4678 & \textbf{0.3301} & \textbf{0.4541} & \textbf{0.3308} \\

VOICE-7M
& 0.487 & 0.361 & 0.460 & 0.350 & 0.564 & 0.471 & 0.557 & 0.498 & 0.495 & 0.465 & 0.512 & 0.463 & \textbf{0.474} & \textbf{0.369} & \textbf{0.439} & \textbf{0.358} & 0.454 & 0.315 & 0.446 & 0.319 \\

VOICE-23M
& \textbf{0.511} & \textbf{0.395} & \textbf{0.494} & \textbf{0.381} & \textbf{0.598} & \textbf{0.492} & \textbf{0.573} & 0.508 & \textbf{0.517} & 0.482 & 0.525 & 0.465 & 0.449 & 0.359 & 0.413 & 0.344 & \textbf{0.469} & 0.328 & 0.454 & 0.330 \\

\midrule

\multicolumn{21}{l}{\textit{Visium-shared gene panel}} \\

GHIST (full)
& 0.1203 & 0.1119 & 0.1088 & 0.1146 & 0.0792 & 0.0779 & 0.1231 & 0.0819 & 0.2328 & 0.2213 & 0.2073 & 0.2100 & 0.0271 & 0.0233 & 0.0434 & 0.0243 & 0.0780 & 0.0427 & 0.0919 & 0.0419 \\

sCellST
& 0.170 & 0.161 & 0.144 & 0.153 & 0.083 & 0.066 & 0.094 & 0.075 & 0.100 & 0.053 & 0.077 & 0.044 & 0.109 & 0.070 & 0.100 & 0.077 & 0.113 & 0.090 & 0.117 & 0.097 \\

DeepSpot-M (zero-shot)
& 0.188 & 0.196 & 0.169 & 0.195 & 0.145 & 0.125 & 0.140 & 0.118 & 0.123 & 0.102 & 0.121 & 0.097 & 0.104 & 0.057 & 0.077 & 0.056 & 0.218 & 0.179 & 0.218 & 0.172 \\

DeepSpot-M (fine-tuned)
& 0.300 & 0.294 & 0.268 & 0.308 & 0.427 & 0.369 & 0.439 & 0.369 & 0.514 & 0.460 & 0.535 & 0.449 & 0.336 & 0.249 & 0.304 & 0.252 & 0.332 & 0.271 & 0.375 & 0.261 \\

VOICE-23M w/o Stage~1
& 0.4298 & 0.4001 & 0.4514 & 0.4153 & \textbf{0.5666} & \textbf{0.4818} & \textbf{0.5537} & \textbf{0.4828} & \textbf{0.5409} & \textbf{0.4950} & \textbf{0.5719} & \textbf{0.4900} & 0.3352 & 0.2623 & 0.3122 & 0.2740 & \textbf{0.5333} & \textbf{0.4517} & \textbf{0.5645} & \textbf{0.4482} \\

VOICE-7M
& 0.390 & 0.369 & 0.384 & 0.367 & 0.478 & 0.409 & 0.488 & 0.418 & 0.442 & 0.440 & 0.460 & 0.437 & \textbf{0.389} & \textbf{0.292} & \textbf{0.384} & \textbf{0.305} & 0.480 & 0.391 & 0.515 & 0.394 \\

VOICE-23M
& \textbf{0.481} & \textbf{0.430} & \textbf{0.475} & \textbf{0.438} & 0.558 & 0.479 & 0.544 & 0.480 & 0.534 & 0.481 & 0.566 & 0.474 & 0.361 & 0.286 & 0.354 & 0.301 & 0.522 & 0.440 & 0.548 & 0.438 \\

\bottomrule
\end{tabular}%
}
\end{table*}
We compare all methods on the three slides containing breast, lung, and skin under the in-slide protocol described in Section~\ref{sec:benchmark}. Table~\ref{tab:single-slide-expanded} compares task-specific models trained
within each slide and two scales of VOICE. From the result, VOICE-23M achieves the highest PCC in all 21 reported comparisons, with consistent gains across breast, lung, and skin tissues. Compared with VOICE-7M, it improves every metric by 0.013--0.047, demonstrating the benefit of increasing training data scale. For example, the S50 PCC increases from 0.712 to 0.758 in breast, from 0.591 to 0.634 in lung, and from 0.849 to 0.868 in skin. In detail, figure~\ref{fig:inslide-spatial} provides cell and gene level views of these
aggregate gains. On lung, VOICE predicts \textit{MYH11} with $r=0.66$, compared
with $0.53$ for DeepSpot-M, $0.41$ for reduced GHIST, and $0.06$ for DeepSpot2Cell. In the enlarged region, its map more closely follows the measured
boundary-localized pattern. Across breast and lung, VOICE also gives the
highest median per-gene PCC in every displayed SVG and HVG subset. 

The \textbf{ablation study} in Table~\ref{tab:inslide} demonstrates the effectiveness of the proposed fusion module. For VOICE-7M, fusion increases the macro-averaged all-gene PCC from 0.3810 to 0.3904 and outperforms the direct branch in all 21 slide-level metric comparisons, confirming that combining direct prediction with retrieval consistently improves predictive performance (Appendix~\ref{app:whygate}). The same table also ablates the two training stages, and Table~\ref{tab:delta} reports their separate contributions. Repeating the fusion analysis with VOICE-23M increases mean all-gene PCC from $0.3982$ to $0.4217$ on the three in-slide benchmarks. A matched 75-slide comparison also shows that Stage~1 increases direct-branch PCC from $0.3421$ to $0.4126$ (Appendix~\ref{app:23m-controls}).

\subsection{Cross-slide generalization to unseen slides and tissues}
\label{sec:xsgen}

\begin{table*}[!t]
\centering\footnotesize
\renewcommand{\arraystretch}{1.08}
\setlength{\tabcolsep}{4pt}
\caption{\textbf{Generalization to unseen tissues.} Head and neck and stomach are absent from training. Both VOICE variants use direct prediction ($\beta=1$); DeepSpot-M (pretrained) uses the same training data as VOICE. Values are PCC on the full target panel. S/H denote the top SVGs/HVGs; bold marks the best result per tissue and metric.}
\label{tab:unseen-tissue}
\begin{tabular*}{\textwidth}{@{\extracolsep{\fill}}llrrrrrrr}
\toprule
Tissue & Model & All & S20 & S50 & S100 & H20 & H50 & H100 \\
\midrule
Head and neck & DeepSpot-M (fine-tuned) & 0.0474 & 0.2806 & 0.2073 & 0.1310 & 0.1135 & 0.1255 & 0.1171 \\
 & DeepSpot-M (pretrained) & \textbf{0.2339} & \textbf{0.6252} & \textbf{0.5460} & \textbf{0.4774} & 0.5033 & 0.4523 & \textbf{0.4402} \\
 & VOICE-23M w/o Stage~1 & 0.2226 & 0.5924 & 0.5262 & 0.4581 & \textbf{0.5192} & \textbf{0.4610} & 0.4311 \\
 & VOICE-23M & 0.2215 & 0.5727 & 0.5039 & 0.4482 & 0.4821 & 0.4432 & 0.4222 \\
\midrule
Stomach & DeepSpot-M (fine-tuned) & 0.0433 & 0.1568 & 0.1341 & 0.1051 & 0.1269 & 0.0908 & 0.0858 \\
 & DeepSpot-M (pretrained) & 0.1407 & \textbf{0.4271} & \textbf{0.3503} & 0.2951 & 0.4229 & 0.3197 & 0.2702 \\
 & VOICE-23M w/o Stage~1 & \textbf{0.1555} & 0.4097 & 0.3484 & \textbf{0.3049} & \textbf{0.4600} & \textbf{0.3565} & \textbf{0.2987} \\
 & VOICE-23M & 0.1516 & 0.3817 & 0.3285 & 0.2930 & 0.4306 & 0.3306 & 0.2811 \\
\bottomrule
\end{tabular*}
\end{table*}

On the full target panel, VOICE variants achieve the highest PCC in all seven metrics on seven of the eight seen-tissue targets, including a tie on Kidney H20 at $0.554$ (Tables~\ref{tab:cross-slide-summary} and~\ref{tab:cross-slide-target}). On Ovary-2, VOICE-23M w/o Stage~1 leads All, H50, and H100, while DeepSpot-M (pretrained) leads the three SVG subsets and H20. Across all ten cross-slide targets, VOICE-23M and its no-Stage-1 variant reach mean all-gene PCCs of $0.2121$ and $0.2089$, respectively, compared with $0.1915$ for DeepSpot-M (pretrained), trained on the same corpus (Tables~\ref{tab:cross-slide-summary} ,Table~\ref{tab:unseen-tissue} and ~\ref{tab:cross-slide-target}).

On the Xenium-shared panel, VOICE variants achieve the highest PCC in all seven metrics on every evaluated slide (Tables~\ref{tab:cross-slide-summary} and~\ref{tab:cross-slide-xenium}). This includes Kidney H20, where the no-Stage-1 variant reaches $0.5375$, exceeding fine-tuned DeepSpot-M at $0.527$. VOICE-23M also outperforms full GHIST on all seven metrics on each of the eight targets, with all-gene PCC gains of $0.0716$--$0.4385$. This advantage holds with GHIST's cell-type and neighborhood training objectives included.

On the Visium-shared panel, VOICE variants lead all $56$ comparisons across eight slides and seven metrics, ahead of sCellST, full GHIST, and DeepSpot-M (Tables~\ref{tab:cross-slide-summary} and~\ref{tab:cross-slide-visium}). VOICE-23M reaches all-gene PCC $0.5438$ on Breast-2, compared with $0.3587$ for fine-tuned DeepSpot-M. The no-Stage-1 variant reaches $0.1324$ on Lung-2 and $0.2267$ on Ovary-2, versus $0.0852$ and $0.1640$, respectively, for the same baseline. These gains extend VOICE's advantage to the gene sets used by spot-supervised methods.

On head and neck and stomach, two tissues absent from the training dataset, both VOICE-23M variants use direct prediction alone ($\beta=1$) without target-specific fine-tuning. Both outperform fine-tuned DeepSpot-M across all seven metrics on both tissues: all-gene PCCs are $0.2215$--$0.2226$ versus $0.0474$ for head and neck and $0.1516$--$0.1555$ versus $0.0433$ for stomach (Table~\ref{tab:unseen-tissue}). Against DeepSpot-M (pretrained), which uses the same training data, the no-Stage-1 variant leads H20 and H50 on head and neck, and All, S100, and all three HVG subsets on stomach. These results show that VOICE's learned morphology--expression relationship transfers to tissue types absent from training.

On the full target panel, scaling from VOICE-7M to VOICE-23M raises the all-gene score on the five slides
with both model scales evaluated. Appendix~\ref{app:ovary-panel} relates the
ovarian ranked-subset results to differences in assay generation between the
training panels at the two model scales.

\begin{figure*}[t]
\centering
\includegraphics[
    width=0.76\textwidth,
    height=0.52\textheight,
    keepaspectratio
]{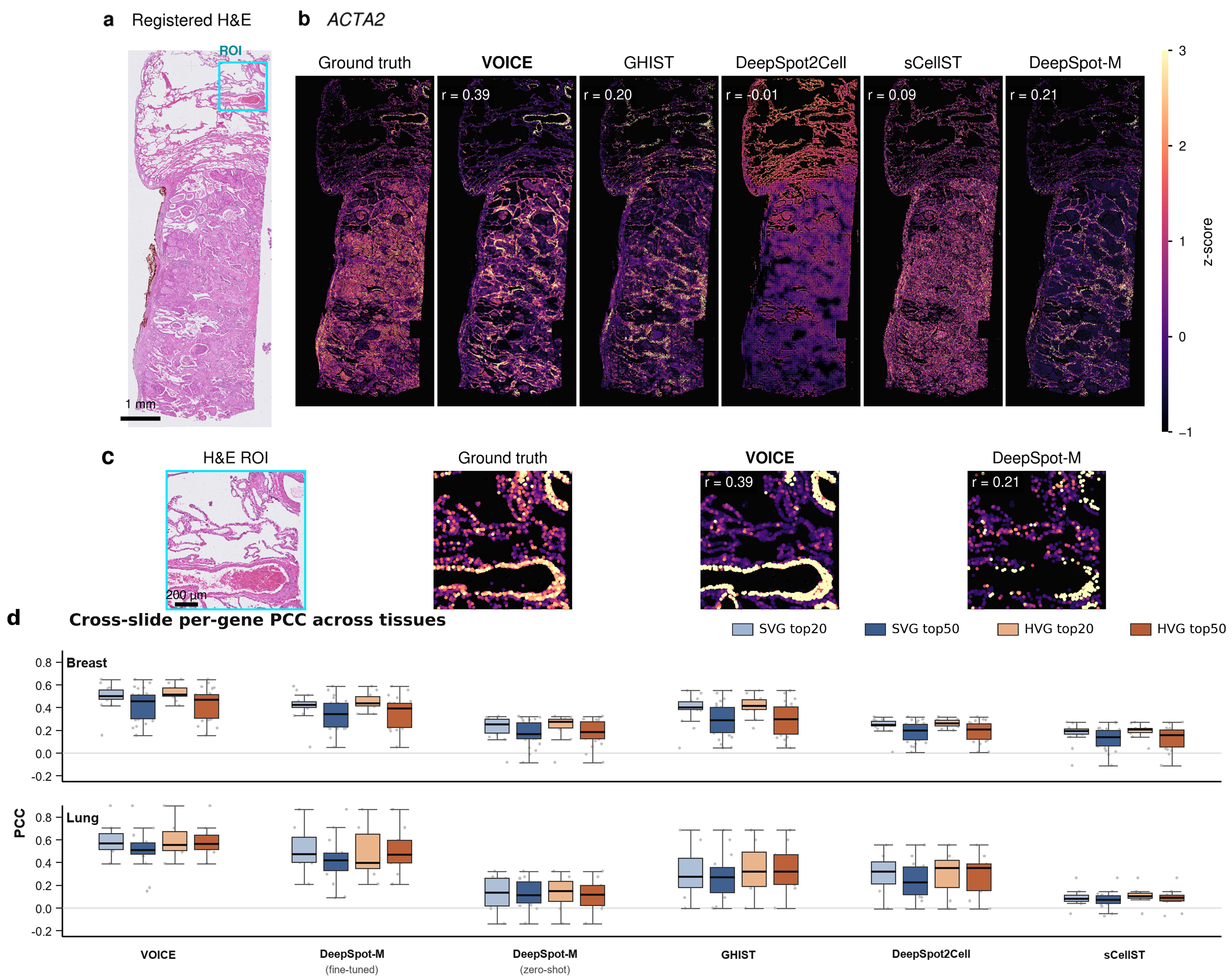}
\caption{\textbf{Cross-slide comparison.} \textbf{(a--c)} Spatial heatmaps on Lung-1 comparing ground truth with predictions from VOICE-23M and baselines;
\textbf{(d)} per-gene PCC on Breast-1 and Lung-1. Expression is shown as within-panel $z$-scores. The displayed GHIST predictions use the reduced variant.}
\Description{Cross-slide comparison on lung, with an H\&E image, measured and predicted ACTA2 maps, an enlarged region, and per-gene PCC distributions on breast and lung.}
\label{fig:crossslide-spatial}
\end{figure*}

Figure~\ref{fig:crossslide-spatial} shows the same advantage at cellular
resolution. On Lung-1, VOICE predicts \textit{ACTA2} with
$r=0.39$, compared with $0.21$ for DeepSpot-M, $0.20$ for reduced GHIST, $0.09$ for
sCellST, and $-0.01$ for DeepSpot2Cell. VOICE also gives the highest
median per-gene PCC in every displayed SVG and HVG subset across Breast-1 and
Lung-1.\\

\textit{Prediction of Genes Measured Only in Other Tissues}
Within tissues represented in training, we further assess whether VOICE can predict genes that were measured only in other tissues. To evaluate this setting, we analyze the direct branch of VOICE-23M on all eight cross-slide targets from these tissues. The retrieval branch is excluded because no same-tissue reference expression is available for these genes. Training support is defined by the $75$ slides used for Stage~2. For each target tissue, we divide training-supported genes into two groups. A \emph{same-tissue gene} is measured in at least one training slide from that tissue. An \emph{other-tissue-only gene} is not measured in any training slide from the target tissue but is measured in at least one training slide from another tissue. Within each group, SVGs and HVGs are ranked separately using expression measured on the target slide.

Table~\ref{tab:cross-tissue-genes} (notation in Appendix~\ref{app:crosstissue-notation}) shows positive correlations for other-tissue-only genes on all six slides containing such genes, with all-gene PCCs of $0.1317$--$0.3526$. On Breast-2, the $28$ other-tissue-only genes reach $0.3005$. The slide-macro all-gene PCCs are $0.2270$ for same-tissue genes and $0.2528$ for other-tissue-only genes, averaging eight and six nonempty groups, respectively. Weighting slides by the number of evaluated genes gives $0.1499$ and $0.1429$; other-tissue-only SVG/HVG subsets retain PCCs of $0.3486$--$0.4643$. The groups differ in gene identity and expression characteristics. These results show that VOICE transfers morphological knowledge to genes never measured in the target tissue's Stage~2 training panels.

\begin{table*}[tp]
\centering\small
\setlength{\tabcolsep}{4pt}
\caption{\textbf{VOICE-23M cross-tissue gene transfer across eight slides.} Values are direct-branch PCC for genes measured in same-tissue Stage-2 training slides or only in other tissues. SVG/HVG subsets are ranked within each group. Empty groups are marked by dashes. Macro averages nonempty slide groups; gene-weighted summaries weight each slide by the number of genes evaluated for that metric.}
\label{tab:cross-tissue-genes}
\resizebox{\textwidth}{!}{%
\begin{tabular}{l cccccccc cccccccc cccccccc}
\toprule
 & \multicolumn{8}{c}{Breast-1} & \multicolumn{8}{c}{Lung-1} & \multicolumn{8}{c}{Kidney} \\
\cmidrule(lr){2-9}\cmidrule(lr){10-17}\cmidrule(lr){18-25}
Training evidence & $n$ & All & S20 & S50 & S100 & H20 & H50 & H100 & $n$ & All & S20 & S50 & S100 & H20 & H50 & H100 & $n$ & All & S20 & S50 & S100 & H20 & H50 & H100 \\
\midrule
Same tissue & 331 & 0.1834 & 0.3778 & 0.3691 & 0.3278 & 0.4042 & 0.3730 & 0.3217 & 419 & 0.2816 & 0.6368 & 0.5647 & 0.4949 & 0.6023 & 0.5594 & 0.5058 & 377 & 0.2311 & 0.5078 & 0.4705 & 0.4278 & 0.5014 & 0.4668 & 0.4325 \\
Other tissues only & 4,755 & 0.1317 & 0.5330 & 0.4872 & 0.4498 & 0.5314 & 0.4930 & 0.4542 & 61 & 0.3068 & 0.4935 & 0.3557 & 0.3068 & 0.5117 & 0.3538 & 0.3068 & 26 & 0.3526 & 0.4340 & 0.3526 & 0.3526 & 0.4151 & 0.3526 & 0.3526 \\
$\Delta$ (other $-$ same) & -- & $-0.0517$ & $+0.1552$ & $+0.1181$ & $+0.1220$ & $+0.1272$ & $+0.1200$ & $+0.1325$ & -- & $+0.0252$ & $-0.1433$ & $-0.2090$ & $-0.1881$ & $-0.0906$ & $-0.2056$ & $-0.1990$ & -- & $+0.1215$ & $-0.0738$ & $-0.1179$ & $-0.0752$ & $-0.0863$ & $-0.1142$ & $-0.0799$ \\
\bottomrule
\end{tabular}%
}
\par\vspace{-0.5pt}
\resizebox{\textwidth}{!}{%
\begin{tabular}{l cccccccc cccccccc cccccccc}
\toprule
 & \multicolumn{8}{c}{Ovary-1} & \multicolumn{8}{c}{Pancreas} & \multicolumn{8}{c}{Breast-2} \\
\cmidrule(lr){2-9}\cmidrule(lr){10-17}\cmidrule(lr){18-25}
Training evidence & $n$ & All & S20 & S50 & S100 & H20 & H50 & H100 & $n$ & All & S20 & S50 & S100 & H20 & H50 & H100 & $n$ & All & S20 & S50 & S100 & H20 & H50 & H100 \\
\midrule
Same tissue & 404 & 0.1371 & 0.4521 & 0.3630 & 0.2814 & 0.4071 & 0.3260 & 0.2857 & 102 & 0.2344 & 0.4962 & 0.3430 & 0.2371 & 0.4662 & 0.3475 & 0.2385 & 283 & 0.4882 & 0.8570 & 0.7696 & 0.6987 & 0.8605 & 0.7616 & 0.6836 \\
Other tissues only & 76 & 0.1690 & 0.3157 & 0.2221 & 0.1690 & 0.3360 & 0.2215 & 0.1690 & 278 & 0.2564 & 0.5454 & 0.4747 & 0.4218 & 0.5856 & 0.4951 & 0.4285 & 28 & 0.3005 & 0.4008 & 0.3005 & 0.3005 & 0.4058 & 0.3005 & 0.3005 \\
$\Delta$ (other $-$ same) & -- & $+0.0319$ & $-0.1364$ & $-0.1409$ & $-0.1124$ & $-0.0711$ & $-0.1045$ & $-0.1167$ & -- & $+0.0220$ & $+0.0492$ & $+0.1317$ & $+0.1847$ & $+0.1194$ & $+0.1476$ & $+0.1900$ & -- & $-0.1877$ & $-0.4562$ & $-0.4691$ & $-0.3982$ & $-0.4547$ & $-0.4612$ & $-0.3832$ \\
\bottomrule
\end{tabular}%
}
\par\vspace{-0.5pt}
\resizebox{\textwidth}{!}{%
\begin{tabular}{l cccccccc cccccccc cccccccc}
\toprule
 & \multicolumn{8}{c}{Lung-2} & \multicolumn{8}{c}{Ovary-2} & \multicolumn{8}{c}{Macro} \\
\cmidrule(lr){2-9}\cmidrule(lr){10-17}\cmidrule(lr){18-25}
Training evidence & $n$ & All & S20 & S50 & S100 & H20 & H50 & H100 & $n$ & All & S20 & S50 & S100 & H20 & H50 & H100 & $n$ & All & S20 & S50 & S100 & H20 & H50 & H100 \\
\midrule
Same tissue & 5,001 & 0.0843 & 0.4458 & 0.3801 & 0.3549 & 0.4977 & 0.4426 & 0.4004 & 5,001 & 0.1763 & 0.4373 & 0.4567 & 0.4471 & 0.5365 & 0.5370 & 0.5247 & -- & 0.2270 & 0.5263 & 0.4646 & 0.4087 & 0.5345 & 0.4767 & 0.4241 \\
Other tissues only & 0 & -- & -- & -- & -- & -- & -- & -- & 0 & -- & -- & -- & -- & -- & -- & -- & -- & 0.2528 & 0.4537 & 0.3655 & 0.3334 & 0.4643 & 0.3694 & 0.3353 \\
$\Delta$ (other $-$ same) & -- & -- & -- & -- & -- & -- & -- & -- & -- & -- & -- & -- & -- & -- & -- & -- & -- & -- & -- & -- & -- & -- & -- & -- \\
\bottomrule
\end{tabular}%
}
\par\vspace{3pt}
\begin{tabular}{lrrrrrrr}
\toprule
Gene-weighted summary & All & S20 & S50 & S100 & H20 & H50 & H100 \\
\midrule
Same tissue & 0.1499 & 0.5263 & 0.4646 & 0.4087 & 0.5345 & 0.4767 & 0.4241 \\
Other tissues only & 0.1429 & 0.4537 & 0.3723 & 0.3486 & 0.4643 & 0.3770 & 0.3514 \\
\bottomrule
\end{tabular}
\end{table*}

\subsection{Recovery of spatial programs and cell states}
\label{sec:downstream}
To assess how VOICE's expression prediction supports biological analysis, we examined spatial domains and extracellular-matrix (ECM) program analysis on the held-out pancreas slide, followed by proliferation and subtype expression differences on Breast-2. 
We first clustered measured expression with spatial neighborhood information to define spatial domains. Then, based on marker gene expression and pathway
scores, part of the regions were annotated as ECM-rich stromal (Appendix~\ref{app:downstream}). Using these expression-defined domains as a reference, we then clustered VOICE-predicted expression. 

\begin{figure*}[t]
\centering
\includegraphics[width=\textwidth]{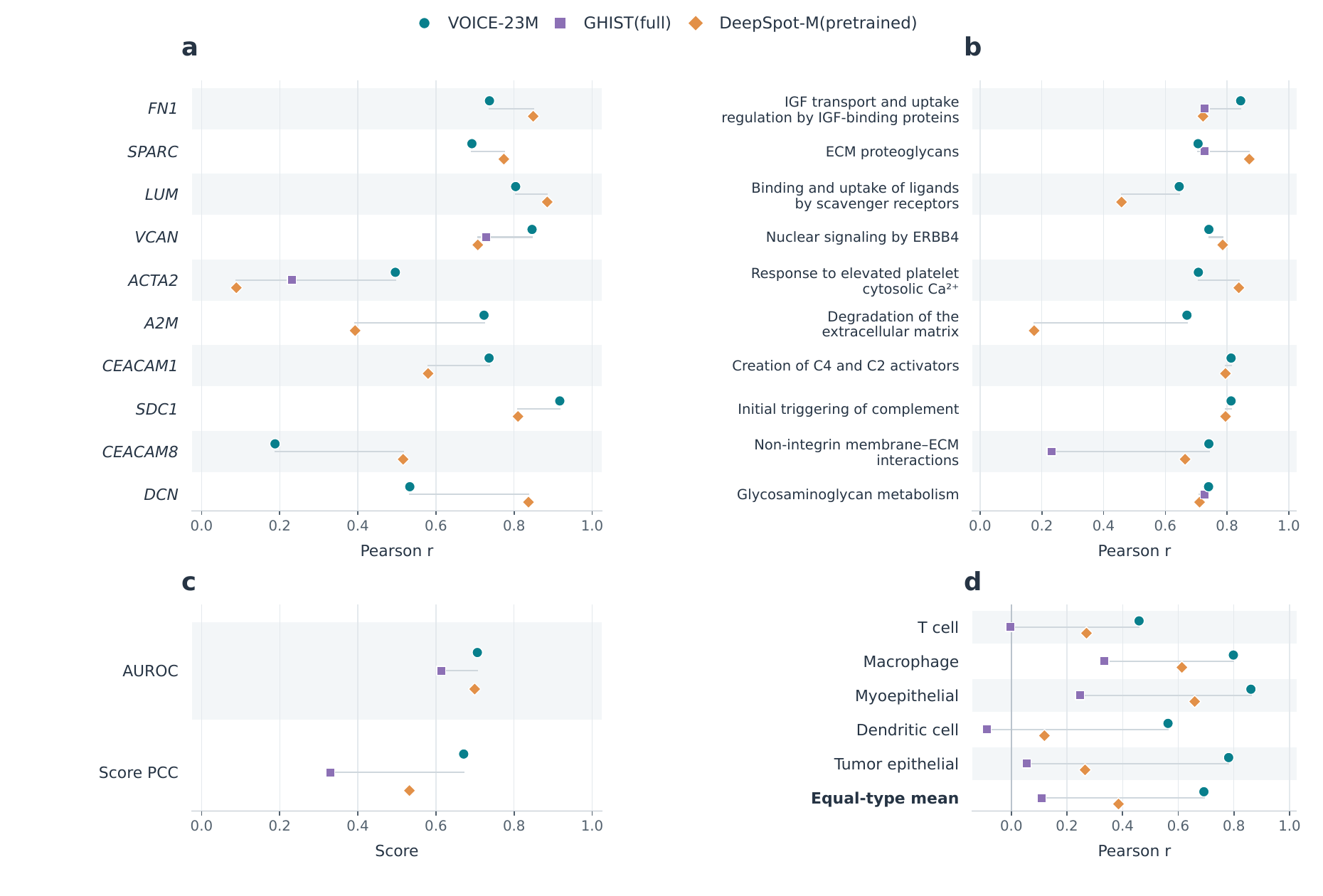}
\caption{\textbf{Biological analyses from predicted expression.} Comparisons use GHIST (full) and DeepSpot-M (pretrained). \textbf{a,b}, Correlation of ECM-gene expression and Reactome pathway scores between measured and projected ECM-rich regions across paired $500\,\mu$m blocks. Each model uses its available genes and cells; missing markers indicate unavailable or undefined correlations. \textbf{c}, Proliferation AUROC and score PCC in Breast-2 tumor epithelial cells. \textbf{d}, Correlation of measured and predicted subtype expression differences, averaged over subtype pairs within each cell type; the final row weights the five cell types equally.}
\Description{Four dot plots compare VOICE, GHIST (full), and DeepSpot-M (pretrained) for ECM genes, Reactome pathways, proliferation, and cell-subtype expression differences.}
\label{fig:downstream}
\end{figure*}

On pancreas, VOICE retains $80.0\%$ of spatial-domain labels. Within a predicted ECM-rich stromal domain, measured and predicted expression share $230$ differentially expressed genes ($q<0.05$), including $20$ ECM genes. Across paired $500\,\mu$m spatial blocks, VOICE achieves the highest correlation among available results for five of ten ECM genes and seven of ten Reactome pathways (Fig.~\ref{fig:downstream}a,b). Examples include \textit{SDC1} ($r=0.917$ versus $0.810$ for pretrained DeepSpot-M), \textit{ACTA2} ($0.496$ versus $0.231$ for full GHIST and $0.089$ for pretrained DeepSpot-M), and ECM degradation ($0.670$ versus $0.175$ for pretrained DeepSpot-M). These results show that VOICE's predictions preserve spatial organization and coordinated ECM activity, supporting analyses of stromal programs from H\&E.

On Breast-2, a six-gene proliferation score distinguishes proliferative from invasive tumor epithelial cells with AUROC $0.706$, compared with $0.613$ for full GHIST and $0.699$ for pretrained DeepSpot-M. Its correlation with the measured score is $0.671$, versus $0.329$ and $0.532$, respectively (Fig.~\ref{fig:downstream}c). VOICE also gives the highest correlation of subtype expression differences in each of five broad cell types, with an equal-type mean of $0.692$ versus $0.109$ for full GHIST and $0.385$ for pretrained DeepSpot-M (Fig.~\ref{fig:downstream}d). Together, these results connect the predicted expression patterns to spatial molecular programs and variation within cell types.

\section{Conclusion and Discussion}
We presented VOICE, a multimodal spatial foundation model for predicting single-cell gene expression from H\&E images. VOICE aligns cellular morphology with transcriptomic representations and combines two complementary prediction strategies: a direct branch that infers expression from morphology and a retrieval branch that transfers expression from similar reference cells. A per-gene fusion mechanism balances the two branches, reflecting that genes differ in how strongly their expression is associated with visible tissue morphology. VOICE improves within-slide prediction and cross-slide performance, including comparisons under matched training data. Its predictions also recover spatial domains, pathway activity, proliferation, and subtype expression differences. Increasing the training scale from 7 million to 23 million cells further improved performance, supporting the value of large-scale and heterogeneous cell-level supervision. A \textbf{key advantage} of VOICE is that it learns from heterogeneous gene panels through a shared gene vocabulary and panel-specific loss masking, enabling prediction of genes absent from same-tissue training data but observed in other tissues or panels.

\section{Limitations \& Ethical Considerations}
VOICE is trained on Xenium data. Its transfer to CosMx is evaluated in Appendix~\ref{app:cosmx-zeroshot}; broader validation across platforms and clinical cohorts remains a direction for future work. The present analyses evaluate expression patterns and their biological utility, while absolute count calibration remains to be assessed.

This study uses publicly available human H\&E and Xenium datasets obtained from published studies and databases. VOICE is intended for research purposes only and should not be used for clinical diagnosis or treatment decisions without further validation.

\section{Code Availability}
\label{sec:code}
The code for training and performing end-to-end prediction of single-cell gene expression from H\&E images is publicly available at \url{https://github.com/Luosanj/VOICE_HE_ST_Pred.git} and model weights are available at \url{https://huggingface.co/zchx/VOICE-23M}.

\section{Generative AI Usage}

The authors used generative AI tools for language editing and manuscript polishing. The authors take full responsibility for the study design, experiments, analyses of all reported results.

\section{Acknowledgment}
This work was supported in part by the National Institutes of Health (NIH) under awards U24DK138515 and OT2OD038003.

\FloatBarrier
\bibliographystyle{ACM-Reference-Format}
\bibliography{references}

@article{omiclip2025,
  author  = {Chen, Weiqing and Zhang, Pengzhi and Tran, Tu N. and Xiao, Yiwei and Li, Shengyu and Shah, Vrutant V. and Cheng, Hao and Brannan, Kristopher W. and Youker, Keith and Lai, Li and Fang, Longhou and Yang, Yu and Le, Nhat-Tu and Abe, Jun-ichi and Chen, Shu-Hsia and Ma, Qin and Chen, Ken and Song, Qianqian and Cooke, John P. and Wang, Guangyu},
  title   = {A visual--omics foundation model to bridge histopathology with spatial transcriptomics},
  journal = {Nature Methods},
  volume  = {22},
  pages   = {1568--1582},
  year    = {2025},
  doi     = {10.1038/s41592-025-02707-1}
}

@article{ghist2025,
  author  = {Fu, Xiaohang and Cao, Yue and Bian, Beilei and Wang, Chuhan and Graham, Dinny and Pathmanathan, Nirmala and Patrick, Ellis and Kim, Jinman and Yang, Jean Yee Hwa},
  title   = {Spatial gene expression at single-cell resolution from histology using deep learning with GHIST},
  journal = {Nature Methods},
  volume  = {22},
  pages   = {1900--1910},
  year    = {2025},
  doi     = {10.1038/s41592-025-02795-z}
}

@article{storm2026,
  author  = {Xiang, Jinxi and Hou, Siyu and Li, Yuchen and Quinton, Ryan and Zhang, Xiaoming and Eweje, Feyisope and Luo, Xiangde and Chen, Yijiang and Li, Zhe and Bergstrom, Colin and Kim, Ted and Willens, Sierra and Olguin, Francesca Maria and Abikenari, Matthew and Heider, Andrew and Rajaram, Sanjeeth and Neal, Joel and Diehn, Maximilian and Zhou, Xiang and Li, Ruijiang},
  title   = {A Multimodal Foundation Model of Spatial Transcriptomics and Histology for Biological Discovery and Clinical Prediction},
  journal = {arXiv preprint arXiv:2604.03630},
  year    = {2026},
  doi     = {10.48550/arXiv.2604.03630}
}

@article{scellst2026,
  author  = {Chadoutaud, Lo{\"i}c and Lerousseau, Marvin and Herrero-Saboya, Daniel and Ostermaier, Julian and Fontugne, Jacqueline and Barillot, Emmanuel and Walter, Thomas},
  title   = {sCellST predicts single-cell gene expression from H\&E images},
  journal = {Nature Communications},
  volume  = {17},
  pages   = {1194},
  year    = {2026},
  doi     = {10.1038/s41467-025-67965-1}
}

@inproceedings{chen2021mocov3,
  author    = {Chen, Xinlei and Xie, Saining and He, Kaiming},
  title     = {An Empirical Study of Training Self-Supervised Vision Transformers},
  booktitle = {Proceedings of the IEEE/CVF International Conference on Computer Vision (ICCV)},
  pages     = {9620--9629},
  year      = {2021},
  doi       = {10.1109/ICCV48922.2021.00950}
}

@article{deepspot2cell2025,
  title={DeepSpot2Cell: Predicting Virtual Single-Cell Spatial Transcriptomics from H\&E images using Spot-Level Supervision},
  author={Nonchev, Kalin and Manaiev, Glib and Koelzer, Viktor H and R{\"a}tsch, Gunnar},
  journal={bioRxiv},
  pages={2025--09},
  year={2025},
  publisher={Cold Spring Harbor Laboratory}
}

@article{nonchev2026deepspotm,
  author  = {Kalin Nonchev and Sebastian Dawo and Karina Silina and Viktor H. Koelzer and Gunnar R{\"a}tsch},
  title   = {DeepSpot-M: A Multimodal Foundation Model for Transcriptome-Wide Virtual Spatial Transcriptomics from Histology},
  journal = {medRxiv},
  year    = {2026},
  doi     = {10.64898/2026.06.19.26356060},
  url     = {https://doi.org/10.64898/2026.06.19.26356060}
}

@article{he2022high,
  title={High-plex imaging of RNA and proteins at subcellular resolution in fixed tissue by spatial molecular imaging},
  author={He, Shanshan and Bhatt, Ruchir and Brown, Carl and Brown, Emily A and Buhr, Derek L and Chantranuvatana, Kan and Danaher, Patrick and Dunaway, Dwayne and Garrison, Ryan G and Geiss, Gary and others},
  journal={Nature biotechnology},
  volume={40},
  number={12},
  pages={1794--1806},
  year={2022},
  publisher={Nature Publishing Group US New York}
}

@article{janesick2023high,
  title={High resolution mapping of the tumor microenvironment using integrated single-cell, spatial and in situ analysis},
  author={Janesick, Amanda and Shelansky, Robert and Gottscho, Andrew D and Wagner, Florian and Williams, Stephen R and Rouault, Morgane and Beliakoff, Ghezal and Morrison, Carolyn A and Oliveira, Michelli F and Sicherman, Jordan T and others},
  journal={Nature communications},
  volume={14},
  number={1},
  pages={8353},
  year={2023},
  publisher={Nature Publishing Group UK London}
}

@article{chen2024towards,
  title={Towards a general-purpose foundation model for computational pathology},
  author={Chen, Richard J and Ding, Tong and Lu, Ming Y and Williamson, Drew FK and Jaume, Guillaume and Song, Andrew H and Chen, Bowen and Zhang, Andrew and Shao, Daniel and Shaban, Muhammad and others},
  journal={Nature medicine},
  volume={30},
  number={3},
  pages={850--862},
  year={2024},
  publisher={Nature Publishing Group US New York}
}

@article{zhang2024istar,
  author  = {Zhang, Daiwei and Schroeder, Amelia and Yan, Hanying and Yang, Haochen and Hu, Jian and Lee, Michelle Y. Y. and Cho, Kyung S. and Susztak, Katalin and Xu, George X. and Feldman, Michael D. and Lee, Edward B. and Furth, Emma E. and Wang, Linghua and Li, Mingyao},
  title   = {Inferring super-resolution tissue architecture by integrating spatial transcriptomics with histology},
  journal = {Nature Biotechnology},
  volume  = {42},
  pages   = {1372--1377},
  year    = {2024},
  doi     = {10.1038/s41587-023-02019-9}
}

@inproceedings{deng2009imagenet,
  title={Imagenet: A large-scale hierarchical image database},
  author={Deng, Jia and Dong, Wei and Socher, Richard and Li, Li-Jia and Li, Kai and Fei-Fei, Li},
  booktitle={2009 IEEE conference on computer vision and pattern recognition},
  pages={248--255},
  year={2009},
  organization={Ieee}
}

@inproceedings{zhou2023uni,
  title={Uni-mol: A universal 3d molecular representation learning framework},
  author={Zhou, Gengmo and Gao, Zhifeng and Ding, Qiankun and Zheng, Hang and Xu, Hongteng and Wei, Zhewei and Zhang, Linfeng and Ke, Guolin},
  booktitle={The eleventh international conference on learning representations},
  year={2023}
}

@article{wang2022transformer,
  title={Transformer-based unsupervised contrastive learning for histopathological image classification},
  author={Wang, Xiyue and Yang, Sen and Zhang, Jun and Wang, Minghui and Zhang, Jing and Yang, Wei and Huang, Junzhou and Han, Xiao},
  journal={Medical image analysis},
  volume={81},
  pages={102559},
  year={2022},
  publisher={Elsevier}
}

@article{cui2024scgpt,
  title={scGPT: toward building a foundation model for single-cell multi-omics using generative AI},
  author={Cui, Haotian and Wang, Chloe and Maan, Hassaan and Pang, Kuan and Luo, Fengning and Duan, Nan and Wang, Bo},
  journal={Nature methods},
  volume={21},
  number={8},
  pages={1470--1480},
  year={2024},
  publisher={Nature Publishing Group US New York}
}

@article{theodoris2023transfer,
  title={Transfer learning enables predictions in network biology},
  author={Theodoris, Christina V and Xiao, Ling and Chopra, Anant and Chaffin, Mark D and Al Sayed, Zeina R and Hill, Matthew C and Mantineo, Helene and Brydon, Elizabeth M and Zeng, Zexian and Liu, X Shirley and others},
  journal={Nature},
  volume={618},
  number={7965},
  pages={616--624},
  year={2023},
  publisher={Nature Publishing Group UK London}
}

@article{hao2024large,
  title={Large-scale foundation model on single-cell transcriptomics},
  author={Hao, Minsheng and Gong, Jing and Zeng, Xin and Liu, Chiming and Guo, Yucheng and Cheng, Xingyi and Wang, Taifeng and Ma, Jianzhu and Zhang, Xuegong and Song, Le},
  journal={Nature methods},
  volume={21},
  number={8},
  pages={1481--1491},
  year={2024},
  publisher={Nature Publishing Group US New York}
}

@article{dong2026dynamic,
  title={Dynamic interaction-aware and causality-disentangled framework for multimodal sentiment analysis},
  author={Dong, Guangyuan and Hong, Ziwei and Liu, Shenghao and Wu, Chenyu and Fang, Yuanyuan and Li, Zihao and Zhang, Xudong and Liu, Bingchen and Zhang, Yuchen and Ding, Haitao and others},
  journal={arXiv preprint arXiv:2605.30994},
  year={2026}
}

@inproceedings{radford2021learning,
  title={Learning transferable visual models from natural language supervision},
  author={Radford, Alec and Kim, Jong Wook and Hallacy, Chris and Ramesh, Aditya and Goh, Gabriel and Agarwal, Sandhini and Sastry, Girish and Askell, Amanda and Mishkin, Pamela and Clark, Jack and others},
  booktitle={International conference on machine learning},
  pages={8748--8763},
  year={2021},
  organization={PmLR}
}

@inproceedings{rombach2022high,
  title={High-resolution image synthesis with latent diffusion models},
  author={Rombach, Robin and Blattmann, Andreas and Lorenz, Dominik and Esser, Patrick and Ommer, Bj{\"o}rn},
  booktitle={2022 IEEE/CVF conference on computer vision and pattern recognition (CVPR)},
  pages={10674--10685},
  year={2022},
  organization={ieee}
}

@inproceedings{zhu2026ants,
  title={Ants: Adaptive negative textual space shaping for ood detection via test-time mllm understanding and reasoning},
  author={Zhu, Wenjie and Zhang, Yabin and Jin, Xin and Zeng, Wenjun and Zhang, Lei},
  booktitle={Proceedings of the IEEE/CVF Conference on Computer Vision and Pattern Recognition},
  pages={20--30},
  year={2026}
}

@inproceedings{zhang2024dual,
  title={Dual memory networks: A versatile adaptation approach for vision-language models},
  author={Zhang, Yabin and Zhu, Wenjie and Tang, Hui and Ma, Zhiyuan and Zhou, Kaiyang and Zhang, Lei},
  booktitle={2024 IEEE/CVF Conference on Computer Vision and Pattern Recognition (CVPR)},
  pages={28718--28728},
  year={2024},
  organization={IEEE}
}

@article{ke2026deformba,
  title={Deformba: Vision State Space Model with Adaptive State Fusion},
  author={Ke, Hongyu and Morris, Jack and Liu, Yongkang and Kitai, Satoshi and Oguchi, Kentaro and Ding, Yi and Wang, Haoxin},
  journal={arXiv preprint arXiv:2605.21308},
  year={2026}
}

@inproceedings{ke2025mambev,
  title={Mambev: Enabling state space models to learn birds-eye-view representations},
  author={Ke, Hongyu and Morris, Jack and Oguchi, Kentaro and Cao, Xiaofei and Liu, Yongkang and Wang, Haoxin and Ding, Yi},
  booktitle={The Thirteenth International Conference on Learning Representations},
  year={2025}
}

@misc{chow2026masked,
      title={Masked Generative Transformer Is What You Need for Image Editing},
      author={Wei Chow and Linfeng Li and Xian Sun and Lingdong Kong and Zefeng Li and Qi Xu and Hang Song and Tian Ye and Xian Wang and Jinbin Bai and Shilin Xu and Xiangtai Li and Junting Pan and Shaoteng Liu and Ran Zhou and Tianshu Yang and Songhua Liu},
      year={2026},
      eprint={2605.10859},
      archivePrefix={arXiv},
      primaryClass={cs.CV},
      url={https://arxiv.org/abs/2605.10859},
      note={CVPR Workshop on HiGen},
}

@misc{kong2026driving,
      title={Is Your Driving World Model an All-Around Player?},
      author={Lingdong Kong and Ao Liang and Tianyi Yan and Hongsi Liu and Wesley Yang and Ziqi Huang and Xian Sun and Wei Yin and Jialong Zuo and Yixuan Hu and Dekai Zhu and Dongyue Lu and Youquan Liu and Guangfeng Jiang and Linfeng Li and Xiangtai Li and Long Zhuo and Lai Xing Ng and Benoit R. Cottereau and Changxin Gao and Liang Pan and Wei Tsang Ooi and Ziwei Liu},
      year={2026},
      eprint={2605.10858},
      archivePrefix={arXiv},
      primaryClass={cs.CV},
      url={https://arxiv.org/abs/2605.10858},
      note={CVPR Workshop on VideoWorldModel},
}

@inproceedings{xiao2026reversible,
  title={Reversible primitive--composition alignment for continual vision--language learning},
  author={Xiao, Canran and Xu, Tianxiang and Ma, Siyuan and Jiang, Yiyang and Gao, Haoyu and Wu, Yuhan},
  booktitle={International Conference on Learning Representations},
  volume={2026},
  pages={88700--88722},
  year={2026}
}

@inproceedings{liu2026affordance,
  title={Affordance-first decomposition for continual learning in video-language understanding},
  author={Liu, Hanzhi and Peng, Ningkang and Chen, Qianyu and Xiao, Canran and others},
  booktitle={Proceedings of the IEEE/CVF conference on computer vision and pattern recognition},
  pages={3908--3919},
  year={2026}
}

@inproceedings{zhou2026comem,
  title={Comem: Compositional concept-graph memory for vision--language adaptation},
  author={Zhou, Heng and Tang, Jing and Li, Yanshu and Xiao, Canran and Hou, Liwei and Ke, Zong and Yao, Jiawei and others},
  booktitle={International Conference on Learning Representations},
  volume={2026},
  pages={20009--20032},
  year={2026}
}

@inproceedings{li2022blip,
  author    = {Li, Junnan and Li, Dongxu and Xiong, Caiming and Hoi, Steven},
  title     = {{BLIP}: Bootstrapping Language-Image Pre-training for Unified Vision-Language Understanding and Generation},
  booktitle = {Proceedings of the 39th International Conference on Machine Learning (ICML)},
  series    = {Proceedings of Machine Learning Research},
  volume    = {162},
  pages     = {12888--12900},
  year      = {2022}
}

@inproceedings{xie2023bleep,
  author    = {Xie, Ronald and Pang, Kuan and Chung, Sai W. and Perciani, Catia T. and MacParland, Sonya A. and Wang, Bo and Bader, Gary D.},
  title     = {Spatially Resolved Gene Expression Prediction from Histology Images via Bi-modal Contrastive Learning},
  booktitle = {Advances in Neural Information Processing Systems 36 (NeurIPS)},
  year      = {2023}
}

@inproceedings{radford2021clip,
  author    = {Radford, Alec and Kim, Jong Wook and Hallacy, Chris and Ramesh, Aditya
               and Goh, Gabriel and Agarwal, Sandhini and Sastry, Girish and Askell, Amanda
               and Mishkin, Pamela and Clark, Jack and Krueger, Gretchen and Sutskever, Ilya},
  title     = {Learning Transferable Visual Models From Natural Language Supervision},
  booktitle = {Proceedings of the 38th International Conference on Machine Learning (ICML)},
  series    = {Proceedings of Machine Learning Research},
  volume    = {139}, pages = {8748--8763}, year = {2021}
}

@article{milacic2024reactome,
  title={The {Reactome} Pathway Knowledgebase 2024},
  author={Milacic, Marija and Beavers, Deidre and Conley, Patrick and Gong, Chuqiao and Gillespie, Marc and Griss, Johannes and Haw, Robin and Jassal, Bijay and Matthews, Lisa and May, Bruce and others},
  journal={Nucleic Acids Research},
  volume={52},
  number={D1},
  pages={D672--D678},
  year={2024},
  doi={10.1093/nar/gkad1025}
}

@article{phoenix2026,
  author = {Tran, Manuel and Gindra, Rushin H. and Putze, Philipp and others},
  title = {Pan-cancer virtual spatial transcriptomics from routine histology with {Phoenix}},
  journal = {bioRxiv},
  year = {2026},
  doi = {10.64898/2026.04.25.720812},
  url = {https://github.com/peng-lab/phoenix}
}

@article{liberzon2011msigdb,
  title={Molecular signatures database ({MSigDB}) 3.0},
  author={Liberzon, Arthur and Subramanian, Aravind and Pinchback, Reid and Thorvaldsd{\'o}ttir, Helga and Tamayo, Pablo and Mesirov, Jill P.},
  journal={Bioinformatics},
  volume={27},
  number={12},
  pages={1739--1740},
  year={2011},
  doi={10.1093/bioinformatics/btr260},
  url={https://pmc.ncbi.nlm.nih.gov/articles/PMC3106198/}
}

@ARTICLE{chiaruttini2022warpy,
    
AUTHOR={Chiaruttini, Nicolas  and Burri, Olivier  and Haub, Peter  and Guiet, Romain  and Sordet-Dessimoz, Jessica  and Seitz, Arne },
           
TITLE={An Open-Source Whole Slide Image Registration Workflow at Cellular Precision Using Fiji, QuPath and Elastix},
          
JOURNAL={Frontiers in Computer Science},
          
VOLUME={Volume 3 - 2021},
  
YEAR={2022},
  
URL={https://www.frontiersin.org/journals/computer-science/articles/10.3389/fcomp.2021.780026},
  
DOI={10.3389/fcomp.2021.780026},
  
ISSN={2624-9898}}

@article{ye2026osef,
  title = {{OSEF}: One-Step Evidence Fusion for Cross-Video Scene Procedure Planning},
  author = {Ye, Zhentong and Zhang, Lei and Zhou, Sijia and Yu, Yingda and Shi, Yuehan and Xuan, Jiaqi and Dong, Shuaiwu and Tong, Guanchao and Zhang, Meimei and Li, Bin},
  journal = {arXiv preprint arXiv:2607.29401},
  year = {2026},
  eprint = {2607.29401},
  archivePrefix = {arXiv},
  primaryClass = {cs.CV},
  doi = {10.48550/arXiv.2607.29401},
  url = {https://arxiv.org/abs/2607.29401}
}

@article{xu2026rivatfuse,
  title = {{RiVaT-Fuse}: Reliability-Calibrated Variational Tensor Fusion with Matrix-Valued Trust for Multimodal Prediction under Modality Uncertainty},
  author = {Xu, Yingfan and Liu, Tieming and Liang, Ye},
  journal = {arXiv preprint arXiv:2609.10798},
  year = {2026},
  eprint = {2609.10798},
  archivePrefix = {arXiv},
  url = {https://arxiv.org/abs/2609.10798}
}

@inproceedings{liang2026oc,
  title = {{OC-Distill}: Ontology-aware Contrastive Learning with Cross-Modal Distillation for ICU Risk Prediction},
  author = {Liang, Zhongyuan and Jo, Junhyung and Lee, Hyang-Jung and Kim, Sang Kyu and Chen, Irene Y.},
  booktitle = {Proceedings of the 11th Machine Learning for Healthcare Conference},
  year = {2026},
  series = {Proceedings of Machine Learning Research},
  volume = {340},
  pages = {1089--1116},
  publisher = {PMLR},
  url = {https://proceedings.mlr.press/v340/liang26a.html}
}

@misc{han2026validationstagecombinatorialfusionanalysis,
      title={Validation-Stage Combinatorial Fusion Analysis for Imbalanced Credit-Card Fraud Detection}, 
      author={Xiao Han and Chenyu Wu},
      year={2026},
      eprint={2606.10393},
      archivePrefix={arXiv},
      primaryClass={cs.LG},
      url={https://arxiv.org/abs/2606.10393}, 
}

@article {lee2024pathomclip,
	author = {Lee, Yongju and Liu, Xinhao and Hao, Minsheng and Liu, Tianyu and Regev, Aviv},
	title = {PathOmCLIP: Connecting tumor histology with spatial gene expression via locally enhanced contrastive learning of Pathology and Single-cell foundation model},
	elocation-id = {2024.12.10.627865},
	year = {2024},
	doi = {10.1101/2024.12.10.627865},
	publisher = {Cold Spring Harbor Laboratory},
	URL = {https://www.biorxiv.org/content/early/2024/12/11/2024.12.10.627865},
	eprint = {https://www.biorxiv.org/content/early/2024/12/11/2024.12.10.627865.full.pdf},
	journal = {bioRxiv}
}
\appendix

\section{Implementation details}

\subsection{Datasets}

All spatial transcriptomics data are publicly available. We obtained the
training slides from HEST-1k version 1.3.0
(\url{https://github.com/mahmoodlab/HEST}) and the 10x Genomics public
datasets portal (\url{https://www.10xgenomics.com/datasets}). The training
corpus contains $75$ human Xenium slides and $23{,}232{,}915$ cells with
matched expression measurements and H\&E locations. It includes Xenium V1 and
Xenium Prime 5K assays across $15$ tissue groups: bone marrow,
bowel or colorectal tissue, brain, breast, cervix, heart, kidney, liver,
lung, lymph node, ovary, pancreas, prostate, skin, and tonsil.
Accession-level provenance for the training and evaluation datasets is provided in Appendix~\ref{app:data-provenance-23m}.

Table~\ref{tab:data-role} gives the sizes and roles of the training
and evaluation sets. The full-scale training set is used to train VOICE-23M.
A 24-slide subset is used to train VOICE-7M for architecture and training
ablations and to assess the effect of data scale. The in-slide evaluation
uses breast ($142{,}272$ cells), lung ($161{,}453$), and skin ($87{,}499$).
The cross-slide evaluation contains ten slides and $3{,}845{,}839$ cells
across seven tissues, with two slides each from breast, lung, and ovary,
and one each from kidney, pancreas, head and neck, and stomach.
Table~\ref{tab:test-provenance} lists all evaluation specimens and their public sources.

\begin{table}[H]
\centering
\footnotesize
\setlength{\tabcolsep}{4pt}
\caption{\textbf{Data scale and experimental role.}}
\label{tab:data-role}
\begin{tabular}{@{}lrr@{}}
\toprule
Data role & Slides & Cells \\
\midrule
Full-scale training set & $75$ & $23{,}232{,}915$ \\
24-slide training subset & $24$ & $6{,}979{,}810$ \\
In-slide evaluation & $3$ & $391{,}224$ \\
Cross-slide evaluation & $10$ & $3{,}845{,}839$ \\
\bottomrule
\end{tabular}
\end{table}

\subsection{Cohort provenance and patient separation}
\label{app:provenance}
The training corpus spans $15$ tissues, $51$ patient groups, and $24$ cohorts. Its public sources include the 10x Genomics portal and HEST-1k, including GEO series GSE250346 and GSE315411. Test-slide identifiers, patient or specimen metadata, and available experiment/ROI identifiers were compared with the training inventory. All ten cross-slide targets were excluded from training and reference banks. Eight have supporting evidence of patient independence; Breast-1 and Ovary-1 do not provide sufficient donor identifiers to resolve their patient relationship to training specimens. Table~\ref{tab:test-provenance} records the evidence for each target.

\subsection{Data processing}
\label{app:preproc}

Each source dataset provides a coordinate transformation that aligns the
Xenium assay with its H\&E image. We used this transformation to locate every
cell centroid and boundary on the H\&E image. For each cell, we extracted the
$55\,\mu\mathrm{m}$ field of view defined in Section~\ref{sec:method} and
resized it to $224\times224$ pixels. A crop that extended beyond the image
was padded with white pixels. We linked each crop to its raw gene counts by
cell identifier and removed cells without a matching expression record.

\subsection{Baseline implementation}
\label{app:baselines}

We ran each baseline from its public implementation and retained its
published architecture and preprocessing. All methods use the train,
validation, and test assignments described in Section~\ref{sec:benchmark}.
The details below describe how each method was trained and adapted to these
shared assignments.

\textit{Simulated Visium spots.}
sCellST and DeepSpot2Cell learn from expression measured over Visium spots,
whereas Xenium measures individual cells. We therefore converted each
Xenium slide into simulated Visium spots for the in-slide experiments. The
spot centers form a hexagonal grid with $100\,\mu\mathrm{m}$ horizontal
spacing and $86.6\,\mu\mathrm{m}$ vertical spacing. Alternating rows are
shifted horizontally by $50\,\mu\mathrm{m}$. A cell contributes to a spot
when its centroid falls within the spot's $55\,\mu\mathrm{m}$-diameter
capture area. We summed the raw counts of all contributing cells to obtain
the expression measured at that spot. Both methods use the same simulated
spots.

\textit{GHIST}
We report both a reduced expression-only GHIST variant and the full model with its auxiliary cell-type and neighborhood objectives. The full model derives auxiliary targets exclusively from the training data and predicts them from H\&E at inference; target-slide cell-type annotations are not required. We omit the optional cell-type-average expression reference at inference. Both variants receive H\&E images and cell masks. We generate the masks
with BIDCell, the cell-segmentation method used by GHIST, under its recommended Xenium settings.

GHIST is trained from scratch on $256\times256$ H\&E crops with its published U-Net3+ image network. Image features within each segmented
nucleus are pooled into a $256$-dimensional cell representation. A feedforward network then predicts the scaled log-expression target used by
the original method, $5\log(1+\text{count})$. The reduced variant minimizes expression mean squared error, while the full variant also retains GHIST's auxiliary losses. Training uses AdamW for $51$ epochs. The initial learning rate is
$10^{-3}$, the weight decay is $10^{-4}$, and the batch size is $8$. We select the model weights with the highest mean Pearson correlation across
genes on the validation region. For cross-slide evaluation, the central $10\%$ of the training slide is reserved for validation and the remaining
$90\%$ is used for training. Only genes present in both the training and target panels are scored.

\textit{sCellST}
sCellST learns a cell-level predictor from expression measured over Visium spots. For real Visium training slides, we use nucleus masks produced by CellViT and distributed with HEST-1k. For Xenium slides, we use the vendor-provided cell masks that are also used throughout our benchmark.

Before expression training, a ResNet-50 image encoder is pretrained for $150$ epochs on cell images from the tissue-matched training slides. We use MoCo v3 contrastive learning~\cite{chen2021mocov3} and a batch size of
$4096$. Each cell is represented by a $72\times72$ pixel crop at $0.25\,\mu\mathrm{m}$ per pixel, and the encoder produces a $2048$-dimensional image feature. During expression training, a three-layer
feedforward network converts each cell feature into an expression prediction. The predictions for cells within the same spot are averaged and trained to match the measured expression of that spot.

Spot counts are normalized to $10{,}000$ total counts and transformed with $\log(1+x)$. We train with AdamW at a learning rate of $10^{-4}$ and a batch size of $128$ spots for at most $400$ epochs. A random $20\%$ of the spots are reserved for validation, and training stops after $20$ epochs without improvement. The targets are the $1000$ genes with the highest expression variance in the training data. The in-slide models are trained on the simulated spots defined above. The cross-slide models are trained on tissue-matched real Visium slides. The HEST-1k identifiers of the individual
training slides are TENX39 for breast, TENX62 for lung, and TENX65 for ovary. Kidney uses eight training slides, and pancreas uses four. After
training, sCellST is applied to individual cells on the Xenium evaluation slide.

\textit{DeepSpot2Cell.}
DeepSpot2Cell also learns from spot-level expression, but it combines the appearance of the target cell with tissue context at two larger scales. Each prediction uses a crop of the target cell, an image tile covering the simulated spot that contains the cell, and tiles from as many as six adjacent spots. A frozen Phikon v2 pathology image encoder converts each $224\times224$ image into a feature vector. Separate feedforward networks
process the target-cell feature, the feature of its containing spot, and the features of the adjacent spots. The method then combines these features to
produce one expression prediction for the target cell.

We use the simulated spots defined above to construct the training targets. Raw cell counts are summed within each spot, normalized to $10{,}000$ total counts, and transformed with $\log(1+x)$. The model learns to match this spot-level target while retaining the target-cell crop as a separate input.
Training minimizes mean squared error with AdamW. The learning rate is $10^{-4}$, the weight decay is $10^{-6}$, the dropout rate is $0.3$, and the
batch size is $32$ spots. We train for at most $100$ epochs with early stopping. At evaluation time, the model receives only H\&E images and cell
locations and returns one prediction per cell.

\textit{DeepSpot-M.}
We initialize DeepSpot-M from its released version 1 model weights. The published model uses a Midnight pathology encoder for the H\&E image and a
fixed numerical representation of each gene produced by scGPT. Its decoder combines the image features with these gene representations and produces a separate prediction for each gene. Each cell is represented by a $224\times224$ image at approximately $0.5\,\mu\mathrm{m}$ per pixel, giving
an approximately $112\,\mu\mathrm{m}$ field of view. We extract this field from the registered Xenium H\&E image and resize it to $224\times224$ pixels. DeepSpot-M can output $19{,}338$ genes. We retain the predictions that belong to the evaluation panel. If a panel gene is not among these outputs, it receives PCC zero. The evaluated cells, expression targets, and data folds are otherwise the same as for the other methods.

For fine-tuning, DeepSpot-M uses the same training slides as GHIST and DeepSpot2Cell. During fine-tuning, we update the rank-$8$ low-rank adaptation (LoRA) layers in the image encoder and the layers that combine image and gene information
to produce gene-specific predictions. The original image-encoder weights and scGPT gene representations remain fixed. This gives approximately $5.7$ million trainable parameters. Training minimizes mean squared error only
over genes measured on the training slide. We use AdamW with a learning rate of $10^{-4}$, weight decay of $10^{-6}$, a cosine learning-rate schedule,
and mixed-precision arithmetic. Training runs for at most $30$ epochs. A spatial strip containing $10\%$ of the training cells is reserved for early
stopping. For zero-shot evaluation, we apply the released model weights without additional training. For DeepSpot-M (pretrained), we pretrain the model on the same 75-slide training corpus as VOICE and apply it to all ten targets without target-specific fine-tuning.


\subsection{Evaluation metrics}
\label{app:metrics}

Table~\ref{tab:panels} gives the number of genes in the three
cross-slide evaluation panels defined in Section~\ref{sec:xsgen}. Control,
viral, and mutation probes are excluded from these counts.

\begin{table}[H]
\centering\footnotesize
\setlength{\tabcolsep}{3pt}
\caption{\textbf{Cross-slide gene-panel sizes.} Dashes indicate that the two unseen tissues are evaluated only on their full target panels.}
\label{tab:panels}
\begin{tabular}{lrrr}
\toprule
Target & Full target & Xenium-shared & Visium-shared\\
\midrule
Breast-1 & $5{,}086$ & $172$ & $390$\\
Lung-1 & $480$ & $198$ & $89$\\
Kidney & $403$ & $377$ & $68$\\
Ovary-1 & $480$ & $342$ & $111$\\
Pancreas & $380$ & $98$ & $82$\\
Breast-2 & $313$ & $280$ & $105$\\
Lung-2 & $5{,}001$ & $222$ & $373$\\
Ovary-2 & $5{,}001$ & $5{,}001$ & $421$\\
Head and neck & $377$ & -- & --\\
Stomach & $377$ & -- & --\\
\bottomrule
\end{tabular}
\end{table}

\textit{Per-gene Pearson correlation.}
Section~\ref{sec:benchmark} defines Pearson correlation coefficient (PCC) as
the agreement between predicted and measured expression across cells. Here
we give its formal definition. Let $\mathcal{C}$ contain the evaluated cells,
let $y_{ig}$ be the measured raw count of gene $g$ in cell $i$, and let
$z_{ig}=\log(1+y_{ig})$ be its log-transformed value. Let $\hat z_{ig}$ be
the corresponding prediction. Then
\begin{equation}
\mathrm{PCC}_g=
\frac{\sum_{i\in\mathcal{C}}\bigl(\hat z_{ig}-\bar{\hat z}_g\bigr)
\bigl(z_{ig}-\bar z_g\bigr)}
{\sqrt{\sum_{i\in\mathcal{C}}\bigl(\hat z_{ig}-\bar{\hat z}_g\bigr)^{2}}\,
 \sqrt{\sum_{i\in\mathcal{C}}\bigl(z_{ig}-\bar z_g\bigr)^{2}}},
\end{equation}
where $\bar z_g$ and $\bar{\hat z}_g$ are the corresponding means over
$\mathcal{C}$. For in-slide evaluation, predictions from all five test bands
are pooled before this quantity is computed. A gene set $\mathcal{G}$ is
summarized by its unweighted mean
$\overline{\mathrm{PCC}}_{\mathcal{G}}=|\mathcal{G}|^{-1}\sum_{g\in\mathcal{G}}
\mathrm{PCC}_g$. A higher value indicates better agreement. Genes that a
method cannot predict receive PCC zero, as specified in the main evaluation
protocol.

\textit{Gene subsets.}
All gene rankings use measured expression from the target slide and are
fixed before any method is scored. Highly variable genes are ranked by the
variance of $z_{ig}$ across cells. Spatially variable genes are ranked by
Moran's $I$, which is higher when nearby cells have more similar expression.
We report the top $20$, $50$, and $100$ genes from each ranking. Full-target and
Visium-shared results, and Xenium-shared results on Breast-1, Lung-1, Kidney,
Ovary-1, and Pancreas, rank genes within the corresponding evaluation panel;
when it contains fewer than
$100$ genes, the top-$100$ average uses every gene in that panel. For the
Xenium-shared results on Breast-2, Lung-2, and Ovary-2 in
Table~\ref{tab:cross-slide-xenium-cont}, all models select top-$k$ genes on the
full target panel and average over their intersection with the shared panel. We denote
the highly variable sets by H20, H50, and H100 and the spatially variable sets
by S20, S50, and S100. The \textsc{all} column averages the genes in the
corresponding evaluation panel.


\subsection{Downstream biological analyses}
\label{app:downstream}
For the pancreas analysis, we standardize measured log-expression, fit a 30-component PCA, average each cell with its 14 nearest spatial neighbors, and fit six k-means domains. The standardization, PCA, and cluster centers are then applied to predicted expression. ECM-rich regions are selected from measured expression using \textit{FN1}, \textit{SPARC}, \textit{DCN}, and \textit{LUM}, together with the Reactome ECM program~\cite{milacic2024reactome}. Selection requires a mean standardized ECM-deposition score above $0.15$, at least two markers with positive blockwise region effects above $0.05$, and a positive Reactome ECM score. We compare measured expression in the selected region with predicted expression in its projected counterpart over paired $500\,\mu$m blocks, retaining blocks with at least five region cells in each arm. Pathway scores average standardized expression over the covered members, using the same members for measured and predicted scores. Differential expression contrasts each region with the remaining cells and uses a false-discovery threshold of $q<0.05$.

Each model uses its available genes. On pancreas, VOICE, GHIST (full), and DeepSpot-M (pretrained) use $380/98/371$ genes, respectively. For each model, regions are selected separately from their own predicted expression. 

For breast, the proliferation score sums gene-wise $z$-scores for the six panel-covered MSigDB genes \textit{CCND1}, \textit{CENPF}, \textit{HMGA1}, \textit{MKI67}, \textit{SQLE}, and \textit{TOP2A}~\cite{liberzon2011msigdb}. We compute AUROC against the provided proliferative/invasive tumor epithelial labels and PCC between predicted and measured scores. For each pair of annotated subtypes within a broad cell type, we divide the difference in each gene's subtype means by its pooled standard deviation. PCC across genes compares these standardized differences between measured and predicted expression. We average over subtype pairs within each cell type, then give the five cell types equal weight. T cells, macrophages, myoepithelial cells, and dendritic cells each contribute one pair; tumor epithelial cells contribute six. 

\section{Ablation studies}
\label{sec:appx-analysis}

\subsection{Ablated components and controls}

The VOICE-7M component studies use the same five vertical bands
and five-fold assignment as the in-slide benchmark of
Section~\ref{sec:inslide} (Table~\ref{tab:inslide}). The direct branch without slide-specific fitting applies the pretrained
Stage~2 model directly to the evaluated slide. The frozen-encoder direct
branch, which ablates both training stages, trains the spatial decoder and
count head on the training bands while keeping the original UNI2-h encoder
fixed. The Stage~2 direct branch follows the same procedure but uses the
image encoder obtained after both training stages.

All retrieval rows average measured expression from neighboring
reference cells according to Eq.~\eqref{eq:knn}, and differ only in the space
in which neighbors are found. The ridge-regression control
first learns a linear map from frozen UNI2-h image features to
log-transformed expression. It then compares cells using the expression
profiles predicted by this map. The projection variant instead compares cells
using the $128$-dimensional output of the Stage~1 alignment projection. The
remaining retrieval rows compare the $1536$-dimensional output of the
image encoder before that projection. The reported VOICE-7M model uses the
final Stage~2 encoder for both branches, as specified in
Section~\ref{sec:gate}. The reference cells and fusion weights come only from
the four training bands in each fold. The two rows marked $\dagger$ use the
same $1536$-dimensional encoder output and retrieval settings. They differ
only in whether the image encoder has completed Stage~1. Macro is the
unweighted mean over the three slides.

\begin{table*}[!t]
\centering
\scriptsize
\renewcommand{\arraystretch}{0.95}
\setlength{\tabcolsep}{4pt}
\caption{\textbf{In-slide component analysis.}}
\label{tab:inslide}
\begin{tabular*}{\textwidth}{@{\extracolsep{\fill}}llccccccc}
\toprule
Slide & Method & All & H20 & H50 & H100 & S20 & S50 & S100 \\
\midrule
\multirow{9}{*}{Breast}
 & Direct branch without slide fitting & 0.3478 & 0.6655 & 0.6131 & 0.5308 & 0.7315 & 0.5998 & 0.5399 \\
 & Ridge-based retrieval from frozen encoder & 0.3621 & 0.7042 & 0.6350 & 0.5528 & 0.7992 & 0.6722 & 0.5857 \\
 & Retrieval with alignment projection, 128 dimensions & 0.3668 & 0.7092 & 0.6419 & 0.5590 & 0.8054 & 0.6773 & 0.5919 \\
 & Retrieval with frozen encoder output, 1536 dimensions$^\dagger$ & 0.3521 & 0.6782 & 0.6101 & 0.5311 & 0.7860 & 0.6554 & 0.5682 \\
 & Retrieval with Stage~1 encoder output$^\dagger$ & 0.3909 & 0.7165 & 0.6518 & 0.5731 & 0.8097 & 0.6862 & 0.6040 \\
 & Retrieval with Stage~2 encoder output & 0.3923 & 0.7203 & 0.6553 & 0.5764 & 0.8113 & 0.6889 & 0.6065 \\
 \cmidrule(l){2-9}
 & Frozen-encoder direct branch & 0.4102 & 0.7393 & 0.6738 & 0.5959 & 0.8227 & 0.7032 & 0.6212 \\
 & Stage~2 direct branch & 0.4160 & 0.7445 & 0.6796 & 0.6012 & 0.8267 & 0.7069 & 0.6253 \\
 \cmidrule(l){2-9}
 & VOICE-7M, direct and retrieval & \textbf{0.4216} & \textbf{0.7468} & \textbf{0.6821} & \textbf{0.6046} & \textbf{0.8288} & \textbf{0.7118} & \textbf{0.6312} \\
\midrule
\multirow{9}{*}{Lung}
 & Direct branch without slide fitting & 0.1952 & 0.5321 & 0.4277 & 0.3719 & 0.4299 & 0.3720 & 0.3518 \\
 & Ridge-based retrieval from frozen encoder & 0.2334 & 0.6296 & 0.5332 & 0.4558 & 0.5529 & 0.4958 & 0.4517 \\
 & Retrieval with alignment projection, 128 dimensions & 0.2359 & 0.6330 & 0.5379 & 0.4601 & 0.5586 & 0.5022 & 0.4578 \\
 & Retrieval with frozen encoder output, 1536 dimensions$^\dagger$ & 0.2283 & 0.5917 & 0.5035 & 0.4337 & 0.5219 & 0.4741 & 0.4341 \\
 & Retrieval with Stage~1 encoder output$^\dagger$ & 0.2698 & 0.6554 & 0.5630 & 0.4888 & 0.6404 & 0.5552 & 0.4983 \\
 & Retrieval with Stage~2 encoder output & 0.2707 & 0.6547 & 0.5630 & 0.4895 & 0.6373 & 0.5558 & 0.5000 \\
 \cmidrule(l){2-9}
 & Frozen-encoder direct branch & 0.2769 & 0.6607 & 0.5704 & 0.4974 & 0.5780 & 0.5294 & 0.4886 \\
 & Stage~2 direct branch & 0.2910 & 0.6748 & 0.5852 & 0.5113 & 0.6555 & 0.5789 & 0.5239 \\
 \cmidrule(l){2-9}
 & VOICE-7M, direct and retrieval & \textbf{0.2996} & \textbf{0.6791} & \textbf{0.5897} & \textbf{0.5171} & \textbf{0.6713} & \textbf{0.5913} & \textbf{0.5338} \\
\midrule
\multirow{9}{*}{Skin}
 & Direct branch without slide fitting & 0.1838 & 0.1568 & 0.2186 & 0.2450 & 0.1649 & 0.2131 & 0.2292 \\
 & Ridge-based retrieval from frozen encoder & 0.4007 & 0.8358 & 0.7131 & 0.6352 & 0.8941 & 0.8276 & 0.7329 \\
 & Retrieval with alignment projection, 128 dimensions & 0.4056 & 0.8415 & 0.7177 & 0.6413 & 0.8934 & 0.8299 & 0.7379 \\
 & Retrieval with frozen encoder output, 1536 dimensions$^\dagger$ & 0.3881 & 0.8075 & 0.6777 & 0.6048 & 0.8796 & 0.8083 & 0.7113 \\
 & Retrieval with Stage~1 encoder output$^\dagger$ & 0.4306 & 0.8484 & 0.7310 & 0.6578 & 0.9018 & 0.8397 & 0.7501 \\
 & Retrieval with Stage~2 encoder output & 0.4299 & 0.8485 & 0.7310 & 0.6575 & 0.9010 & 0.8396 & 0.7494 \\
 \cmidrule(l){2-9}
 & Frozen-encoder direct branch & 0.4286 & 0.8476 & 0.7307 & 0.6566 & 0.8909 & 0.8312 & 0.7353 \\
 & Stage~2 direct branch & 0.4361 & 0.8552 & 0.7424 & 0.6678 & 0.8966 & 0.8381 & 0.7444 \\
 \cmidrule(l){2-9}
 & VOICE-7M, direct and retrieval & \textbf{0.4500} & \textbf{0.8589} & \textbf{0.7498} & \textbf{0.6767} & \textbf{0.9079} & \textbf{0.8489} & \textbf{0.7627} \\
\midrule
\midrule
\multirow{9}{*}{\textbf{Macro}}
 & Direct branch without slide fitting & 0.2423 & 0.4514 & 0.4198 & 0.3826 & 0.4421 & 0.3950 & 0.3736 \\
 & Ridge-based retrieval from frozen encoder & 0.3321 & 0.7232 & 0.6271 & 0.5479 & 0.7487 & 0.6652 & 0.5901 \\
 & Retrieval with alignment projection, 128 dimensions & 0.3361 & 0.7279 & 0.6325 & 0.5534 & 0.7525 & 0.6698 & 0.5959 \\
 & Retrieval with frozen encoder output, 1536 dimensions$^\dagger$ & 0.3228 & 0.6925 & 0.5971 & 0.5232 & 0.7292 & 0.6459 & 0.5712 \\
 & Retrieval with Stage~1 encoder output$^\dagger$ & 0.3638 & 0.7401 & 0.6486 & 0.5732 & 0.7840 & 0.6937 & 0.6175 \\
 & Retrieval with Stage~2 encoder output & 0.3643 & 0.7412 & 0.6498 & 0.5745 & 0.7832 & 0.6948 & 0.6186 \\
 \cmidrule(l){2-9}
 & Frozen-encoder direct branch & 0.3719 & 0.7492 & 0.6583 & 0.5833 & 0.7639 & 0.6879 & 0.6151 \\
 & Stage~2 direct branch & 0.3810 & 0.7582 & 0.6691 & 0.5934 & 0.7929 & 0.7080 & 0.6312 \\
 \cmidrule(l){2-9}
 & VOICE-7M, direct and retrieval & \textbf{0.3904} & \textbf{0.7616} & \textbf{0.6739} & \textbf{0.5995} & \textbf{0.8026} & \textbf{0.7173} & \textbf{0.6426} \\
\bottomrule
\end{tabular*}
\end{table*}

Stage~1 encoder retrieval exceeds the matched frozen-encoder retrieval
row on all $21$ slide and metric combinations. Continuing to Stage~2
changes retrieval by at most $0.0038$ in this comparison. The Stage~2 direct
branch also exceeds the frozen-encoder direct branch in all $21$
combinations. Combining the Stage~2 direct and retrieval predictions gives
the highest value in every column. Its macro All PCC is $0.3904$,
compared with $0.3810$ for the Stage~2 direct branch alone.

\subsection{Contribution of the two training stages}
Table~\ref{tab:delta} adds no new experiment. It re-expresses two pairs of
rows from Table~\ref{tab:inslide} as differences, so that the effect of
removing each training stage can be read directly. The Stage~1 comparison
changes whether the image encoder has received contrastive training while
keeping the feature dimension, reference cells, and retrieval settings fixed.
The Stage~2 comparison contrasts the complete Stage~2 direct branch with the
ablated version that keeps UNI2-h frozen. Both direct models use the same
architecture and are trained and validated on the same spatial bands in each
fold, but their encoder and decoder weights are initialized differently. The
Stage~2 comparison therefore measures the difference between two complete
direct systems rather than the effect of the encoder alone. Each entry
subtracts the ablated score from the Stage~1 or Stage~2 score. A positive
value therefore favors the model that includes the corresponding training
stage.

\begin{table*}[!t]
\centering
\footnotesize
\renewcommand{\arraystretch}{0.95}
\setlength{\tabcolsep}{4pt}
\caption{\textbf{Effects of Stage~1 and Stage~2 training.}}
\label{tab:delta}
\begin{tabular*}{\textwidth}{@{\extracolsep{\fill}}lccccccc}
\toprule
Slide & $\Delta$All & $\Delta$H20 & $\Delta$H50 & $\Delta$H100 & $\Delta$S20 & $\Delta$S50 & $\Delta$S100 \\
\midrule
\multicolumn{8}{l}{\emph{Stage~1 encoder retrieval $-$ frozen encoder retrieval at matched $K$ and $\tau$}} \\
Breast & $+0.0388$ & $+0.0383$ & $+0.0417$ & $+0.0420$ & $+0.0237$ & $+0.0309$ & $+0.0358$ \\
Lung            & $+0.0415$ & $+0.0637$ & $+0.0595$ & $+0.0551$ & $+0.1185$ & $+0.0812$ & $+0.0641$ \\
Skin   & $+0.0426$ & $+0.0409$ & $+0.0534$ & $+0.0530$ & $+0.0222$ & $+0.0314$ & $+0.0388$ \\
\textbf{Macro}  & $\mathbf{+0.0409}$ & $+0.0477$ & $+0.0515$ & $+0.0500$ & $\mathbf{+0.0548}$ & $+0.0478$ & $+0.0462$ \\
\midrule
\multicolumn{8}{l}{\emph{Stage~2 direct branch $-$ frozen-encoder direct branch at matched fitting}} \\
Breast & $+0.0058$ & $+0.0051$ & $+0.0058$ & $+0.0052$ & $+0.0040$ & $+0.0037$ & $+0.0041$ \\
Lung            & $+0.0140$ & $+0.0141$ & $+0.0148$ & $+0.0140$ & $+0.0775$ & $+0.0494$ & $+0.0352$ \\
Skin   & $+0.0075$ & $+0.0076$ & $+0.0117$ & $+0.0112$ & $+0.0057$ & $+0.0069$ & $+0.0090$ \\
\textbf{Macro}  & $\mathbf{+0.0091}$ & $+0.0089$ & $+0.0108$ & $+0.0101$ & $\mathbf{+0.0290}$ & $+0.0200$ & $+0.0161$ \\
\bottomrule
\end{tabular*}
\end{table*}

Stage~1 increases macro All PCC for retrieval by $0.0409$. The
increase is similar on breast, lung, and skin, ranging from $0.0388$ to
$0.0426$. The Stage~2 direct system improves macro All PCC by
$0.0091$. Its largest gain occurs on lung, where S20 increases by $0.0775$.
In both blocks, the macro S20 increase is larger than the H20 increase.
Table~\ref{tab:inslide} further shows that continuing from Stage~1
to Stage~2 changes every macro retrieval score by at most $0.0013$. Across
these ablations, Stage~1 contributes the larger change to retrieval, whereas
Stage~2 changes the direct branch more than it changes retrieval.

\subsection{Controlled comparisons at the 23M scale}
\label{app:23m-controls}
To isolate contrastive pre-alignment, we remove Stage~1 while keeping the 75-slide Stage~2 corpus, architecture, and optimization recipe fixed. The three in-slide benchmarks were represented during Stage~1, whereas the cross-slide targets were excluded from both stages. Stage~1 raises mean direct-branch all-gene PCC from $0.3421$ to $0.4126$ on the three in-slide benchmarks and from $0.2061$ to $0.2096$ on Breast-1, Lung-1, Kidney, Ovary-1, and Pancreas (Table~\ref{tab:23m-controls}). With the same reference banks and retrieval settings, it also raises mean covered-gene retrieval PCC from $0.158$ for UNI2-h to $0.200$, improving all five targets. Thus, alignment benefits both direct prediction and reference matching, with larger direct gains on the in-slide benchmarks.

The full-scale fusion ablation increases mean all-gene PCC from $0.3982$ to $0.4217$ within slides and from $0.2240$ to $0.2284$ on Breast-2, Lung-2, and Ovary-2, improving all three. These gains show that retrieval adds information beyond direct prediction at the 23M scale and that this benefit carries over to held-out slides, extending the branch comparison in Table~\ref{tab:inslide}.

\begin{table}[t]
\centering\footnotesize
\setlength{\tabcolsep}{5pt}
\caption{\textbf{Stage~1 and fusion at the 23M scale.} Values are all-gene PCC; means give each slide equal weight.}
\label{tab:23m-controls}
\begin{tabular}{@{}lrr@{}}
\toprule
Direct prediction & w/o Stage~1 & with Stage~1 \\
\midrule
In-slide breast & 0.3649 & 0.4565 \\
In-slide lung & 0.2938 & 0.3146 \\
In-slide skin & 0.3675 & 0.4668 \\
In-slide mean & 0.3421 & 0.4126 \\
\midrule
Breast-1 & 0.1125 & 0.1345 \\
Kidney & 0.2436 & 0.2363 \\
Lung-1 & 0.2889 & 0.2848 \\
Ovary-1 & 0.1206 & 0.1421 \\
Pancreas & 0.2651 & 0.2505 \\
Cross-slide mean (5 slides) & 0.2061 & 0.2096 \\
\midrule
Fusion comparison & Direct & Fused \\
\midrule
Three in-slide targets & 0.3982 & 0.4217 \\
Breast-2, Lung-2, Ovary-2 & 0.2240 & 0.2284 \\
\bottomrule
\end{tabular}
\end{table}

\subsection{The contrastive objective}
\label{app:objective}
Stage~1 uses the CLIP objective~\cite{radford2021clip}, a symmetric InfoNCE with a hard diagonal target,
(Eq.~\eqref{eq:infonce}). We test the two established alternatives to that choice using the 24-slide training subset in Table~\ref{tab:data-role}.

BLEEP~\cite{xie2023bleep} replaces the one-hot target with a soft one that spreads
mass over cells of similar expression. Writing $a_i$ and $b_i$ for the
$\ell_2$-normalised projections of Eq.~\eqref{eq:infonce}, and $\hat h_i$ and
$\hat u_i$ for the morphology feature and the scFoundation embedding they are
projected from, standardised per dimension over the corpus and then
$\ell_2$-normalised,
\begin{equation}
\begin{split}
  p_{ij} &= \frac{\exp\!\big(\langle a_i, b_j\rangle/\tau_c\big)}
                 {\sum_{k\in\mathcal{B}} \exp\!\big(\langle a_i, b_k\rangle/\tau_c\big)},
  \\[3pt]
  M_{ij} &= \tfrac{1}{2}\big(\langle \hat h_i, \hat h_j\rangle
                           + \langle \hat u_i, \hat u_j\rangle\big),
  \qquad
  Q_{ij} = \frac{\exp\!\big(M_{ij}/\tau_{\mathrm{tgt}}\big)}
                {\sum_{k\in\mathcal{B}} \exp\!\big(M_{ik}/\tau_{\mathrm{tgt}}\big)},
  \\[3pt]
  \mathcal{L}_{\mathrm{soft}} &= -\frac{1}{2|\mathcal{B}|}
    \sum_{i,j\in\mathcal{B}} \Big[Q_{ij}\,\log p_{ij} + Q_{ji}\,\log p_{ji}\Big].
\end{split}
\label{eq:softtarget}
\end{equation}
This is the \emph{BLEEP soft target} row of
Tables~\ref{tab:objective-inslide} and~\ref{tab:objective-cross}.

BLIP~\cite{li2022blip} instead keeps the one-hot target and adds a head that scores
whether a pair matches,
\begin{equation}
\begin{split}
  m_{ij} &= \sigma\!\Big(\mathbf{w}^{\top}\psi\big(\big[\,\tilde a_i \,;\, \tilde b_j \,;\,
             \tilde a_i \odot \tilde b_j \,;\, |\tilde a_i - \tilde b_j|\,\big]\big)\Big),
  \\[3pt]
  n(i) &= \operatorname*{arg\,max}_{j \neq i}\ \langle a_i, b_j\rangle,
  \\[3pt]
  \mathcal{L}_{\text{match}} &= -\frac{1}{|\mathcal{B}|}\sum_{i\in\mathcal{B}}
    \Big[\log m_{ii} + \log\big(1 - m_{i\,n(i)}\big)\Big],
\end{split}
  \label{eq:matchhead}
\end{equation}
where $\tilde a_i$ and $\tilde b_j$ are the projections \emph{before} $\ell_2$
normalisation, $\odot$ is the elementwise product, and $\psi$ is an MLP layer. The MLP layer's input is the pooled cell feature, and
$n(i)$ is the hardest mismatched cell in the batch under the current alignment.
Stage~1 optimises $\mathcal{L}_{\text{Stage-1}} + \lambda\mathcal{L}_{\text{match}}$
with $\lambda=1$. The three \emph{BLIP matching head}
rows share this encoder and differ only in how the head is used at retrieval.
\emph{Loss only} discards it at retrieval, so the head acts only through training, \emph{reweight} multiplies the neighbour
weights of Eq.~\eqref{eq:knn} by $m_{i\,\pi_k(i)}$ and renormalises, and
\emph{rerank} takes the top $100$ neighbours by cosine and keeps the $20$ the head
scores highest. BLIP's third objective, image-grounded language modelling, is not added. It supervises generation of the paired modality from the image, which is what Stage~2 already does through the count head of Eq.~\eqref{eq:nb}, so adding it to Stage~1 would duplicate a component the model already has. 

\paragraph{Protocol.} Every experiment holds the image encoder fixed and train only the projection
towers. These numbers are on their own
scale, using frozen $128$-d towers at $K{=}20$. 

We run the experiments under both evaluation protocols of
Section~\ref{sec:benchmark}. In the in-slide benchmark each slide retrieves against its
own training bands (Table~\ref{tab:objective-inslide}), and in the cross-slide benchmark each slide is queried against other slides of the same tissue
(Table~\ref{tab:objective-cross}).

\begin{table*}[t]
\centering
\small
\setlength{\tabcolsep}{5pt}
\caption{\textbf{Contrastive objectives under the in-slide protocol.} The three in-slide
evaluation slides of Table~\ref{tab:data-role}, each retrieving against its
own four training bands under the inslide protocol.
All three are held out of tower training.}
\label{tab:objective-inslide}
\begin{tabular*}{\textwidth}{@{\extracolsep{\fill}}llccccccc}
\toprule
Slide & Objective & All & H20 & H50 & H100 & S20 & S50 & S100 \\
\midrule
\multirow{5}{*}{Breast}
 & BLIP matching head, rerank    & 0.3186 & 0.6649 & 0.5945 & 0.5095 & 0.7819 & 0.6392 & 0.5468 \\
 & BLIP matching head, reweight  & 0.3625 & 0.7054 & 0.6377 & 0.5549 & 0.8036 & 0.6746 & 0.5884 \\
 & BLIP matching head, loss only & 0.3677 & 0.7091 & 0.6420 & \textbf{0.5596} & 0.8047 & \textbf{0.6774} & \textbf{0.5921} \\
 & BLEEP soft target             & \textbf{0.3683} & 0.7069 & 0.6406 & 0.5590 & 0.8022 & 0.6753 & 0.5907 \\
 & \textbf{CLIP (ours)}       & 0.3668 & \textbf{0.7092} & \textbf{0.6419} & 0.5590 & \textbf{0.8054} & 0.6773 & 0.5919 \\
\midrule
\multirow{5}{*}{Lung}
 & BLIP matching head, rerank    & 0.1929 & 0.6044 & 0.4990 & 0.4107 & 0.5140 & 0.4466 & 0.3992 \\
 & BLIP matching head, reweight  & 0.2292 & 0.6291 & 0.5328 & 0.4537 & 0.5442 & 0.4911 & 0.4485 \\
 & BLIP matching head, loss only & 0.2348 & 0.6314 & 0.5369 & \textbf{0.4593} & 0.5473 & 0.4965 & 0.4548 \\
 & BLEEP soft target             & \textbf{0.2359} & 0.6304 & 0.5362 & \textbf{0.4593} & 0.5491 & 0.4967 & 0.4550 \\
 & \textbf{CLIP (ours)}       & \textbf{0.2359} & \textbf{0.6330} & \textbf{0.5379} & \textbf{0.4601} & \textbf{0.5586} & \textbf{0.5022} & \textbf{0.4578} \\
\midrule
\multirow{5}{*}{Skin}
 & BLIP matching head, rerank    & 0.3708 & 0.8324 & 0.6934 & 0.6149 & 0.8844 & 0.8180 & 0.7130 \\
 & BLIP matching head, reweight  & 0.4019 & 0.8401 & 0.7148 & 0.6381 & 0.8927 & 0.8290 & 0.7358 \\
 & BLIP matching head, loss only & 0.4051 & 0.8408 & 0.7167 & 0.6403 & 0.8931 & 0.8297 & 0.7375 \\
 & BLEEP soft target             & \textbf{0.4056} & 0.8397 & 0.7158 & 0.6397 & 0.8917 & 0.8282 & 0.7361 \\
 & \textbf{CLIP (ours)}       & \textbf{0.4056} & \textbf{0.8415} & \textbf{0.7177} & \textbf{0.6413} & \textbf{0.8934} & \textbf{0.8299} & \textbf{0.7379} \\
\midrule
\multirow{5}{*}{\textbf{Macro}}
 & BLIP matching head, rerank    & 0.2941 & 0.7006 & 0.5956 & 0.5117 & 0.7268 & 0.6346 & 0.5530 \\
 & BLIP matching head, reweight  & 0.3312 & 0.7249 & 0.6285 & 0.5489 & 0.7468 & 0.6649 & 0.5909 \\
 & BLIP matching head, loss only & 0.3359 & 0.7271 & 0.6318 & 0.5531 & 0.7484 & 0.6679 & 0.5948 \\
 & BLEEP soft target             & \textbf{0.3366} & 0.7257 & 0.6308 & 0.5526 & 0.7477 & 0.6667 & 0.5939 \\
 & \textbf{CLIP (ours)}       & 0.3361 & \textbf{0.7279} & \textbf{0.6325} & \textbf{0.5534} & \textbf{0.7525} & \textbf{0.6698} & \textbf{0.5959} \\
\bottomrule
\end{tabular*}
\end{table*}

\begin{table*}[t]
\centering
\small
\setlength{\tabcolsep}{5pt}
\caption{\textbf{Contrastive objectives under the cross-slide protocol.} The same
six objectives and the same frozen-encoder protocol as
Table~\ref{tab:objective-inslide}, evaluated on Breast-1, Lung-1, Kidney,
Ovary-1, and Pancreas; each slide is queried against a bank of other slides
of the same tissue over their shared panel. The four non-breast sections are absent from the
pretraining manifest and so are held out of tower training; the breast section is not,
so its row measures partly memorised structure. Macro is over all five sections;
excluding the breast section leaves every objective's sign and ranking unchanged.}
\label{tab:objective-cross}
\begin{tabular*}{\textwidth}{@{\extracolsep{\fill}}llccccccc}
\toprule
Slide & Objective & All & H20 & H50 & H100 & S20 & S50 & S100 \\
\midrule
\multirow{5}{*}{Breast-1}
 & BLIP matching head, rerank    & 0.1456 & 0.4000 & 0.3001 & 0.2116 & 0.4346 & 0.3146 & 0.2180 \\
 & BLIP matching head, reweight  & 0.1616 & 0.4231 & 0.3194 & 0.2308 & 0.4475 & 0.3355 & 0.2392 \\
 & BLIP matching head, loss only & 0.1644 & \textbf{0.4270} & 0.3227 & 0.2341 & \textbf{0.4495} & 0.3389 & 0.2427 \\
 & BLEEP soft target             & \textbf{0.1675} & 0.4262 & \textbf{0.3234} & \textbf{0.2363} & 0.4449 & \textbf{0.3394} & \textbf{0.2459} \\
 & \textbf{CLIP (ours)}       & 0.1636 & 0.4245 & 0.3219 & 0.2333 & 0.4476 & 0.3382 & 0.2418 \\
\midrule
\multirow{5}{*}{Lung-1}
 & BLIP matching head, rerank    & 0.1941 & 0.3878 & 0.3048 & 0.1956 & 0.3563 & 0.3046 & 0.1960 \\
 & BLIP matching head, reweight  & 0.2433 & 0.4508 & 0.3707 & 0.2451 & 0.4142 & 0.3658 & 0.2454 \\
 & BLIP matching head, loss only & 0.2512 & 0.4593 & 0.3800 & 0.2530 & 0.4221 & 0.3746 & 0.2533 \\
 & BLEEP soft target             & \textbf{0.2625} & \textbf{0.4743} & \textbf{0.3949} & \textbf{0.2643} & \textbf{0.4369} & \textbf{0.3890} & \textbf{0.2647} \\
 & \textbf{CLIP (ours)}       & 0.2543 & 0.4636 & 0.3837 & 0.2561 & 0.4264 & 0.3785 & 0.2564 \\
\midrule
\multirow{5}{*}{Kidney}
 & BLIP matching head, rerank    & 0.1503 & 0.3603 & 0.3559 & 0.3300 & 0.3383 & 0.3299 & 0.3125 \\
 & BLIP matching head, reweight  & 0.1912 & 0.4598 & 0.4407 & 0.4053 & 0.4309 & 0.4143 & 0.3867 \\
 & BLIP matching head, loss only & 0.1963 & 0.4689 & 0.4484 & 0.4127 & 0.4405 & \textbf{0.4227} & 0.3946 \\
 & BLEEP soft target             & \textbf{0.1973} & \textbf{0.4783} & \textbf{0.4509} & \textbf{0.4147} & \textbf{0.4425} & 0.4214 & \textbf{0.3950} \\
 & \textbf{CLIP (ours)}       & 0.1963 & 0.4691 & 0.4484 & 0.4141 & 0.4378 & 0.4210 & 0.3942 \\
\midrule
\multirow{5}{*}{Ovary-1}
 & BLIP matching head, rerank    & 0.0906 & 0.3346 & 0.2526 & 0.1681 & 0.3580 & 0.2416 & 0.1600 \\
 & BLIP matching head, reweight  & 0.1015 & 0.3424 & 0.2708 & 0.1848 & 0.3818 & 0.2638 & 0.1777 \\
 & BLIP matching head, loss only & \textbf{0.1020} & \textbf{0.3429} & \textbf{0.2715} & \textbf{0.1855} & \textbf{0.3826} & \textbf{0.2647} & \textbf{0.1785} \\
 & BLEEP soft target             & 0.1009 & 0.3198 & 0.2602 & 0.1830 & 0.3620 & 0.2581 & 0.1755 \\
 & \textbf{CLIP (ours)}       & 0.1015 & 0.3357 & 0.2675 & 0.1845 & 0.3775 & 0.2615 & 0.1775 \\
\midrule
\multirow{5}{*}{Pancreas}
 & BLIP matching head, rerank    & 0.1327 & 0.3664 & 0.2170 & 0.1327 & 0.3550 & 0.2165 & 0.1327 \\
 & BLIP matching head, reweight  & 0.1624 & 0.4057 & 0.2542 & 0.1624 & 0.3907 & 0.2521 & 0.1624 \\
 & BLIP matching head, loss only & 0.1694 & 0.4141 & 0.2629 & 0.1694 & 0.4007 & 0.2603 & 0.1694 \\
 & BLEEP soft target             & 0.1685 & 0.4127 & 0.2613 & 0.1685 & 0.3941 & 0.2590 & 0.1685 \\
 & \textbf{CLIP (ours)}       & \textbf{0.1724} & \textbf{0.4198} & \textbf{0.2653} & \textbf{0.1724} & \textbf{0.4065} & \textbf{0.2648} & \textbf{0.1724} \\
\midrule
\multirow{5}{*}{\textbf{Macro}}
 & BLIP matching head, rerank    & 0.1427 & 0.3698 & 0.2861 & 0.2076 & 0.3684 & 0.2815 & 0.2039 \\
 & BLIP matching head, reweight  & 0.1720 & 0.4164 & 0.3312 & 0.2457 & 0.4130 & 0.3263 & 0.2423 \\
 & BLIP matching head, loss only & 0.1767 & 0.4224 & 0.3371 & 0.2509 & 0.4191 & 0.3322 & 0.2477 \\
 & BLEEP soft target             & \textbf{0.1793} & 0.4223 & \textbf{0.3381} & \textbf{0.2534} & 0.4161 & \textbf{0.3334} & \textbf{0.2499} \\
 & \textbf{CLIP (ours)}       & 0.1776 & \textbf{0.4225} & 0.3374 & 0.2521 & \textbf{0.4192} & 0.3328 & 0.2485 \\
\bottomrule
\end{tabular*}
\end{table*}

\textbf{Adding the matching head to the loss doesn't improve.} Compared with the \emph{BLIP matching head, loss only} model, CLIP is ahead on
all seven metrics under both protocols, always by less than $0.005$.

\textbf{Using the matching head at inference hurts.} Re-weighting the $K$ neighbours by the BLIP matching
head's score loses all seven metrics under both protocols, and the two-stage
rerank loses all seven on every individual slide of both tables, giving up
about $0.04$ All either way. The head is not uninformative, but retrieval
averages $K$ neighbours, and sharpening the ranking works against an
aggregation whose benefit comes from the breadth of the neighbourhood rather
than the precision of its top match.

\textbf{The soft target ties.} The BLEEP model wins some metrics from CLIP. For cross-slide it wins five of seven, losing H20 and S20. For in-slide wins only all gene pcc, losing all other metrics. Every
margin is of order $10^{-3}$. We then keep the hard
diagonal target for simplicity because it needs no target temperature and places no
constraint on how batches are assembled.

\subsection{Why per-gene fusion helps}
\label{app:whygate}
\begin{figure*}[t]
\centering
\includegraphics[width=\textwidth]{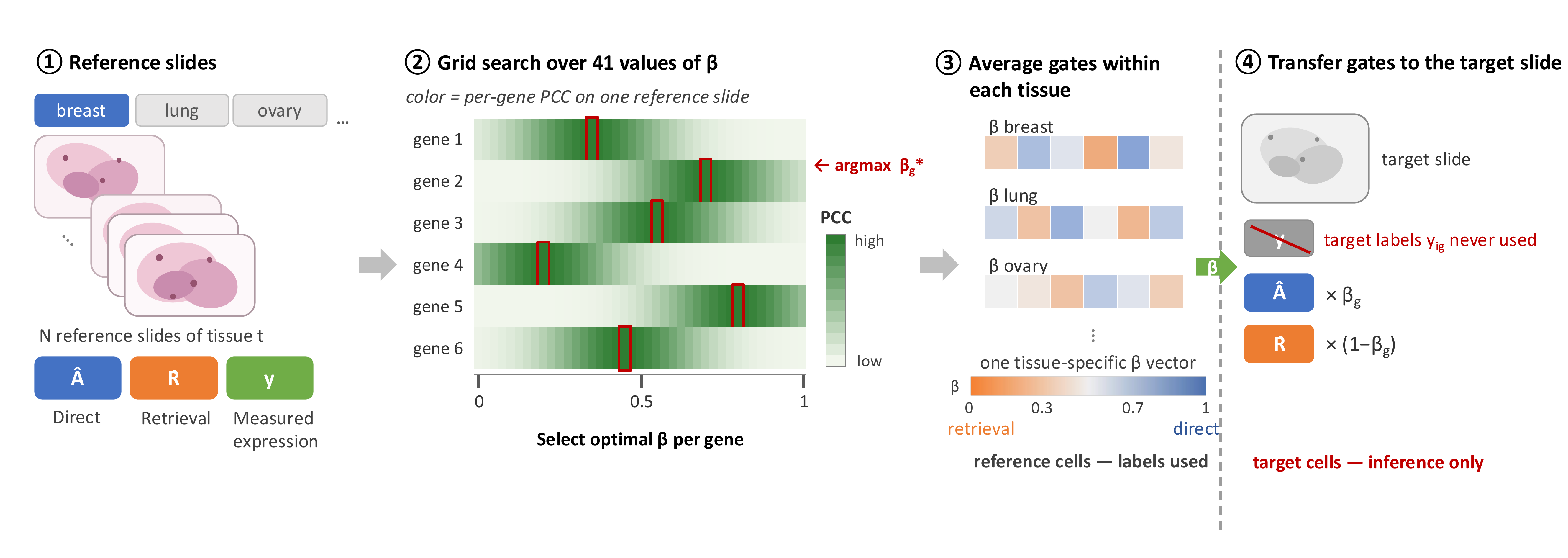}
\caption{\textbf{Fusion-weight workflow.}}
\Description{Reference slides provide direct, retrieval, and measured expression values. A gene-specific gate is selected on each reference slide, averaged within tissue, and transferred to an unlabeled target slide to fuse the two predictions.}
\label{fig:gate}
\end{figure*}

The two single-branch ablations in Table~\ref{tab:inslide} explain why
neither branch can be dropped. The Stage~2 direct branch has a higher macro
All PCC than the Stage~2 retrieval branch, yet retrieval is better for the
S20, S50, and S100 subsets on skin, so neither branch is uniformly
better across genes and slides. As described in Section~\ref{sec:gate} and
illustrated in Fig.~\ref{fig:gate}, using the In-slide protocol desctibed in Section~\ref{sec:benchmark}, Stage~3
selects one weight $\beta_g$ per gene on the training bands, choosing the
value that maximizes the correlation between the fused prediction and
measured expression, and that weight is then fixed when predicting the test
band. Because the two branches read the same morphology representation
through different mechanisms, their errors are only partly shared, so a
gene-specific convex combination can be more accurate than either branch on
its own. This is what allows VOICE-7M to exceed both individual branches in
every column of Table~\ref{tab:inslide}.

\subsection{Fusion-weight stability and calibration}
\label{app:gate-calibration}
The relative accuracy of the two branches varies by gene: retrieval outperforms direct prediction for $362$ of the $784$ reference-covered genes ($46.2\%$) across Breast-1, Lung-1, Kidney, Ovary-1, and Pancreas.

We compare averaging gene-specific weights optimized separately on training reference slides with fitting them after pooling cells from same-tissue reference slides. Pooled fitting centers each slide before combining its cells. The resulting weights are evaluated on the five held-out slides in Table~\ref{tab:beta-fitting}. Averaging achieves higher all-gene PCC on every slide, with mean PCC $0.2111$ versus $0.2101$ and Lung-1 PCC $0.2945$ versus $0.2932$. This supports the slide-wise rule in Section~\ref{sec:gate}.

\begin{table*}[tp]
\centering\footnotesize
\renewcommand{\arraystretch}{1.05}
\setlength{\tabcolsep}{4pt}
\caption{\textbf{Slide-wise averaging versus pooled fitting of fusion weights.} Both rules fit a weight for each gene using the same training reference cells. Values are PCC on held-out slides; Macro is the equal-slide mean. Bold marks the higher value within each comparison.}
\label{tab:beta-fitting}
\begin{tabular*}{\textwidth}{@{\extracolsep{\fill}}llrrrrrrr}
\toprule
Target & Weight fitting & All & S20 & S50 & S100 & H20 & H50 & H100 \\
\midrule
Breast-1 & Slide-wise average & \textbf{0.1355} & \textbf{0.5634} & \textbf{0.5020} & \textbf{0.4672} & 0.5539 & \textbf{0.5219} & \textbf{0.4752} \\
 & Pooled fitting & 0.1354 & 0.5613 & 0.5007 & 0.4665 & \textbf{0.5542} & 0.5210 & 0.4743 \\
\midrule
Lung-1 & Slide-wise average & \textbf{0.2945} & \textbf{0.6363} & \textbf{0.5923} & \textbf{0.5165} & \textbf{0.6687} & \textbf{0.5808} & \textbf{0.5392} \\
 & Pooled fitting & 0.2932 & 0.6330 & 0.5888 & 0.5145 & 0.6573 & 0.5772 & 0.5365 \\
\midrule
Kidney & Slide-wise average & \textbf{0.2290} & \textbf{0.3197} & \textbf{0.3608} & \textbf{0.3933} & 0.1577 & \textbf{0.3296} & \textbf{0.3733} \\
 & Pooled fitting & 0.2280 & 0.3183 & 0.3603 & 0.3919 & \textbf{0.1578} & 0.3291 & 0.3729 \\
\midrule
Ovary-1 & Slide-wise average & \textbf{0.1453} & \textbf{0.4363} & \textbf{0.3872} & \textbf{0.3168} & \textbf{0.4217} & \textbf{0.3685} & \textbf{0.3192} \\
 & Pooled fitting & 0.1430 & 0.4202 & 0.3794 & 0.3095 & 0.4063 & 0.3559 & 0.3112 \\
\midrule
Pancreas & Slide-wise average & \textbf{0.2511} & \textbf{0.4951} & \textbf{0.5032} & \textbf{0.4487} & \textbf{0.5927} & \textbf{0.5268} & \textbf{0.4599} \\
 & Pooled fitting & 0.2509 & 0.4930 & 0.5021 & 0.4482 & 0.5920 & 0.5257 & 0.4593 \\
\midrule
Macro & Slide-wise average & \textbf{0.2111} & \textbf{0.4901} & \textbf{0.4691} & \textbf{0.4285} & \textbf{0.4790} & \textbf{0.4655} & \textbf{0.4333} \\
 & Pooled fitting & 0.2101 & 0.4852 & 0.4663 & 0.4261 & 0.4735 & 0.4618 & 0.4308 \\
\bottomrule
\end{tabular*}
\end{table*}

We also compare the grid-fitted weights with an MLP gate using standardized direct and retrieval predictions, their element-wise product, the cell's mean absolute branch difference, and a learned 16-dimensional gene embedding. The MLP has two 64-unit hidden layers and predicts a convex weight for each cell and gene with a sigmoid output. Under within-slide spatial five-fold cross-validation on the five slides in Table~\ref{tab:mlp-gate}, grid fitting achieves higher PCC on all seven metrics for every slide. Its mean all-gene PCC is $0.4448$, compared with $0.4366$ for the MLP trained with squared-error loss. This favors the simpler per-gene weights fitted directly to the evaluation objective.

\begin{table*}[tp]
\centering\footnotesize
\renewcommand{\arraystretch}{1.05}
\setlength{\tabcolsep}{4pt}
\caption{\textbf{Within-slide evaluation of grid fitting and an MLP gate.} Fusion is fitted on four spatial bands and evaluated on the held-out band; PCC uses predictions pooled over the five test bands. Breast rep1 and rep2 are the two replicates of sample~1. Macro averages the five slides. Bold marks the higher PCC in each comparison.}
\label{tab:mlp-gate}
\begin{tabular*}{\textwidth}{@{\extracolsep{\fill}}llrrrrrrr}
\toprule
Slide & Fusion method & All & S20 & S50 & S100 & H20 & H50 & H100 \\
\midrule
Breast rep1 & Grid fitting & \textbf{0.4837} & \textbf{0.8581} & \textbf{0.7753} & \textbf{0.7071} & \textbf{0.8563} & \textbf{0.7652} & \textbf{0.6871} \\
 & MLP gate & 0.4771 & 0.8541 & 0.7697 & 0.7008 & 0.8527 & 0.7606 & 0.6825 \\
\midrule
Breast rep2 & Grid fitting & \textbf{0.4780} & \textbf{0.8576} & \textbf{0.7751} & \textbf{0.7053} & \textbf{0.8608} & \textbf{0.7628} & \textbf{0.6870} \\
 & MLP gate & 0.4693 & 0.8521 & 0.7661 & 0.6957 & 0.8555 & 0.7567 & 0.6807 \\
\midrule
Breast sample2 & Grid fitting & \textbf{0.4631} & \textbf{0.8564} & \textbf{0.7575} & \textbf{0.6774} & \textbf{0.7875} & \textbf{0.7245} & \textbf{0.6477} \\
 & MLP gate & 0.4544 & 0.8507 & 0.7476 & 0.6678 & 0.7821 & 0.7189 & 0.6416 \\
\midrule
Lung & Grid fitting & \textbf{0.3222} & \textbf{0.7149} & \textbf{0.6331} & \textbf{0.5661} & \textbf{0.6947} & \textbf{0.6060} & \textbf{0.5349} \\
 & MLP gate & 0.3156 & 0.7069 & 0.6255 & 0.5595 & 0.6897 & 0.6013 & 0.5299 \\
\midrule
Skin & Grid fitting & \textbf{0.4771} & \textbf{0.9252} & \textbf{0.8677} & \textbf{0.7880} & \textbf{0.8720} & \textbf{0.7724} & \textbf{0.6987} \\
 & MLP gate & 0.4667 & 0.9161 & 0.8598 & 0.7767 & 0.8677 & 0.7659 & 0.6919 \\
\midrule
Macro & Grid fitting & \textbf{0.4448} & \textbf{0.8424} & \textbf{0.7617} & \textbf{0.6888} & \textbf{0.8143} & \textbf{0.7262} & \textbf{0.6511} \\
 & MLP gate & 0.4366 & 0.8360 & 0.7537 & 0.6801 & 0.8095 & 0.7207 & 0.6453 \\
\bottomrule
\end{tabular*}
\end{table*}

Table~\ref{tab:beta-stability} shows that gene-specific weights are generally stable across reference slides within a tissue: median weight SD is $0.000$--$0.018$ outside breast. Pancreas and kidney have mean absolute cross-slide weight differences of $0.029$ and $0.016$, despite low rank correlations. The small absolute variation supports a tissue-level average that reflects consistent branch preferences across reference slides, with limited influence from atypical slides for most genes. Breast is the exception, with larger between-slide differences in the preferred balance of direct prediction and retrieval.

To examine calibration on held-out slides, we start from the same Stage~1 checkpoint and train Stage~2 on a donor-grouped split of 53 slides, reserving 22 for calibration. With Stage~2 fixed, held-out fitting lowers mean $\beta$ from $0.848$ to $0.812$ and raises test PCC from $0.2090$ to $0.2106$, improving all five targets and all seven metrics. Thus, held-out calibration shifts some weight toward retrieval and improves transfer. Training Stage~2 on all 75 slides with training-slide calibration gives $0.2128$ (Table~\ref{tab:gate-calibration}), indicating that retaining the full training corpus provides the larger overall benefit.

\begin{table*}[t]
\centering\footnotesize
\setlength{\tabcolsep}{6pt}
\caption{\textbf{Support and stability of tissue-specific fusion weights.} Support lists the number of slides:number of genes. SD summarizes slide-specific weights across supported genes. Pair $\rho$ is the median Spearman correlation over finite slide pairs; pair difference is the median mean absolute weight difference. Three breast pairs have undefined $\rho$ because one weight vector is constant.}
\label{tab:beta-stability}
\begin{tabular}{lrrrr}
\toprule
Tissue & Support (slides:genes) & SD median / P90 & Pair $\rho$ (pairs) & Pair difference \\
\midrule
Breast & 1:100; 4:280 & 0.471 / 0.489 & 0.424 (3) & 0.482 \\
Lung & 2:5003; 4:239 & 0.018 / 0.071 & 0.494 (6) & 0.151 \\
Ovary & 2:4767; 3:278 & 0.000 / 0.035 & 0.099 (3) & 0.049 \\
Pancreas & 2:377 & 0.018 / 0.053 & 0.226 (1) & 0.029 \\
Kidney & 2:377 & 0.000 / 0.035 & 0.063 (1) & 0.016 \\
\bottomrule
\end{tabular}
\end{table*}

\begin{table}[t]
\centering\footnotesize
\setlength{\tabcolsep}{4pt}
\caption{\textbf{Fusion calibration.} Mean all-gene PCC on Breast-1, Lung-1, Kidney, Ovary-1, and Pancreas. The 53/22-slide split groups slides by donor.}
\label{tab:gate-calibration}
\begin{tabular}{@{}llr@{}}
\toprule
Stage~2 training & Gate fitting & PCC \\
\midrule
75 slides & Training slides & 0.2128 \\
53 slides & Same 53 slides & 0.2090 \\
53 slides & Held-out 22 slides & 0.2106 \\
\bottomrule
\end{tabular}
\end{table}

\subsection{Cell-feature construction}
\label{sec:ablation}

This subsection reports one further ablation and one sensitivity check, both
on the in-slide evaluation breast slide under the five-fold in-slide protocol
defined in Section~\ref{sec:benchmark}, holding the decoder, training
schedule, and evaluation fixed. Every variant follows the frozen-encoder
direct branch of Table~\ref{tab:inslide}, which trains the spatial decoder
and count head on the training bands while keeping the original UNI2-h
encoder fixed. These scores are therefore comparable to that row and
not to the complete VOICE-7M model, so they isolate the cell-feature
construction from the effect of Stage~1 and Stage~2 training. UNI2-h
represents each crop as a $16\times16$ grid of $256$ image tokens, where each
token summarizes a local image region. The standard cell-mask pooling in
Eq.~\eqref{eq:cellmask} weights these tokens by how much they overlap the
target cell. The ablation replaces this cell-specific weighting with an
unweighted average over all $256$ tokens at a matched crop size, and
therefore isolates the pooling rule. The sensitivity check varies the crop
size while retaining cell-mask pooling, and therefore isolates how much
surrounding tissue the crop contains. Table~\ref{tab:ablation} reports both
comparisons.

\begin{table*}[!t]
\centering
\footnotesize
\renewcommand{\arraystretch}{0.95}
\setlength{\tabcolsep}{4pt}
\caption{\textbf{Cell-feature ablations on in-slide evaluation breast slide.}}
\label{tab:ablation}
\begin{tabular*}{\textwidth}{@{\extracolsep{\fill}}lccccccc}
\toprule
Variant & All & H20 & H50 & H100 & S20 & S50 & S100 \\
\midrule
Cell-mask pooling, 224-pixel crop   & \textbf{0.4102} & \textbf{0.7393} & \textbf{0.6738} & \textbf{0.5959} & \textbf{0.8227} & \textbf{0.7032} & \textbf{0.6212} \\
Cell-mask pooling, 300-pixel crop   & 0.4093 & 0.7376 & 0.6714 & 0.5939 & 0.8212 & 0.7019 & 0.6197 \\
Uniform average over 300-pixel crop & 0.3229 & 0.6617 & 0.5829 & 0.4999 & 0.7722 & 0.6395 & 0.5429 \\
\cmidrule(l){1-8}
$\Delta$ (300-pixel $-$ 224-pixel cell mask)  & $-0.0009$ & $-0.0017$ & $-0.0024$ & $-0.0020$ & $-0.0015$ & $-0.0013$ & $-0.0015$ \\
$\Delta$ (uniform average $-$ cell mask at 300 pixels) & $\mathbf{-0.0864}$ & $-0.0759$ & $-0.0885$ & $-0.0940$ & $-0.0490$ & $-0.0623$ & $-0.0768$ \\
\bottomrule
\end{tabular*}
\end{table*}

At a matched $300$-pixel crop, replacing cell-mask pooling with a uniform
average over the whole crop lowers All PCC from $0.4093$ to
$0.3229$, and lowers every reported subset by $0.0490$ to $0.0940$. Because
the crop is identical in both rows, the difference is attributable to the
pooling rule rather than to the amount of surrounding tissue the crop
contains. Increasing the crop from $224$ to $300$ pixels while retaining
cell-mask pooling changes every score by at most $0.0024$. Concentrating the
pooled representation on the target cell is therefore what matters, not the
size of the field of view around it.


\subsection{Sensitivity to cell-boundary perturbations}
\label{app:mask-sensitivity}
We perturb the pooling masks with $1\,\mu$m erosion, dilation, and translation on Breast-1, Lung-1, Kidney, Ovary-1, and Pancreas, using approximately 30,000 cells per slide sampled in complete spatial blocks. Images, cell coordinates, model weights, reference banks, and fusion weights remain fixed; the features and predictions are recomputed for each perturbed mask. Translation is averaged over four directions. Table~\ref{tab:mask-sensitivity} reports each slide's PCC and mask-area changes. Erosion removes $35.6$--$49.1\%$ of mask area and dilation adds $42.4$--$63.1\%$, yet the largest slide-level PCC decrease is $0.0059$. Mean PCC decreases are $0.0003$, $0.0032$, and $0.0002$ for erosion, dilation, and translation, respectively. This stability under substantial area changes shows that cell-specific pooling tolerates imprecise cell boundaries.

\begin{table*}[t]
\centering\footnotesize
\setlength{\tabcolsep}{6pt}
\caption{\textbf{Sensitivity to $1\,\mu$m cell-boundary perturbations.} All-gene PCC is evaluated on the same sampled cells for all conditions. Translation averages four directions. Area change is the mean percentage change from the original mask; translation preserves area.}
\label{tab:mask-sensitivity}
\begin{tabular}{lrrrrrr}
\toprule
& \multicolumn{4}{c}{All-gene PCC} & \multicolumn{2}{c}{Area change (\%)} \\
\cmidrule(lr){2-5}\cmidrule(lr){6-7}
Slide & Original & Erosion & Dilation & Translation & Erosion & Dilation \\
\midrule
Breast-1 & 0.1358 & 0.1356 & 0.1349 & 0.1361 & $-49.1$ & $+63.1$ \\
Lung-1 & 0.2931 & 0.2949 & 0.2872 & 0.2926 & $-35.6$ & $+42.4$ \\
Kidney & 0.2393 & 0.2361 & 0.2372 & 0.2377 & $-45.3$ & $+56.6$ \\
Ovary-1 & 0.1465 & 0.1462 & 0.1442 & 0.1469 & $-41.2$ & $+50.2$ \\
Pancreas & 0.2504 & 0.2509 & 0.2456 & 0.2508 & $-41.2$ & $+50.9$ \\
\bottomrule
\end{tabular}
\end{table*}

\section{Cross-slide analyses}

\subsection{Notation for the cross-tissue gene comparison}
\label{app:crosstissue-notation}
Table~\ref{tab:cross-tissue-genes} reports the VOICE-23M direct branch on
the same-tissue and other-tissue-only gene groups defined in
Section~\ref{sec:xsgen}. The two groups are disjoint and cover genes measured
in at least one of the $75$ Stage~2 training slides. Breast-2 excludes
\textit{FBLIM1} and \textit{REXO4}, which are absent from those training panels.
The column $n$ gives the number of genes in a group. All $5{,}001$ genes on
Lung-2 and Ovary-2 have same-tissue training support, so their other-tissue-only
groups have $n=0$ and undefined PCC, shown as dashes.

Macro averages the per-slide scores over nonempty groups: eight slides for
same-tissue genes and six for other-tissue-only genes. The gene-weighted
summary weights each slide's score by its number of evaluated genes in that
group and metric: $n$ for All and $\min(k,n)$ for S$k$ or H$k$.
Each slide--gene pair therefore receives equal weight within a metric.
$\Delta$ is the other-tissue-only score minus the same-tissue score within
each slide where both groups are nonempty.

\subsection{Complete cross-slide results}
\label{sec:xsgen_appendix}
Tables~\ref{tab:cross-slide-target}--\ref{tab:cross-slide-visium} report the
remaining cross-slide results, including the All, S100, and H100 columns
omitted from Table~\ref{tab:cross-slide-summary}, on the full target,
Xenium-shared, and Visium-shared panels defined in
Section~\ref{sec:benchmark}. Both shared-panel comparisons cover eight targets
from tissues represented in training. Full target-panel results also include the
two unseen tissues in Table~\ref{tab:unseen-tissue} in the main text, giving ten
targets overall. Gene subsets follow the
selection rules in Section~\ref{app:metrics}.

\begin{table*}[!t]
\centering
\small
\setlength{\tabcolsep}{4pt}
\caption{
\textbf{Cross-slide generalization: full target panel.}
 DeepSpot-M (pretrained) is pretrained on the same training data as VOICE; the no-Stage-1 model uses the same Stage-2 recipe. Bold marks the best reported result in each column.
}
\label{tab:cross-slide-target}

\resizebox{\textwidth}{!}{%
\begin{tabular}{l ccccccc ccccccc ccccccc}
\toprule

& \multicolumn{7}{c}{Breast-1}
& \multicolumn{7}{c}{Lung-1}
& \multicolumn{7}{c}{Kidney} \\

\cmidrule(lr){2-8}
\cmidrule(lr){9-15}
\cmidrule(lr){16-22}

Model
& All & S20 & S50 & S100 & H20 & H50 & H100
& All & S20 & S50 & S100 & H20 & H50 & H100
& All & S20 & S50 & S100 & H20 & H50 & H100 \\

\midrule

DeepSpot-M (zero-shot)
& 0.047 & 0.193 & 0.184 & 0.188 & 0.214 & 0.196 & 0.186 & 0.060 & 0.172 & 0.148 & 0.123 & 0.153 & 0.136 & 0.125 & 0.040 & 0.177 & 0.127 & 0.109 & 0.117 & 0.108 & 0.082 \\

DeepSpot-M (fine-tuned)
& 0.079 & 0.348 & 0.339 & 0.323 & 0.355 & 0.332 & 0.282 & 0.219 & 0.522 & 0.447 & 0.396 & 0.500 & 0.437 & 0.385 & 0.221 & 0.563 & 0.479 & 0.435 & \textbf{0.554} & 0.473 & 0.441 \\

DeepSpot-M (pretrained)
& 0.1255 & 0.4257 & 0.4136 & 0.4052 & 0.4523 & 0.4122 & 0.4082 & 0.2739 & 0.5882 & 0.5485 & 0.4808 & 0.6661 & 0.5624 & 0.5212 & 0.2221 & 0.5507 & 0.4795 & 0.4299 & 0.5483 & 0.4588 & 0.4254 \\
VOICE-23M w/o Stage~1
& 0.1130 & \textbf{0.5336} & \textbf{0.4856} & 0.4389 & 0.5430 & 0.4991 & 0.4577 & \textbf{0.2965} & 0.6231 & \textbf{0.6017} & 0.5224 & 0.6596 & 0.5850 & \textbf{0.5492} & \textbf{0.2443} & 0.5888 & 0.4887 & \textbf{0.4602} & 0.5334 & \textbf{0.4878} & \textbf{0.4604} \\

VOICE-7M
& 0.088 & 0.424 & 0.382 & 0.373 & 0.469 & 0.441 & 0.414 & 0.248 & 0.601 & 0.530 & 0.463 & 0.653 & 0.547 & 0.498 & 0.230 & 0.489 & 0.448 & 0.424 & 0.456 & 0.432 & 0.430 \\

VOICE-23M
& \textbf{0.135} & 0.515 & 0.480 & \textbf{0.453} & \textbf{0.559} & \textbf{0.519} & \textbf{0.473} & 0.293 & \textbf{0.673} & 0.599 & \textbf{0.526} & \textbf{0.684} & \textbf{0.589} & 0.540 & 0.241 & \textbf{0.602} & \textbf{0.507} & 0.456 & \textbf{0.554} & 0.477 & 0.454 \\

\bottomrule
\end{tabular}%
}

\par\vspace{-0.5pt}

\resizebox{0.70\textwidth}{!}{%
\begin{tabular}{l ccccccc ccccccc}
\toprule

& \multicolumn{7}{c}{Ovary-1}
& \multicolumn{7}{c}{Pancreas} \\

\cmidrule(lr){2-8}
\cmidrule(lr){9-15}

Model
& All & S20 & S50 & S100 & H20 & H50 & H100
& All & S20 & S50 & S100 & H20 & H50 & H100 \\

\midrule

DeepSpot-M (zero-shot)
& 0.020 & 0.125 & 0.081 & 0.052 & 0.038 & 0.057 & 0.059 & 0.100 & 0.308 & 0.260 & 0.217 & 0.312 & 0.223 & 0.179 \\

DeepSpot-M (fine-tuned)
& 0.107 & 0.418 & 0.323 & 0.255 & 0.327 & 0.283 & 0.241 & 0.142 & 0.458 & 0.378 & 0.294 & 0.487 & 0.360 & 0.274 \\

DeepSpot-M (pretrained)
& 0.1140 & 0.3719 & 0.3134 & 0.2560 & 0.3075 & 0.2875 & 0.2566 & 0.2338 & 0.4451 & 0.4814 & 0.4229 & 0.6180 & 0.5273 & 0.4445 \\
VOICE-23M w/o Stage~1
& 0.1239 & 0.4167 & 0.3551 & 0.2870 & 0.3553 & 0.3186 & 0.2781 & \textbf{0.2656} & 0.5174 & \textbf{0.5264} & \textbf{0.4671} & \textbf{0.6329} & \textbf{0.5557} & \textbf{0.4798} \\

VOICE-7M
& 0.132 & \textbf{0.464} & \textbf{0.390} & \textbf{0.316} & \textbf{0.439} & \textbf{0.381} & \textbf{0.317} & 0.234 & 0.564 & 0.510 & 0.442 & 0.618 & 0.527 & 0.440 \\

VOICE-23M
& \textbf{0.146} & 0.433 & 0.379 & 0.313 & 0.420 & 0.360 & 0.312 & 0.248 & \textbf{0.566} & 0.517 & 0.465 & 0.592 & 0.519 & 0.456 \\

\bottomrule
\end{tabular}%
}
\end{table*}

\begin{table*}[!t]
\ContinuedFloat
\centering\scriptsize
\renewcommand{\arraystretch}{0.95}
\setlength{\tabcolsep}{4pt}
\caption{\textbf{Cross-slide generalization: full target panel, continued.}}
\label{tab:cross-slide-target-cont}
\begin{tabular*}{\textwidth}{@{\extracolsep{\fill}}llrrrrrrr}
\toprule
Target & Model & All & S20 & S50 & S100 & H20 & H50 & H100 \\
\midrule
Breast-2 & DeepSpot-M (fine-tuned) & 0.2938 & 0.6941 & 0.5349 & 0.4752 & 0.7118 & 0.5545 & 0.4649 \\
 & DeepSpot-M (pretrained) & 0.3026 & 0.6519 & 0.5612 & 0.5050 & 0.6659 & 0.5579 & 0.4775 \\
 & VOICE-23M w/o Stage~1 & 0.3745 & 0.7284 & 0.6329 & 0.5822 & 0.7395 & 0.6526 & 0.5849 \\
 & VOICE-23M & \textbf{0.4242} & \textbf{0.8106} & \textbf{0.7182} & \textbf{0.6508} & \textbf{0.8156} & \textbf{0.7121} & \textbf{0.6362} \\
\midrule
Lung-2 & DeepSpot-M (fine-tuned) & 0.0619 & 0.4032 & 0.3184 & 0.2803 & 0.3678 & 0.3154 & 0.2991 \\
 & DeepSpot-M (pretrained) & 0.0841 & 0.4740 & 0.3871 & 0.3498 & 0.4987 & 0.4020 & 0.3667 \\
 & VOICE-23M w/o Stage~1 & \textbf{0.0983} & \textbf{0.4906} & \textbf{0.4235} & \textbf{0.3847} & \textbf{0.5350} & \textbf{0.4633} & \textbf{0.4219} \\
 & VOICE-23M & 0.0846 & 0.4480 & 0.3815 & 0.3572 & 0.4978 & 0.4431 & 0.4011 \\
\midrule
Ovary-2 & DeepSpot-M (fine-tuned) & 0.1457 & 0.4292 & 0.4363 & 0.3969 & 0.4823 & 0.4932 & 0.4679 \\
 & DeepSpot-M (pretrained) & 0.1840 & \textbf{0.5260} & \textbf{0.5403} & \textbf{0.4962} & \textbf{0.6149} & 0.5927 & 0.5638 \\
 & VOICE-23M w/o Stage~1 & \textbf{0.1949} & 0.4943 & 0.5178 & 0.4946 & 0.6093 & \textbf{0.6015} & \textbf{0.5856} \\
 & VOICE-23M & 0.1765 & 0.4382 & 0.4589 & 0.4487 & 0.5396 & 0.5385 & 0.5261 \\
\bottomrule
\end{tabular*}
\end{table*}

\begin{table*}[!t]
\centering
\small
\setlength{\tabcolsep}{4pt}
\caption{
\textbf{Cross-slide generalization: Xenium-shared panel.}
 GHIST (full) retains cell-type and neighborhood supervision. All rows use the same gene lists and rankings within the shared panel (Appendix~\ref{app:metrics}). Bold marks the best result in each column.
}
\label{tab:cross-slide-xenium}

\resizebox{\textwidth}{!}{%
\begin{tabular}{l ccccccc ccccccc ccccccc}
\toprule

& \multicolumn{7}{c}{Breast-1}
& \multicolumn{7}{c}{Lung-1}
& \multicolumn{7}{c}{Kidney} \\

\cmidrule(lr){2-8}
\cmidrule(lr){9-15}
\cmidrule(lr){16-22}

Model
& All & S20 & S50 & S100 & H20 & H50 & H100
& All & S20 & S50 & S100 & H20 & H50 & H100
& All & S20 & S50 & S100 & H20 & H50 & H100 \\

\midrule

GHIST (reduced)
& 0.084 & 0.367 & 0.223 & 0.134 & 0.327 & 0.220 & 0.134 & 0.116 & 0.280 & 0.229 & 0.192 & 0.281 & 0.259 & 0.201 & 0.123 & 0.333 & 0.323 & 0.271 & 0.405 & 0.340 & 0.287 \\

GHIST (full)
& 0.0950 & 0.3747 & 0.2301 & 0.1476 & 0.3397 & 0.2266 & 0.1453 & 0.1045 & 0.2906 & 0.2139 & 0.1733 & 0.2836 & 0.2331 & 0.1772 & 0.1102 & 0.2548 & 0.2549 & 0.2298 & 0.2945 & 0.2654 & 0.2403 \\

DeepSpot2Cell
& 0.061 & 0.231 & 0.149 & 0.094 & 0.215 & 0.144 & 0.091 & 0.105 & 0.312 & 0.221 & 0.168 & 0.280 & 0.220 & 0.167 & 0.050 & 0.165 & 0.145 & 0.118 & 0.187 & 0.149 & 0.118 \\

DeepSpot-M (zero-shot)
& 0.056 & 0.187 & 0.125 & 0.084 & 0.155 & 0.124 & 0.086 & 0.055 & 0.162 & 0.123 & 0.101 & 0.176 & 0.122 & 0.100 & 0.038 & 0.166 & 0.139 & 0.104 & 0.124 & 0.108 & 0.081 \\

DeepSpot-M (fine-tuned)
& 0.150 & 0.423 & 0.304 & 0.223 & 0.392 & 0.294 & 0.210 & 0.232 & 0.448 & 0.387 & 0.350 & 0.424 & 0.421 & 0.349 & 0.211 & 0.464 & 0.457 & 0.414 & 0.527 & 0.458 & 0.414 \\

VOICE-23M w/o Stage~1
& 0.1955 & 0.4977 & 0.3864 & 0.2843 & 0.4797 & 0.3697 & 0.2677 & \textbf{0.2816} & 0.5898 & 0.4899 & \textbf{0.4258} & 0.5659 & \textbf{0.5098} & \textbf{0.4254} & \textbf{0.2412} & 0.5091 & \textbf{0.4870} & \textbf{0.4551} & \textbf{0.5375} & \textbf{0.4761} & \textbf{0.4544} \\

VOICE-7M
& 0.185 & 0.487 & 0.361 & 0.268 & 0.460 & 0.350 & 0.255 & 0.265 & 0.564 & 0.471 & 0.406 & 0.557 & 0.498 & 0.406 & 0.229 & 0.495 & 0.465 & 0.431 & 0.512 & 0.463 & 0.435 \\

VOICE-23M
& \textbf{0.205} & \textbf{0.511} & \textbf{0.395} & \textbf{0.296} & \textbf{0.494} & \textbf{0.381} & \textbf{0.279} & 0.281 & \textbf{0.598} & \textbf{0.492} & 0.423 & \textbf{0.573} & 0.508 & 0.421 & 0.233 & \textbf{0.517} & 0.482 & 0.436 & 0.525 & 0.465 & 0.437 \\

\bottomrule
\end{tabular}%
}

\par\vspace{-0.5pt}

\resizebox{0.70\textwidth}{!}{%
\begin{tabular}{l ccccccc ccccccc}
\toprule

& \multicolumn{7}{c}{Ovary-1}
& \multicolumn{7}{c}{Pancreas} \\

\cmidrule(lr){2-8}
\cmidrule(lr){9-15}

Model
& All & S20 & S50 & S100 & H20 & H50 & H100
& All & S20 & S50 & S100 & H20 & H50 & H100 \\

\midrule

GHIST (reduced)
& 0.053 & 0.230 & 0.170 & 0.137 & 0.297 & 0.236 & 0.156 & 0.088 & 0.245 & 0.148 & 0.088 & 0.279 & 0.155 & 0.088 \\

GHIST (full)
& 0.0152 & 0.0868 & 0.0660 & 0.0477 & 0.1315 & 0.0859 & 0.0506 & 0.0984 & 0.2860 & 0.1650 & 0.0984 & 0.2840 & 0.1603 & 0.0984 \\

DeepSpot2Cell
& 0.043 & 0.267 & 0.170 & 0.126 & 0.247 & 0.196 & 0.135 & 0.068 & 0.181 & 0.114 & 0.068 & 0.162 & 0.104 & 0.068 \\

DeepSpot-M (zero-shot)
& 0.022 & 0.108 & 0.068 & 0.057 & 0.067 & 0.080 & 0.057 & 0.107 & 0.276 & 0.173 & 0.107 & 0.227 & 0.163 & 0.107 \\

DeepSpot-M (fine-tuned)
& 0.106 & 0.413 & 0.299 & 0.224 & 0.322 & 0.270 & 0.225 & 0.173 & 0.424 & 0.269 & 0.173 & 0.414 & 0.273 & 0.173 \\

VOICE-23M w/o Stage~1
& 0.1200 & 0.3986 & 0.3218 & 0.2452 & 0.3281 & 0.2854 & 0.2475 & \textbf{0.2349} & 0.4678 & \textbf{0.3301} & \textbf{0.2349} & \textbf{0.4541} & \textbf{0.3308} & \textbf{0.2349} \\

VOICE-7M
& 0.128 & \textbf{0.474} & \textbf{0.369} & 0.270 & \textbf{0.439} & \textbf{0.358} & 0.285 & 0.216 & 0.454 & 0.315 & 0.216 & 0.446 & 0.319 & 0.216 \\

VOICE-23M
& \textbf{0.144} & 0.449 & 0.359 & \textbf{0.276} & 0.413 & 0.344 & \textbf{0.288} & 0.227 & \textbf{0.469} & 0.328 & 0.227 & 0.454 & 0.330 & 0.227 \\

\bottomrule
\end{tabular}%
}
\end{table*}

\begin{table*}[!t]
\ContinuedFloat
\centering\scriptsize
\renewcommand{\arraystretch}{0.95}
\setlength{\tabcolsep}{4pt}
\caption{\textbf{Cross-slide generalization: Xenium-shared panel, continued.} All models use full-target top-$k$ genes intersected with the shared panel for these three slides. Bold marks the best result per slide and metric.}
\label{tab:cross-slide-xenium-cont}
\begin{tabular*}{\textwidth}{@{\extracolsep{\fill}}llrrrrrrr}
\toprule
Target & Model & All & S20 & S50 & S100 & H20 & H50 & H100 \\
\midrule
Breast-2 & GHIST (full) & 0.0349 & 0.1165 & 0.0752 & 0.0635 & 0.1247 & 0.0871 & 0.0666 \\
 & VOICE-23M w/o Stage~1 & 0.3940 & 0.7284 & 0.6379 & 0.5902 & 0.7395 & 0.6537 & 0.5862 \\
 & VOICE-23M & \textbf{0.4734} & \textbf{0.8492} & \textbf{0.7593} & \textbf{0.6916} & \textbf{0.8525} & \textbf{0.7536} & \textbf{0.6780} \\
\midrule
Lung-2 & GHIST (full) & 0.0502 & 0.2126 & 0.2018 & 0.1853 & 0.4544 & 0.3931 & 0.3523 \\
 & VOICE-23M w/o Stage~1 & \textbf{0.1266} & \textbf{0.4025} & \textbf{0.3764} & \textbf{0.3552} & 0.6197 & \textbf{0.5297} & 0.4781 \\
 & VOICE-23M & 0.1218 & 0.3843 & 0.3597 & 0.3525 & \textbf{0.6425} & 0.5275 & \textbf{0.4878} \\
\midrule
Ovary-2 & GHIST (full) & 0.0331 & 0.1335 & 0.1283 & 0.1380 & 0.2077 & 0.2194 & 0.2210 \\
 & VOICE-23M w/o Stage~1 & \textbf{0.1949} & \textbf{0.4943} & \textbf{0.5178} & \textbf{0.4946} & \textbf{0.6093} & \textbf{0.6015} & \textbf{0.5856} \\
 & VOICE-23M & 0.1765 & 0.4382 & 0.4589 & 0.4487 & 0.5396 & 0.5385 & 0.5261 \\
\bottomrule
\end{tabular*}
\end{table*}

\begin{table*}[!t]
\centering
\small
\setlength{\tabcolsep}{4pt}
\caption{
\textbf{Cross-slide generalization on the Visium-shared gene panel.}
Results cover eight targets from tissues represented in training. All models use
the same gene lists and rankings within each shared panel. Bold marks the best
result per slide and metric.
}
\label{tab:cross-slide-visium}

\resizebox{\textwidth}{!}{%
\begin{tabular}{l ccccccc ccccccc ccccccc}
\toprule

& \multicolumn{7}{c}{Breast-1}
& \multicolumn{7}{c}{Lung-1}
& \multicolumn{7}{c}{Kidney} \\

\cmidrule(lr){2-8}
\cmidrule(lr){9-15}
\cmidrule(lr){16-22}

Model
& All & S20 & S50 & S100 & H20 & H50 & H100
& All & S20 & S50 & S100 & H20 & H50 & H100
& All & S20 & S50 & S100 & H20 & H50 & H100 \\

\midrule

GHIST (full)
& 0.0228 & 0.1203 & 0.1119 & 0.0627 & 0.1088 & 0.1146 & 0.0643 & 0.0614 & 0.0792 & 0.0779 & 0.0614 & 0.1231 & 0.0819 & 0.0614 & 0.1889 & 0.2328 & 0.2213 & 0.1889 & 0.2073 & 0.2100 & 0.1889 \\

sCellST
& 0.044 & 0.170 & 0.161 & 0.113 & 0.144 & 0.153 & 0.109 & 0.057 & 0.083 & 0.066 & 0.057 & 0.094 & 0.075 & 0.057 & 0.045 & 0.100 & 0.053 & 0.045 & 0.077 & 0.044 & 0.045 \\

DeepSpot-M zero-shot
& 0.071 & 0.188 & 0.196 & 0.156 & 0.169 & 0.195 & 0.155 & 0.092 & 0.145 & 0.125 & 0.092 & 0.140 & 0.118 & 0.092 & 0.086 & 0.123 & 0.102 & 0.086 & 0.121 & 0.097 & 0.086 \\

DeepSpot-M fine-tuned
& 0.133 & 0.300 & 0.294 & 0.235 & 0.268 & 0.308 & 0.247 & 0.287 & 0.427 & 0.369 & 0.287 & 0.439 & 0.369 & 0.287 & 0.399 & 0.514 & 0.460 & 0.399 & 0.535 & 0.449 & 0.399 \\

VOICE-23M w/o Stage~1
& 0.1851 & 0.4298 & 0.4001 & 0.3210 & 0.4514 & 0.4153 & 0.3382 & \textbf{0.3739} & \textbf{0.5666} & \textbf{0.4818} & \textbf{0.3739} & \textbf{0.5537} & \textbf{0.4828} & \textbf{0.3739} & \textbf{0.4441} & \textbf{0.5409} & \textbf{0.4950} & \textbf{0.4441} & \textbf{0.5719} & \textbf{0.4900} & \textbf{0.4441} \\

VOICE-7M
& 0.145 & 0.390 & 0.369 & 0.287 & 0.384 & 0.367 & 0.294 & 0.318 & 0.478 & 0.409 & 0.318 & 0.488 & 0.418 & 0.318 & 0.399 & 0.442 & 0.440 & 0.399 & 0.460 & 0.437 & 0.399 \\

VOICE-23M
& \textbf{0.201} & \textbf{0.481} & \textbf{0.430} & \textbf{0.345} & \textbf{0.475} & \textbf{0.438} & \textbf{0.361} & 0.373 & 0.558 & 0.479 & 0.373 & 0.544 & 0.480 & 0.373 & 0.428 & 0.534 & 0.481 & 0.428 & 0.566 & 0.474 & 0.428 \\

\bottomrule
\end{tabular}%
}

\par\vspace{-0.5pt}

\resizebox{0.70\textwidth}{!}{%
\begin{tabular}{l ccccccc ccccccc}
\toprule

& \multicolumn{7}{c}{Ovary-1}
& \multicolumn{7}{c}{Pancreas} \\

\cmidrule(lr){2-8}
\cmidrule(lr){9-15}

Model
& All & S20 & S50 & S100 & H20 & H50 & H100
& All & S20 & S50 & S100 & H20 & H50 & H100 \\

\midrule

GHIST (full)
& 0.0098 & 0.0271 & 0.0233 & 0.0109 & 0.0434 & 0.0243 & 0.0109 & 0.0364 & 0.0780 & 0.0427 & 0.0364 & 0.0919 & 0.0419 & 0.0364 \\

sCellST
& 0.042 & 0.109 & 0.070 & 0.046 & 0.100 & 0.077 & 0.046 & 0.075 & 0.113 & 0.090 & 0.075 & 0.117 & 0.097 & 0.075 \\

DeepSpot-M zero-shot
& 0.031 & 0.104 & 0.057 & 0.035 & 0.077 & 0.056 & 0.035 & 0.130 & 0.218 & 0.179 & 0.130 & 0.218 & 0.172 & 0.130 \\

DeepSpot-M fine-tuned
& 0.153 & 0.336 & 0.249 & 0.166 & 0.304 & 0.252 & 0.167 & 0.212 & 0.332 & 0.271 & 0.212 & 0.375 & 0.261 & 0.212 \\

VOICE-23M w/o Stage~1
& 0.1703 & 0.3352 & 0.2623 & 0.1831 & 0.3122 & 0.2740 & 0.1844 & \textbf{0.3527} & \textbf{0.5333} & \textbf{0.4517} & \textbf{0.3527} & \textbf{0.5645} & \textbf{0.4482} & \textbf{0.3527} \\

VOICE-7M
& 0.179 & \textbf{0.389} & \textbf{0.292} & 0.192 & \textbf{0.384} & \textbf{0.305} & 0.195 & 0.310 & 0.480 & 0.391 & 0.310 & 0.515 & 0.394 & 0.310 \\

VOICE-23M
& \textbf{0.189} & 0.361 & 0.286 & \textbf{0.204} & 0.354 & 0.301 & \textbf{0.204} & 0.350 & 0.522 & 0.440 & 0.350 & 0.548 & 0.438 & 0.350 \\

\bottomrule
\end{tabular}%
}
\end{table*}

\begin{table*}[tp]
\ContinuedFloat
\centering\scriptsize
\renewcommand{\arraystretch}{0.95}
\setlength{\tabcolsep}{4pt}
\caption{\textbf{Cross-slide generalization on the Visium-shared gene panel, continued.} The shared panels contain $105$, $373$, and $421$ genes for Breast-2, Lung-2, and Ovary-2, respectively. All models use the same genes and within-panel SVG/HVG rankings. Bold marks the best result per slide and metric.}
\label{tab:cross-slide-visium-cont}
\begin{tabular*}{\textwidth}{@{\extracolsep{\fill}}llrrrrrrr}
\toprule
Target & Model & All & S20 & S50 & S100 & H20 & H50 & H100 \\
\midrule
Breast-2 & GHIST (full) & 0.0421 & 0.1055 & 0.0653 & 0.0434 & 0.1140 & 0.0706 & 0.0442 \\
 & sCellST & 0.1382 & 0.2447 & 0.1865 & 0.1432 & 0.2465 & 0.1928 & 0.1449 \\
 & DeepSpot-M (zero-shot) & 0.1545 & 0.2645 & 0.2096 & 0.1604 & 0.2613 & 0.2087 & 0.1606 \\
 & DeepSpot-M (fine-tuned) & 0.3587 & 0.6183 & 0.4883 & 0.3716 & 0.6439 & 0.5000 & 0.3728 \\
 & VOICE-23M w/o Stage~1 & 0.4587 & 0.7036 & 0.6030 & 0.4761 & 0.6992 & 0.6061 & 0.4764 \\
 & VOICE-23M & \textbf{0.5438} & \textbf{0.8086} & \textbf{0.7064} & \textbf{0.5639} & \textbf{0.7950} & \textbf{0.6943} & \textbf{0.5627} \\
\midrule
Lung-2 & GHIST (full) & 0.0088 & 0.0313 & 0.0279 & 0.0216 & 0.0236 & 0.0133 & 0.0186 \\
 & sCellST & 0.0150 & 0.0468 & 0.0344 & 0.0278 & 0.0861 & 0.0554 & 0.0408 \\
 & DeepSpot-M (zero-shot) & 0.0135 & 0.0607 & 0.0477 & 0.0326 & 0.0676 & 0.0481 & 0.0352 \\
 & DeepSpot-M (fine-tuned) & 0.0852 & 0.2921 & 0.2236 & 0.1787 & 0.3369 & 0.2419 & 0.1911 \\
 & VOICE-23M w/o Stage~1 & \textbf{0.1324} & \textbf{0.3866} & \textbf{0.3117} & \textbf{0.2590} & \textbf{0.4637} & \textbf{0.3419} & \textbf{0.2697} \\
 & VOICE-23M & 0.1170 & 0.3595 & 0.2929 & 0.2419 & 0.4545 & 0.3190 & 0.2510 \\
\midrule
Ovary-2 & GHIST (full) & 0.0249 & 0.0043 & 0.0105 & 0.0208 & 0.0422 & 0.0514 & 0.0601 \\
 & sCellST & 0.0591 & 0.1928 & 0.1525 & 0.1304 & 0.1725 & 0.1455 & 0.1264 \\
 & DeepSpot-M (zero-shot) & 0.0595 & 0.1336 & 0.1182 & 0.0948 & 0.1396 & 0.1220 & 0.0988 \\
 & DeepSpot-M (fine-tuned) & 0.1640 & 0.3202 & 0.2949 & 0.2598 & 0.3934 & 0.3611 & 0.3258 \\
 & VOICE-23M w/o Stage~1 & \textbf{0.2267} & \textbf{0.4299} & \textbf{0.3984} & \textbf{0.3412} & \textbf{0.5333} & \textbf{0.4606} & \textbf{0.4135} \\
 & VOICE-23M & 0.2028 & 0.4105 & 0.3731 & 0.3121 & 0.5095 & 0.4349 & 0.3917 \\
\bottomrule
\end{tabular*}
\end{table*}

\subsection{Ovarian training panels across model scales}
\label{app:ovary-panel}

Scaling from VOICE-7M to VOICE-23M raises the all-gene score on Breast-1,
Lung-1, Kidney, Ovary-1, and Pancreas, and raises every ranked subset on
Breast-1, Lung-1, and Kidney. Ovary-1 is the exception: its all-gene score rises by $0.014$ while all six
ranked subsets fall, by $0.003$ to $0.031$
(Table~\ref{tab:cross-slide-target}). Pancreas loses two subsets, and no other
tissue loses any.

The ovarian training panels differ in assay generation between model scales.
Ovary-1 and the single ovary slide in the $24$-slide subset are both
Xenium V1, carrying $480$ and $477$ genes. Both ovary slides that the
full-scale set adds are Xenium Prime 5K: $200{,}815$ cells over $5{,}001$
genes and $407{,}124$ cells over $5{,}090$ genes. Scaling therefore takes the
V1 share of ovary training cells from all of them to roughly a quarter, so a
V1 target slide now draws most of its same-tissue supervision from the other
generation.

A gene shared by the two panels is not measured identically on them. Probe
set, panel size and per-gene detection sensitivity all differ, so the same
gene carries a different count distribution on Prime than on V1, and the
shared count head of Section~\ref{sec:decoder} receives both as supervision
for one output. This offers a mechanism for the split we observe: the Prime
slides raise the all-gene score by covering an order of magnitude more genes,
while the ranked subsets, the high-signal genes on which a scale mismatch
between generations bites hardest, move the other way.

\section{Cross-Platform Generalization to CosMx Data}
Every training pair used by VOICE is a single cell whose morphology, and
measured transcriptome are belong to the same cell. Xenium
distributes its H\&E and DAPI images in a common registered frame. However, CosMx does not provide an equivalent registration for
externally acquired H\&E. To generalize our model to CosMx data, we therefore developed a nucleus-level procedure that establishes the missing
correspondence directly to make
VOICE extend to CosMx. 

\subsection{Procedure}
\label{app:dapi-align}
The procedure works at two scales. A whole-slide registration first places the
two images from H\&E and DAPI images in a common coordinate frame. Nucleus-level matching then decides which
individual cell corresponds to. It treats nuclei as landmarks shared by the two modalities and filters candidate nucleus-to-nucleus matches until only
geometrically self-consistent ones survive.

\begin{enumerate}
\item \textbf{Whole-slide registration.} We register the H\&E image to the
CosMx DAPI image with Warpy~\cite{chiaruttini2022warpy}, an open-source Fiji
workflow for whole-slide registration. Registration is computed with elastix
and can be refined interactively in BigWarp. Warpy stores the transform and leaves both
images at their original resolution, so the same transform maps CosMx cell
coordinates onto H\&E pixels without resampling either image. 

\item \textbf{Nucleus segmentation and patch extraction.} We segment nuclei
from the H\&E image with Cellpose and convert the label mask to per-nucleus
polygons; CosMx supplies its own cell centroids and boundaries on the
molecular side. We then crop a patch of fixed radius around every nucleus
centre in both modalities.

\item \textbf{Local patch matching.} For each H\&E nucleus, we score every
DAPI nuclei within a search radius based on the image similarity of their patches.
To tolerate the residual offset that step~1 leaves behind, we score four shifted versions from top, down, right, and left directions of each patch and keep the worst score over the shifts, so a
match must hold under perturbation. The top-ranked candidates are retained.

\item \textbf{Field-of-view masking.} We restrict matches to the true imaged
field of view and discard a margin at the tile border, because a nucleus cut
by an edge cannot be scored reliably.

\item \textbf{Nucleus-shape validation.} We overlay the two nucleus polygons
of each surviving pair in the common frame and require them to overlap. A true
match is the same nucleus seen twice and therefore agrees in position and
extent, whereas a locally ambiguous match usually does not.
\end{enumerate}

The runs below use normalised cross-correlation as the patch similarity, a
$20$ px search radius, a $20$ px patch radius, shifts up to $10$ px, the top
$3$ candidates per H\&E nucleus, a $30$ px border margin, and a
nucleus-polygon intersection filter on the rank-one candidate.

\subsection{Whole-slide check on pair quality}

Three public CosMx 6K slides, from ovary, liver and colon, allow a direct check of the nucleus-level steps. Their releases already fix the registration.
between the H\&E and DAPI images exactly, as a scale with no offset, so the agreement between the original registration and our align results provides a test of our processing pipeline.

We ran steps~1--5 over each slide in full, tiling it into $6000 \times 6000$
H\&E-pixel blocks with a $20$ px search radius and a $30$ px border discard, a block boundary only affects nuclei that the procedure already drops. Tiling is harmless because these steps are local by construction. Every nucleus that Cellpose returns on the slide is fed to the processing pipeline, and the
agreement below is measured over all of them.

The matching and filtering steps are reliable. Of the pairs the procedure retains, $98.7$--$99.8\%$ pairs get the same registration as the original one (Table~\ref{tab:hedapi-qc}).

\begin{table}[t]
\centering
\small
\caption{\textbf{Whole-slide pairing of H\&E with CosMx.} Every tissue block
of each slide. \emph{Agreement} is the fraction of retained pairs whose
matched CosMx cell coincides with the cell that the vendor H\&E--DAPI
registration places at that H\&E nucleus; the procedure has no access to that
registration. \emph{Pairs} is the number of retained pairs the fraction is
computed over.}
\label{tab:hedapi-qc}
\begin{tabular}{lrr}
\hline
Slide & Pairs & Agreement (\%) \\
\hline
Ovary & $14{,}320$ & 99.2 \\
Liver & $31{,}570$ & 99.8 \\
Colon & $25{,}058$ & 98.7 \\
\hline
\end{tabular}
\end{table}

\subsection{CosMx Zero-Shot Prediction}
\label{app:cosmx-zeroshot}
Running the pairing procedure of Section~\ref{app:dapi-align} on the three
CosMx slides gives an H\&E crop for each paired CosMx cell, and those cells can
then be predicted directly. This measures how much of VOICE survives a change
of platform before any CosMx data is used.

The transfer is strictly zero-shot. No CosMx expression data are used for model training. We cut the H\&E crops to the same
$55\,\mu$m field of view as the Xenium corpus, so the physical scale matches by
construction. The retrieval bank is drawn from same-tissue \emph{Xenium}
slides, and we fit the Stage-3 weights on those same Xenium reference slides
and transfer them, exactly as in Section~\ref{sec:xsgen}. Table~\ref{tab:cosmx-zeroshot} reports the result.

We score the $3{,}097$ genes that the CosMx $6{,}174$-gene panel shares with
the $6{,}029$-gene vocabulary of Section~\ref{sec:decoder}, since the count
head cannot predict the rest. How many of those genes the Xenium bank measures
varies by tissue. The gene numbers are $2{,}672$ on ovary, $577$ on colon, and $310$ on
liver. This limits the retrieval branch alone since the direct branch needs no bank
and predicts all $3{,}097$ genes. For the retrieval row we therefore depart from
the protocol of Section~\ref{sec:benchmark} and average only over the genes it
can reach. 
Table~\ref{tab:cosmx-zeroshot} reports the result.

\begin{table*}[t]
\centering
\small
\setlength{\tabcolsep}{4pt}
\caption{\textbf{Zero-shot transfer from Xenium to CosMx.} Bold marks the best
value in each column.}
\label{tab:cosmx-zeroshot}
\begin{tabular*}{\textwidth}{@{\extracolsep{\fill}}llccccccc}
\toprule
Slide & Branch & All & H20 & H50 & H100 & S20 & S50 & S100 \\
\midrule
\multirow{3}{*}{Ovary}
 & VOICE-23M Direct                & 0.0911 & 0.1844 & 0.1773 & 0.1711 & 0.1470 & 0.1496 & 0.1480 \\
 & VOICE-23M Retrieval             & 0.0774 & \textbf{0.2201} & \textbf{0.2105} & \textbf{0.1849} & \textbf{0.1786} & \textbf{0.1695} & \textbf{0.1607} \\
 & \textbf{VOICE-23M Fused} & \textbf{0.0917} & 0.1853 & 0.1783 & 0.1722 & 0.1480 & 0.1503 & 0.1489 \\
\midrule
\multirow{3}{*}{Liver}
 & VOICE-23M Direct                & 0.0494 & 0.1884 & 0.1602 & 0.1420 & 0.1899 & 0.1564 & 0.1376 \\
 & VOICE-23M Retrieval             & 0.0453 & 0.1775 & \textbf{0.1661} & \textbf{0.1633} & 0.1775 & 0.1511 & \textbf{0.1534} \\
 & \textbf{VOICE-23M Fused} & \textbf{0.0494} & \textbf{0.1889} & 0.1605 & 0.1421 & \textbf{0.1903} & \textbf{0.1568} & 0.1378 \\
\midrule
\multirow{3}{*}{Colon}
 & VOICE-23M Direct                & 0.0762 & 0.1227 & 0.1394 & 0.1363 & 0.0862 & 0.1023 & 0.1166 \\
 & VOICE-23M Retrieval             & 0.0569 & 0.0949 & 0.0919 & 0.0924 & 0.0477 & 0.0667 & 0.0772 \\
 & \textbf{VOICE-23M Fused} & \textbf{0.0762} & \textbf{0.1232} & \textbf{0.1397} & \textbf{0.1366} & \textbf{0.0862} & \textbf{0.1024} & \textbf{0.1168} \\
\midrule
\multirow{3}{*}{\textbf{Macro}}
 & VOICE-23M Direct                & 0.0722 & 0.1652 & 0.1590 & 0.1498 & 0.1410 & 0.1361 & 0.1340 \\
 & VOICE-23M Retrieval             & 0.0599 & 0.1642 & 0.1562 & 0.1469 & 0.1346 & 0.1291 & 0.1304 \\
 & \textbf{VOICE-23M Fused} & \textbf{0.0724} & \textbf{0.1658} & \textbf{0.1595} & \textbf{0.1503} & \textbf{0.1415} & \textbf{0.1365} & \textbf{0.1345} \\
\bottomrule
\end{tabular*}
\end{table*}

Prediction transfers but weakens sharply. The macro All score of $0.0724$ is
well below the $0.2126$ that the same model reaches across the Xenium slides
Breast-1, Lung-1, Kidney, Ovary-1, and Pancreas (Table~\ref{tab:cross-slide-target}), and that gap is the
result of the platform shift since nothing from the target platform is used. Which branch predicts better follows the bank coverage above. On ovary, where the bank
reaches $2{,}672$ genes, the retrieval branch is the stronger of the two on
every ranked subset, reaching $0.2201$ on H20 against $0.1844$ for the direct
branch while remaining behind on All, so transferring measurements from
morphologically similar cells survives the platform change better than
predicting them does. On colon, where the bank reaches only $577$, retrieval
is behind on all seven metrics, and the fusion falls back on the direct branch
and differs from it by at most $0.0005$.

The transferred gate is conservative. Fitting the per-gene weights on the
CosMx slide itself raises the macro All score from $0.0724$ to $0.0756$ and
H50 from $0.1595$ to $0.1667$. That variant uses target labels, so it is an
upper bound rather than a deployable number. The gap between the two is the
cost of never touching the target platform, and it is what a small amount of
CosMx supervision would be expected to recover.

\onecolumn
\section{Data provenance}
\label{app:data-provenance-23m}

Tables~\ref{tab:train-provenance-23m} and~\ref{tab:test-provenance}
list the $75$ training slides, three in-slide evaluation slides, and ten
cross-slide evaluation slides. Cross-slide identifiers distinguish specimens
from the same tissue and are used consistently throughout the paper.
Table~\ref{tab:train-provenance-7m} lists the training subset used for VOICE-7M.

\begingroup
\small
\setlength{\tabcolsep}{4pt}
\setlength{\LTcapwidth}{\textwidth}
\renewcommand{\arraystretch}{1.08}
\begin{longtable}{@{}p{0.43\textwidth}p{0.16\textwidth}rrp{0.17\textwidth}@{}}
\caption{\textbf{Training-data provenance for VOICE-23M.} The training split contains $75$ Xenium slides and $23{,}232{,}915$ cells across $15$ tissues. Rows group processed slides from the same public release; counts denote retained cells. HEST identifiers refer to samples reconstructed at single-cell resolution.}
\label{tab:train-provenance-23m}\\
\toprule
Source / dataset and samples & Tissue (condition) & Slides & Cells & Accession / link \\
\midrule
\endfirsthead
\multicolumn{5}{l}{\footnotesize\textit{Table~\thetable\ (continued)}}\\
\toprule
Source / dataset and samples & Tissue (condition) & Slides & Cells & Accession / link \\
\midrule
\endhead
\midrule
\multicolumn{5}{r}{\footnotesize\textit{Continued on next page}}\\
\endfoot
\bottomrule
\endlastfoot
\raggedright Human bone and bone marrow, custom add-on panel & \raggedright Bone, bone marrow (normal, ALL) & 3 & 344{,}181 & \raggedright \href{https://www.10xgenomics.com/datasets/human-bone-and-bone-marrow-data-with-custom-add-on-panel-1-standard}{10x Genomics} \tabularnewline
\raggedright FFPE brain cancer, IO panel and add-on; HEST TENX138 & \raggedright Brain (glioblastoma) & 1 & 816{,}519 & \raggedright \href{https://www.10xgenomics.com/datasets/ffpe-human-brain-cancer-data-with-human-immuno-oncology-profiling-panel-and-custom-add-on-1-standard}{10x Genomics} \tabularnewline
\raggedright FFPE breast, custom add-on panel; samples 3--5 (IDC, ILC) & \raggedright Breast cancer & 3 & 1{,}517{,}094 & \raggedright \href{https://www.10xgenomics.com/datasets/ffpe-human-breast-with-custom-add-on-panel-1-standard}{10x Genomics} \tabularnewline
\raggedright FFPE breast, pre-designed panel; samples 6--7 (IDC, ILC) & \raggedright Breast cancer & 2 & 931{,}598 & \raggedright \href{https://www.10xgenomics.com/datasets/ffpe-human-breast-with-pre-designed-panel-1-standard}{10x Genomics} \tabularnewline
\raggedright FFPE breast, entire sample area; sample 8, replicates 1--2 & \raggedright Breast cancer & 2 & 1{,}778{,}489 & \raggedright \href{https://www.10xgenomics.com/datasets/ffpe-human-breast-using-the-entire-sample-area-1-standard}{10x Genomics} \tabularnewline
\raggedright FFPE breast with biomarkers; HEST TENX191--TENX202, patients 1--4 & \raggedright Breast (DCIS, IDC, normal) & 12 & 3{,}429{,}268 & \raggedright \href{https://www.10xgenomics.com/datasets/xenium-ffpe-human-breast-biomarkers}{10x Genomics} \tabularnewline
\raggedright Xenium Prime FFPE cervical cancer; HEST TENX188 & \raggedright Cervical cancer & 1 & 836{,}989 & \raggedright \href{https://www.10xgenomics.com/datasets/xenium-prime-ffpe-human-cervical-cancer}{10x Genomics} \tabularnewline
\raggedright Human colon preview, non-diseased and cancer add-on & \raggedright Colon (normal, cancer) & 2 & 858{,}041 & \raggedright \href{https://www.10xgenomics.com/datasets/human-colon-preview-data-xenium-human-colon-gene-expression-panel-1-standard}{10x Genomics} \tabularnewline
\raggedright FFPE colorectal cancer, IO panel and add-on & \raggedright Colorectal cancer & 1 & 388{,}175 & \raggedright \href{https://www.10xgenomics.com/datasets/ffpe-human-colorectal-cancer-data-with-human-immuno-oncology-profiling-panel-and-custom-add-on-1-standard}{10x Genomics} \tabularnewline
\raggedright Colorectal cancer, Xenium sections; HEST TENX147--TENX149, patients 1, 2 and 5 & \raggedright Colorectal cancer & 3 & 922{,}936 & \raggedright \href{https://www.10xgenomics.com/products/visium-hd-spatial-gene-expression/dataset-human-crc}{10x Genomics} \tabularnewline
\raggedright Human heart, multi-tissue and cancer panel & \raggedright Heart (normal) & 1 & 26{,}364 & \raggedright \href{https://www.10xgenomics.com/datasets/human-heart-data-xenium-human-multi-tissue-and-cancer-panel-1-standard}{10x Genomics} \tabularnewline
\raggedright Human kidney preview, non-diseased and cancer sections & \raggedright Kidney (normal, papillary RCC) & 2 & 154{,}068 & \raggedright \href{https://www.10xgenomics.com/datasets/human-kidney-preview-data-xenium-human-multi-tissue-and-cancer-panel-1-standard}{10x Genomics} \tabularnewline
\raggedright Human liver, non-diseased and cancer sections & \raggedright Liver (normal, cancer) & 2 & 401{,}890 & \raggedright \href{https://www.10xgenomics.com/datasets/human-liver-data-xenium-human-multi-tissue-and-cancer-panel-1-standard}{10x Genomics} \tabularnewline
\raggedright Post-Xenium technical note, lung cancer; Xenium V1 and Prime 5K & \raggedright Lung cancer & 2 & 556{,}957 & \raggedright \href{https://www.10xgenomics.com/datasets/xenium-human-lung-cancer-post-xenium-technote}{10x Genomics} \tabularnewline
\raggedright FFPE lung cancer, multimodal segmentation; HEST TENX118 & \raggedright Lung cancer & 1 & 161{,}157 & \raggedright \href{https://www.10xgenomics.com/datasets/preview-data-ffpe-human-lung-cancer-with-xenium-multimodal-cell-segmentation-1-standard}{10x Genomics} \tabularnewline
\raggedright Pulmonary fibrosis cohort; HEST NCBI856--861, 864--867, 870, 873, 875--876 and 879--884 & \raggedright Lung (ILD, IPF, control) & 20 & 1{,}043{,}163 & \raggedright \href{https://www.ncbi.nlm.nih.gov/geo/query/acc.cgi?acc=GSE250346}{GEO GSE250346} \tabularnewline
\raggedright Lung tissue microarrays; HEST NCBI885--NCBI888 & \raggedright Lung & 4 & 4{,}427{,}689 & \raggedright \href{https://www.ncbi.nlm.nih.gov/geo/query/acc.cgi?acc=GSE315411}{GEO GSE315411}; GSM9427181--GSM9427184 \tabularnewline
\raggedright Xenium Prime preview, human lymph node & \raggedright Lymph node (reactive) & 1 & 708{,}981 & \raggedright \href{https://www.10xgenomics.com/datasets/preview-data-xenium-prime-gene-expression}{10x Genomics} \tabularnewline
\raggedright Fresh-frozen ovarian cancer, Xenium V1 and Prime 5K & \raggedright Ovarian cancer & 2 & 405{,}885 & \raggedright \href{https://www.10xgenomics.com/datasets/xenium-comparison-fresh-frozen-human-ovarian-cancer}{10x Genomics} \tabularnewline
\raggedright Xenium Prime FFPE ovarian cancer, XRrun & \raggedright Ovarian cancer & 1 & 407{,}124 & \raggedright \href{https://www.10xgenomics.com/datasets/xenium-prime-ffpe-human-ovarian-cancer}{10x Genomics} \tabularnewline
\raggedright FFPE pancreas, multimodal segmentation; HEST patient 1 & \raggedright Pancreatic cancer & 1 & 140{,}702 & \raggedright \href{https://www.10xgenomics.com/datasets/ffpe-human-pancreas-with-xenium-multimodal-cell-segmentation-1-standard}{10x Genomics} \tabularnewline
\raggedright Pancreatic cancer, multi-tissue panel and add-on; HEST patient 2 & \raggedright Pancreatic cancer & 1 & 190{,}964 & \raggedright \href{https://www.10xgenomics.com/datasets/pancreatic-cancer-with-xenium-human-multi-tissue-and-cancer-panel-1-standard}{10x Genomics} \tabularnewline
\raggedright Xenium Prime FFPE human prostate & \raggedright Prostate cancer & 1 & 193{,}000 & \raggedright \href{https://www.10xgenomics.com/datasets/xenium-prime-ffpe-human-prostate}{10x Genomics} \tabularnewline
\raggedright Human skin, two non-diseased sections & \raggedright Skin (normal) & 2 & 158{,}579 & \raggedright \href{https://www.10xgenomics.com/datasets/human-skin-data-xenium-human-multi-tissue-and-cancer-panel-1-standard}{10x Genomics} \tabularnewline
\raggedright Human skin preview; HEST TENX115 & \raggedright Skin (melanoma) & 1 & 106{,}791 & \raggedright \href{https://www.10xgenomics.com/datasets/human-skin-preview-data-xenium-human-skin-gene-expression-panel-1-standard}{10x Genomics} \tabularnewline
\raggedright Xenium Prime FFPE skin; HEST TENX158 & \raggedright Skin (melanoma) & 1 & 112{,}322 & \raggedright \href{https://www.10xgenomics.com/datasets/xenium-prime-ffpe-human-skin}{10x Genomics} \tabularnewline
\raggedright Human tonsil, follicular and reactive hyperplasia sections & \raggedright Tonsil & 2 & 2{,}213{,}989 & \raggedright \href{https://www.10xgenomics.com/datasets/human-tonsil-data-xenium-human-multi-tissue-and-cancer-panel-1-standard}{10x Genomics} \tabularnewline
\midrule
\textbf{Training total} & & \textbf{75} & \textbf{23{,}232{,}915} & \\
\end{longtable}

\clearpage
\begin{longtable}{@{}p{0.12\textwidth}p{0.28\textwidth}rp{0.17\textwidth}p{0.25\textwidth}@{}}
\caption{\textbf{Evaluation-data provenance for VOICE-23M.} The three in-slide benchmarks contain $391{,}224$ cells and use spatial five-fold cross-validation. The ten cross-slide targets contain $3{,}845{,}839$ cells across seven tissues and are evaluated without target-slide fitting. Counts denote retained cells. Head and neck and stomach are absent from training.}
\label{tab:test-provenance}\\
\toprule
Slide & Dataset and specimen & Cells & Accession / link & Cross-slide separation \\
\midrule
\endfirsthead
\multicolumn{5}{l}{\footnotesize\textit{Table~\thetable\ (continued)}}\\
\toprule
Slide & Dataset and specimen & Cells & Accession / link & Cross-slide separation \\
\midrule
\endhead
\midrule
\multicolumn{5}{r}{\footnotesize\textit{Continued on next page}}\\
\endfoot
\bottomrule
\endlastfoot
\multicolumn{5}{l}{\textit{In-slide evaluation: three slides}}\\
\raggedright Breast & \raggedright FFPE breast cancer, Sample~2 (Janesick et al.) & 142{,}272 & \raggedright \href{https://www.ncbi.nlm.nih.gov/geo/query/acc.cgi?acc=GSE243275}{GEO GSE243275}; \href{https://www.10xgenomics.com/products/xenium-in-situ/preview-dataset-human-breast}{10x Genomics} & \raggedright -- \tabularnewline
\raggedright Lung & \raggedright FFPE lung cancer, multimodal segmentation; Sample~1 & 161{,}453 & \raggedright \href{https://www.10xgenomics.com/datasets/preview-data-ffpe-human-lung-cancer-with-xenium-multimodal-cell-segmentation-1-standard}{10x Genomics} & \raggedright -- \tabularnewline
\raggedright Skin & \raggedright Human skin preview, melanoma add-on; Sample~1 & 87{,}499 & \raggedright \href{https://www.10xgenomics.com/datasets/human-skin-preview-data-xenium-human-skin-gene-expression-panel-add-on-1-standard}{10x Genomics} & \raggedright -- \tabularnewline
\midrule
\multicolumn{2}{l}{\textbf{Subtotal (3 slides)}} & \textbf{391{,}224} & & \\
\midrule
\multicolumn{5}{l}{\textit{Cross-slide evaluation: ten slides}}\\
\raggedright Breast-1 & \raggedright Xenium Prime FFPE breast cancer; Discovery Life Sciences & 699{,}078 & \raggedright \href{https://www.10xgenomics.com/datasets/xenium-prime-ffpe-human-breast-cancer}{10x Genomics} & \raggedright Distinct slide; donor identifiers unavailable. \tabularnewline
\raggedright Breast-2 & \raggedright FFPE breast cancer, Sample~1, Replicate~2; Janesick et al. & 112{,}068 & \raggedright \href{https://www.ncbi.nlm.nih.gov/geo/query/acc.cgi?acc=GSE243275}{GEO GSE243275} & \raggedright Both sections from Sample~1 excluded from training. \tabularnewline
\raggedright Head and neck & \raggedright S8; primary tumor~1; p16-negative oropharyngeal SCC & 18{,}421 & \raggedright \href{https://www.ncbi.nlm.nih.gov/geo/query/acc.cgi?acc=GSE324934}{GEO GSE324934}; GSM9589065 & \raggedright Independent cohort; specimen absent from training. \tabularnewline
\raggedright Kidney & \raggedright Xenium Protein FFPE renal carcinoma; Slide~01 & 465{,}534 & \raggedright \href{https://www.10xgenomics.com/datasets/xenium-protein-ffpe-human-renal-carcinoma}{10x Genomics} & \raggedright Training kidney specimens supplied by AcePix; test specimen by DLS. \tabularnewline
\raggedright Lung-1 & \raggedright FFPE lung cancer, IO panel and add-on; HEST TENX141, patient~1 & 160{,}444 & \raggedright \href{https://www.10xgenomics.com/datasets/ffpe-human-lung-cancer-data-with-human-immuno-oncology-profiling-panel-and-custom-add-on-1-standard}{10x Genomics} & \raggedright Distinct HEST patient held out. \tabularnewline
\raggedright Lung-2 & \raggedright CHUV primary NSCLC; patient L1 & 66{,}729 & \raggedright \href{https://www.ncbi.nlm.nih.gov/geo/query/acc.cgi?acc=GSE311609}{GEO GSE311609}; GSM9509134 & \raggedright Independent cohort; patient absent from training. \tabularnewline
\raggedright Ovary-1 & \raggedright FFPE ovarian cancer, IO panel and add-on; HEST TENX142 & 247{,}636 & \raggedright \href{https://www.10xgenomics.com/datasets/ffpe-human-ovarian-cancer-data-with-human-immuno-oncology-profiling-panel-and-custom-add-on-1-standard}{10x Genomics} & \raggedright Distinct experiment and ROI identifiers; donor identifiers unavailable. \tabularnewline
\raggedright Ovary-2 & \raggedright Xenium Prime fresh-frozen ovary; BioIVT donor~1 & 1{,}151{,}806 & \raggedright \href{https://www.10xgenomics.com/datasets/xenium-prime-fresh-frozen-human-ovary}{10x Genomics} & \raggedright Distinct sample; supplier differs from training. \tabularnewline
\raggedright Pancreas & \raggedright FFPE pancreatic ductal adenocarcinoma, IO panel; HEST TENX140, patient~3 & 234{,}856 & \raggedright \href{https://www.10xgenomics.com/datasets/ffpe-human-ductal-adenocarcinoma-data-with-human-immuno-oncology-profiling-panel-1-standard}{10x Genomics} & \raggedright Training pancreas specimens are HEST patients~1 and~2. \tabularnewline
\raggedright Stomach & \raggedright iSCALE, SNUH patient P1; specimen BS06-9313-8 & 689{,}267 & \raggedright \href{https://zenodo.org/records/15164980}{Zenodo 15164980} & \raggedright Independent cohort; patient and specimen absent from training. \tabularnewline
\midrule
\multicolumn{2}{l}{\textbf{Subtotal (10 slides)}} & \textbf{3{,}845{,}839} & & \\
\end{longtable}
\endgroup

\clearpage
\begingroup
\small
\setlength{\tabcolsep}{4pt}
\setlength{\LTcapwidth}{\textwidth}
\renewcommand{\arraystretch}{1.08}
\begin{longtable}{@{}p{0.43\textwidth}p{0.16\textwidth}rrp{0.17\textwidth}@{}}
\caption{\textbf{Training-data provenance for VOICE-7M.} This subset of the VOICE-23M training corpus contains $24$ Xenium slides and $6{,}979{,}810$ cells across $11$ tissues. Rows group slides from the same public release; counts denote retained cells.}
\label{tab:train-provenance-7m}\\
\toprule
Source / dataset and samples & Tissue (condition) & Slides & Cells & Accession / link \\
\midrule
\endfirsthead
\multicolumn{5}{l}{\footnotesize\textit{Table~\thetable\ (continued)}}\\
\toprule
Source / dataset and samples & Tissue (condition) & Slides & Cells & Accession / link \\
\midrule
\endhead
\midrule
\multicolumn{5}{r}{\footnotesize\textit{Continued on next page}}\\
\endfoot
\bottomrule
\endlastfoot
\raggedright Human bone and bone marrow, custom add-on panel & \raggedright Bone, bone marrow (normal, ALL) & 3 & 344{,}181 & \raggedright \href{https://www.10xgenomics.com/datasets/human-bone-and-bone-marrow-data-with-custom-add-on-panel-1-standard}{10x Genomics} \tabularnewline
\raggedright FFPE breast, custom add-on panel; samples 3--5 (IDC, ILC) & \raggedright Breast cancer & 3 & 1{,}517{,}094 & \raggedright \href{https://www.10xgenomics.com/datasets/ffpe-human-breast-with-custom-add-on-panel-1-standard}{10x Genomics} \tabularnewline
\raggedright FFPE breast, pre-designed panel; samples 6--7 (IDC, ILC) & \raggedright Breast cancer & 2 & 931{,}598 & \raggedright \href{https://www.10xgenomics.com/datasets/ffpe-human-breast-with-pre-designed-panel-1-standard}{10x Genomics} \tabularnewline
\raggedright FFPE breast, entire sample area; sample 8, replicates 1--2 & \raggedright Breast cancer & 2 & 1{,}778{,}489 & \raggedright \href{https://www.10xgenomics.com/datasets/ffpe-human-breast-using-the-entire-sample-area-1-standard}{10x Genomics} \tabularnewline
\raggedright Human colon preview, non-diseased section & \raggedright Colon (normal) & 1 & 270{,}984 & \raggedright \href{https://www.10xgenomics.com/datasets/human-colon-preview-data-xenium-human-colon-gene-expression-panel-1-standard}{10x Genomics} \tabularnewline
\raggedright FFPE colorectal cancer, IO panel and add-on & \raggedright Colorectal cancer & 1 & 388{,}175 & \raggedright \href{https://www.10xgenomics.com/datasets/ffpe-human-colorectal-cancer-data-with-human-immuno-oncology-profiling-panel-and-custom-add-on-1-standard}{10x Genomics} \tabularnewline
\raggedright Human heart, multi-tissue and cancer panel & \raggedright Heart (normal) & 1 & 26{,}364 & \raggedright \href{https://www.10xgenomics.com/datasets/human-heart-data-xenium-human-multi-tissue-and-cancer-panel-1-standard}{10x Genomics} \tabularnewline
\raggedright Human kidney preview, non-diseased and cancer sections & \raggedright Kidney (normal, papillary RCC) & 2 & 154{,}068 & \raggedright \href{https://www.10xgenomics.com/datasets/human-kidney-preview-data-xenium-human-multi-tissue-and-cancer-panel-1-standard}{10x Genomics} \tabularnewline
\raggedright Human liver, non-diseased and cancer sections & \raggedright Liver (normal, cancer) & 2 & 401{,}890 & \raggedright \href{https://www.10xgenomics.com/datasets/human-liver-data-xenium-human-multi-tissue-and-cancer-panel-1-standard}{10x Genomics} \tabularnewline
\raggedright Post-Xenium technical note, lung cancer; Xenium V1 & \raggedright Lung cancer & 1 & 278{,}652 & \raggedright \href{https://www.10xgenomics.com/datasets/xenium-human-lung-cancer-post-xenium-technote}{10x Genomics} \tabularnewline
\raggedright Fresh-frozen ovarian cancer, Xenium V1 & \raggedright Ovarian cancer & 1 & 205{,}070 & \raggedright \href{https://www.10xgenomics.com/datasets/xenium-comparison-fresh-frozen-human-ovarian-cancer}{10x Genomics} \tabularnewline
\raggedright FFPE pancreas, multimodal segmentation; HEST patient 1 & \raggedright Pancreatic cancer & 1 & 140{,}702 & \raggedright \href{https://www.10xgenomics.com/datasets/ffpe-human-pancreas-with-xenium-multimodal-cell-segmentation-1-standard}{10x Genomics} \tabularnewline
\raggedright Pancreatic cancer, multi-tissue panel and add-on; HEST patient 2 & \raggedright Pancreatic cancer & 1 & 190{,}964 & \raggedright \href{https://www.10xgenomics.com/datasets/pancreatic-cancer-with-xenium-human-multi-tissue-and-cancer-panel-1-standard}{10x Genomics} \tabularnewline
\raggedright Xenium Prime FFPE human prostate & \raggedright Prostate cancer & 1 & 193{,}000 & \raggedright \href{https://www.10xgenomics.com/datasets/xenium-prime-ffpe-human-prostate}{10x Genomics} \tabularnewline
\raggedright Human skin, two non-diseased sections & \raggedright Skin (normal) & 2 & 158{,}579 & \raggedright \href{https://www.10xgenomics.com/datasets/human-skin-data-xenium-human-multi-tissue-and-cancer-panel-1-standard}{10x Genomics} \tabularnewline
\midrule
\textbf{Training total} & & \textbf{24} & \textbf{6{,}979{,}810} & \\
\end{longtable}
\endgroup

\end{document}